%% file: main.tex
\documentclass[lettersize,journal]{IEEEtran}
\usepackage{amsmath,amsfonts}
\usepackage{algorithmic}
\usepackage{algorithm}
\usepackage{array}
\usepackage[caption=false,font=normalsize,labelfont=sf,textfont=sf]{subfig}
\usepackage{textcomp}
\usepackage{stfloats}
\usepackage{url}
\usepackage{verbatim}
\usepackage{graphicx}
\usepackage{caption}
\usepackage{subcaption}
\usepackage{xcolor}
\usepackage{cite}
\usepackage{booktabs}
\usepackage{makecell}
\usepackage{multirow}
\usepackage[hidelinks]{hyperref}
\usepackage{cleveref}
\usepackage[acronym,nonumberlist]{glossaries}
\input{acronyms.tex}
\definecolor{lightgray}{rgb}{0.8, 0.8, 0.8}

\usepackage{pifont}
\newcommand{\cmark}{\ding{51}}%
\newcommand{\xmark}{\ding{55}}%

\begin{document}

\title{A Systematic Literature Review \\on Vehicular Collaborative Perception -- \\A Computer Vision Perspective}

\author{Lei Wan, Jianxin Zhao, Andreas Wiedholz, Manuel Bied, Mateus Martinez de Lucena, \\Abhishek Dinkar Jagtap, Andreas Festag,~\IEEEmembership{Senior Member,~IEEE}, \\Antônio Augusto Fröhlich,~\IEEEmembership{Senior~Member,~IEEE}, Hannan Ejaz Keen, Alexey Vinel.~\IEEEmembership{Senior~Member,~IEEE}

\thanks{Lei Wan, Andreas Wiedholz and Hannan Ejaz Keen are with XITASO GmbH, 86153 Augsburg, Germany
(e-mail: \{lei.wan, andreas.wiedholz, \\hannan.keen\}@xitaso.com).}
\thanks{Lei Wan, Jianxin Zhao, Manuel Bied and Alexey Vinel are with Karlsruhe Institute of Technology (KIT), 76133 Karlsruhe, Germany. 
(e-mail: lei.wan@partner.kit.edu, \{jianxin.zhao, manuel.bied, alexey.vinel\}@kit.edu).}
\thanks{Abhishek Dinkar Jagtap and Andreas Festag are with Technische Hochschule Ingolstadt (THI), CARISSMA Institute of Electric, Connected and Secure Mobility (C-ECOS), 85049 Ingolstadt, Germany (e-mail: \{abhishekdinkar.jagtap, andreas.festag\}@carissma.eu).}
\thanks{Mateus Martinez de Lucena and Antônio Augusto Fröhlich are with Federal University of Santa Catarina (UFSC), Brazil (e-mail: lucena@lisha.ufsc.br and guto@lisha.ufsc.br).}
\thanks{Figure~\ref{fig:cp_schematic} was created with the Car2Car Communication Consortium \mbox{(C2C-CC)} illustration toolkit.}
}

\markboth{Submitted to IEEE Transactions on Intelligent Transportation Systems.}%
{L. Wan \MakeLowercase{\textit{et al.}}: A Systematic Literature Review on Vehicular Collaborative Perception -- A Computer Vision Perspective}


\maketitle


\begin{abstract}
The effectiveness of autonomous vehicles relies on reliable perception capabilities. Despite significant advancements in artificial intelligence and sensor fusion technologies, current single-vehicle perception systems continue to encounter limitations, notably visual occlusions and limited long-range detection capabilities. \acrfull{CP}, enabled by \acrfull{V2V} and \acrfull{V2I} communication, has emerged as a promising solution to mitigate these issues and enhance the reliability of autonomous systems. Beyond advancements in communication, the computer vision community is increasingly focusing on improving vehicular perception through collaborative approaches. However, a systematic literature review that thoroughly examines existing work and reduces subjective bias is still lacking. Such a systematic approach helps identify research gaps, recognize common trends across studies, and inform future research directions. In response, this study follows the PRISMA 2020 guidelines and includes 106 peer-reviewed articles. These publications are analyzed based on modalities, collaboration schemes, and key perception tasks. Through a comparative analysis, this review illustrates how different methods address practical issues such as pose errors, temporal latency, communication constraints, domain shifts, heterogeneity, and adversarial attacks. Furthermore, it critically examines evaluation methodologies, highlighting a misalignment between current metrics and CP’s fundamental objectives. By delving into all relevant topics in-depth, this review offers valuable insights into challenges, opportunities, and risks, serving as a reference for advancing research in vehicular collaborative perception.

\end{abstract}


\input{sections/01_introduction}
\input{sections/02_research_methodology}
\input{sections/03_taxonomy}
\input{sections/04_evaluation}
\input{sections/05_challenges_opportunities_risks}
\input{sections/06_conclusion}
\input{sections/07_appendix}

\bibliographystyle{IEEEtran}
\bibliography{IEEEabrv,main}


\begin{IEEEbiography}[{\includegraphics[width=1in,height=1.25in,clip,keepaspectratio]{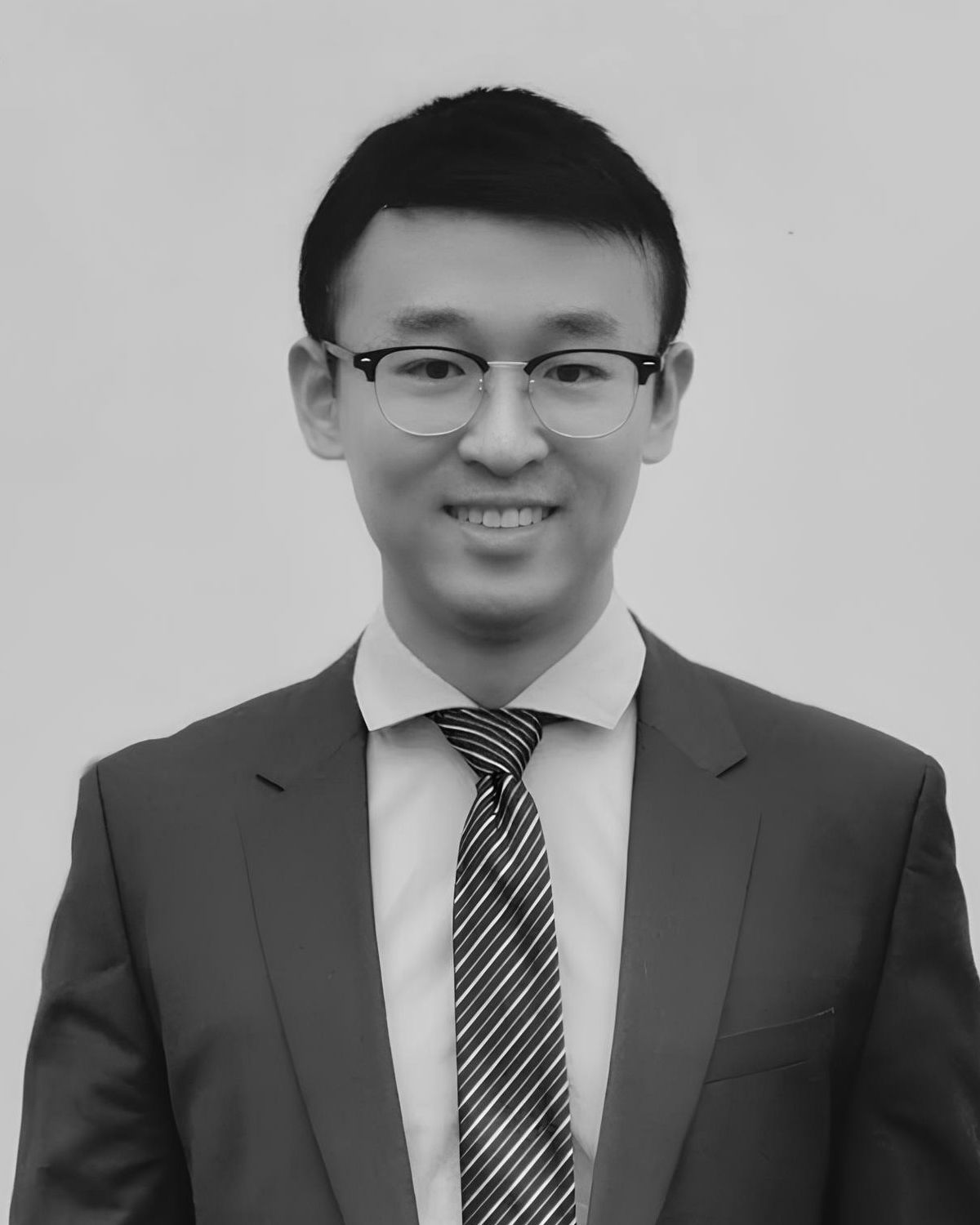}}]{Lei Wan}
is an external Ph.D student at the Karlsruhe Institute of Technology (KIT), Germany, and works as a full-time research engineer at XITASO GmbH, Germany. He received the Bachelor’s degree in Mechanical Engineering from Harbin Engineering University, China, and the Master's Degree in Mechatronics and Information Technology from the Karlsruhe Institute of Technology, Germany. His research fields are related to autonomous driving, computer vision, sensor fusion, and collaborative perception with V2X, whose goal is to improve the safety of autonomous driving through the development of robust perception systems.
\end{IEEEbiography}

\begin{IEEEbiography}[{\includegraphics[width=1in,height=1.25in,clip,keepaspectratio]{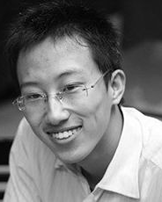}}]{Jianxin Zhao} 
is a postdoc at the Karlsruhe Institute of Technology (KIT), Germany. His research interests include scientific computation, machine learning, their application and optimization in the real world. In the postdoctoral work, he is researching the use of machine learning methods in cooperative perception, with a focus on the impact of high dynamism in the cooperative autonomous systems.
\end{IEEEbiography}

\begin{IEEEbiography}[{\includegraphics[width=1in,height=1.25in,clip,keepaspectratio]{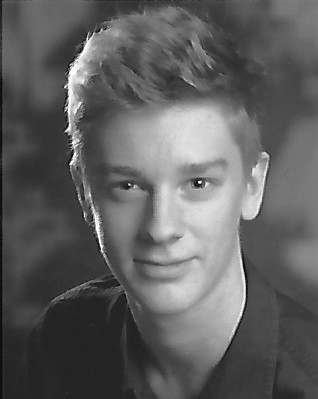}}]{Andreas Wiedholz}
is a full-time researcher at XITASO GmbH, Germany. He received his Bachelor's degree in Computer Engineering and his Master's degree in Robotics and AI at the Technical University of Applied Sciences in Augsburg. His research interests are related to perception and modeling of autonomous behavior in robotic systems.
\end{IEEEbiography}

\begin{IEEEbiography}[{\includegraphics[width=1in,height=1.25in,clip,keepaspectratio]{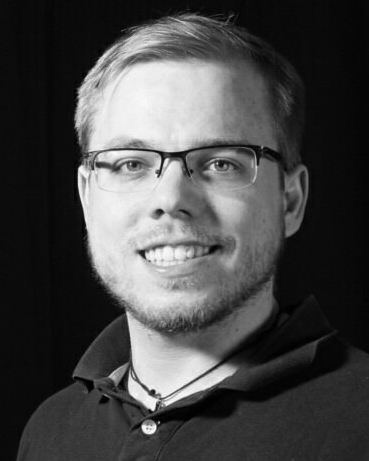}}]{Manuel Bied}
is a postdoctoral researcher at the Karlsruhe Institute of Technology (KIT), Germany. He received his Ph.D. degree in robotics from Sorbonne University, France and his Master's and Bachelor's degree in Electrical Engineering and Information Technology from Technical University Darmstadt, Germany.  His research interests include human-robot interaction, V2X, machine learning, and collective perception. In his work, he is focusing on the use of robots in traffic with the aim of increasing road safety.

\end{IEEEbiography}

\begin{IEEEbiography}[{\includegraphics[width=1in,height=1.25in,clip,keepaspectratio]{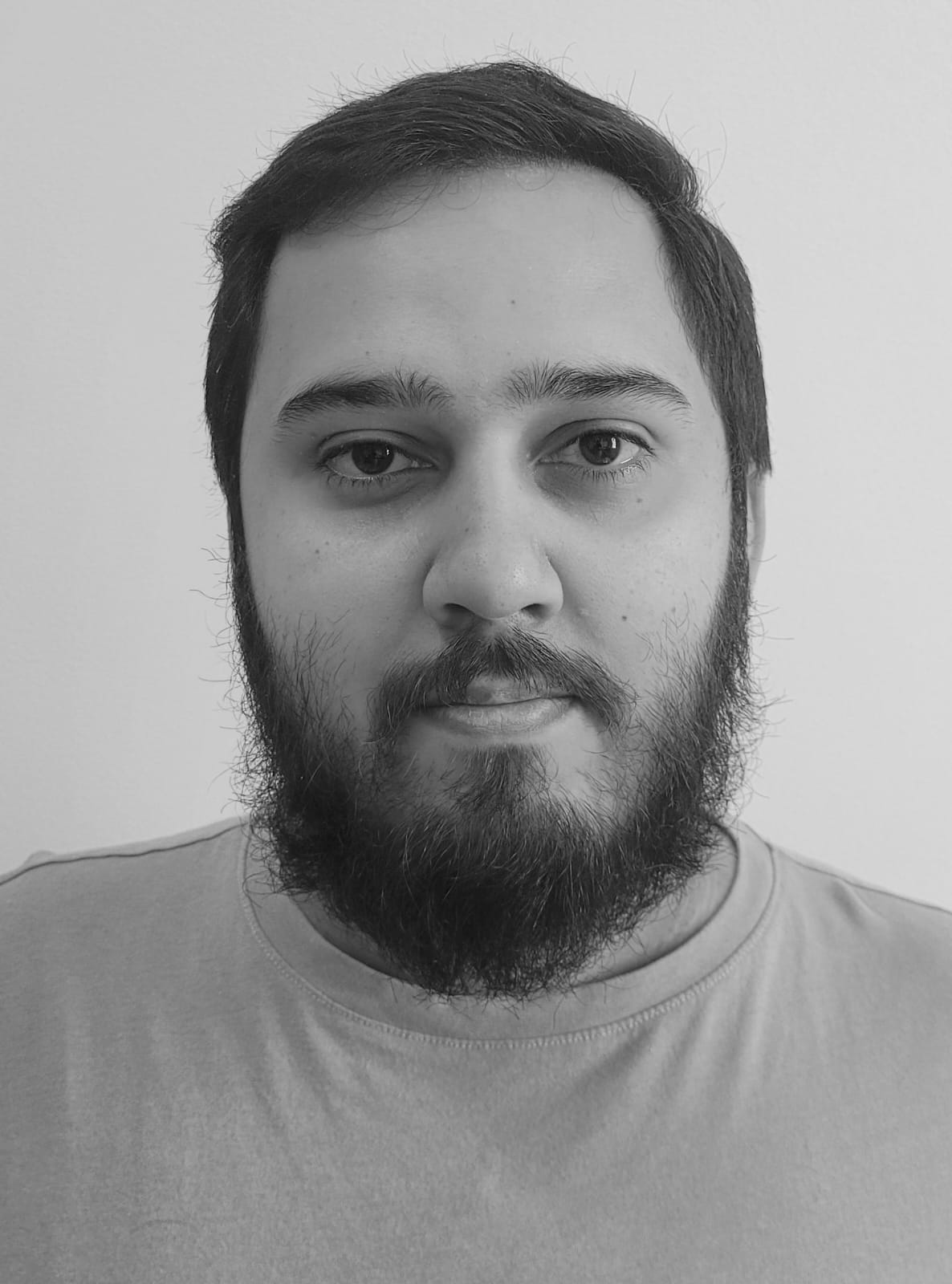}}]{Mateus Martinez de Lucena}
is a Federal University of Santa Catarina doctoral candidate supervised by Prof. Dr. Antônio Augusto Fröhlich. He graduated in Computer Engineering at the Federal University of Amazonas and completed his Masters in Computer Science at UFSC. He is researching collaborative perception in vehicular networks from a consensus perspective.
\end{IEEEbiography}

\begin{IEEEbiography}[{\includegraphics[width=1in,height=1.25in,clip,keepaspectratio]{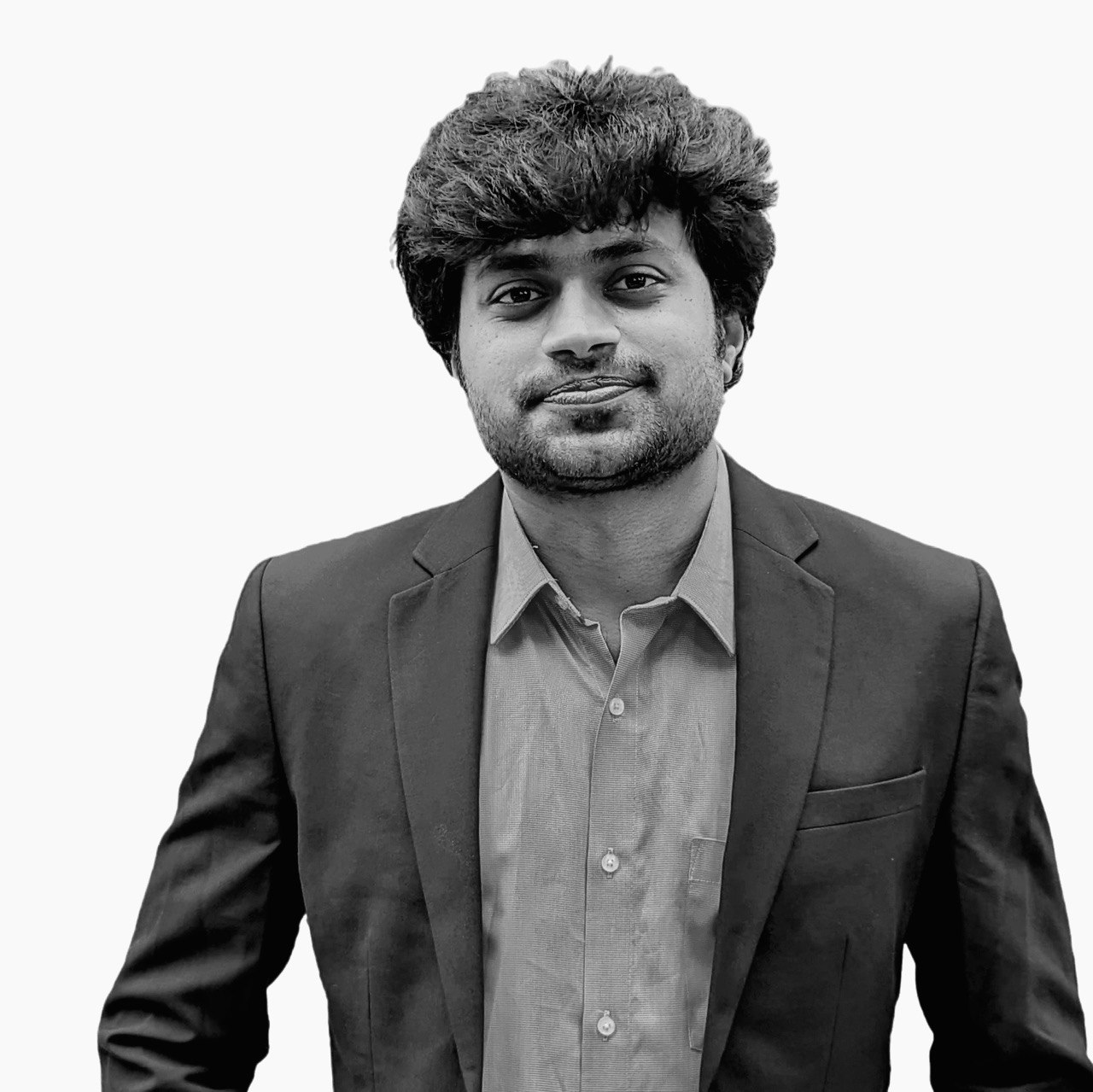}}]{Abhishek Dinkar Jagtap}
 is a Ph.D. student at Technische Hochschule Ingolstadt, Germany, and affiliated with the Center of Automotive Research on Integrated Safety Systems and Measurement Area CARISSMA, a leading research and testing facility for vehicle safety. He obtained his Master’s degree in Robotics and Autonomous Systems from Universität zu Lübeck, Germany. His research focuses on Vehicle-to-Everything (V2X) communication, computer vision, and cooperative perception. Specifically, he aims to enhance the efficiency of V2X communication-based cooperative perception by leveraging intermediate features from deep learning models. 
\end{IEEEbiography}

\begin{IEEEbiography}[{\includegraphics[width=1in,height=1.25in,clip,keepaspectratio]{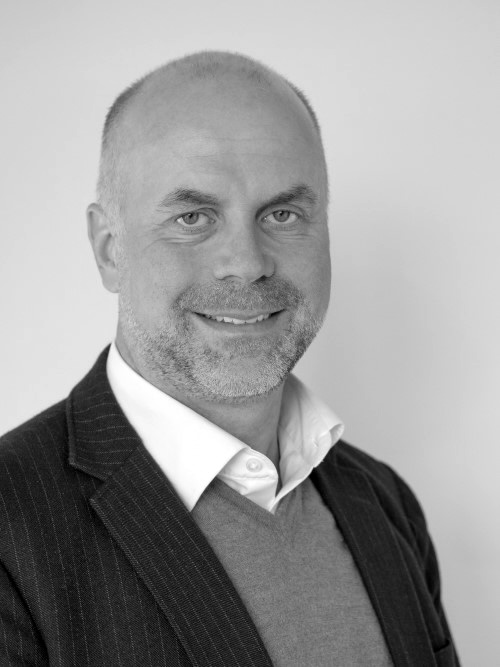}}]{Andreas Festag} received his Ph.D. degree in electrical engineering from the Berlin Institute of Technology, Berlin, Germany, in 2003. He has held research positions at the Telecommunication Networks Group, Berlin Institute of Technology, Heinrich-Hertz-Institute (HHI), Berlin, NEC Laboratories, Heidelberg, Germany, Vodafone Chair Mobile Communication Systems, Dresden University of Technology, Dresden, Germany, and the Fraunhofer Institute for Transportation and Infrastructure Systems, Dresden. Currently, he is a Professor at Technische Hochschule Ingolstadt, Germany and is affiliated with the Center of Automotive Research on Integrated Safety Systems and Measurement Area CARISSMA. He is also deputy head of the Fraunhofer application center for ``Connected Mobility and Infrastructure`` in Ingolstadt. His research interests include architecture, design, and performance evaluation of wireless and mobile communication systems and protocols, with an emphasis on vehicular communication and cooperative intelligent transportation systems (C-ITS).
\end{IEEEbiography}

\begin{IEEEbiography}[{\includegraphics[width=1in,height=1.25in,clip,keepaspectratio]{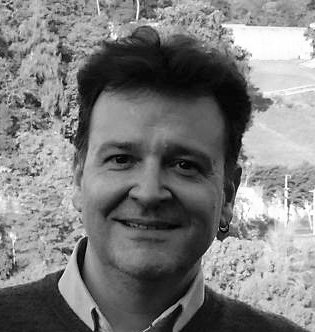}}]{Antônio Augusto Fröhlich}
is a Full Professor at the Federal University of Santa Catarina (UFSC), Brazil, where he leads the Software/Hardware Integration Lab (LISHA) since 2001. With a Ph.D. in Computer Engineering from the Technical University of Berlin, Germany, he has coordinated several R\&D projects on embedded systems, including the ALTATV Open, Free, Scalable Digital TV Platform, the CIA² research network on Smart Cities and the Internet of Things, and the Smart Campus project at UFSC. Significant contributions from these projects materialized within the Brazilian Digital Television System (SBTVD) and IoT technology for Energy Distribution, Smart Cities, and Precision Agriculture. He is a senior member of ACM, IEEE, and SBC.
\end{IEEEbiography}

\begin{IEEEbiography}[{\includegraphics[width=1in,height=1.25in,clip,keepaspectratio]{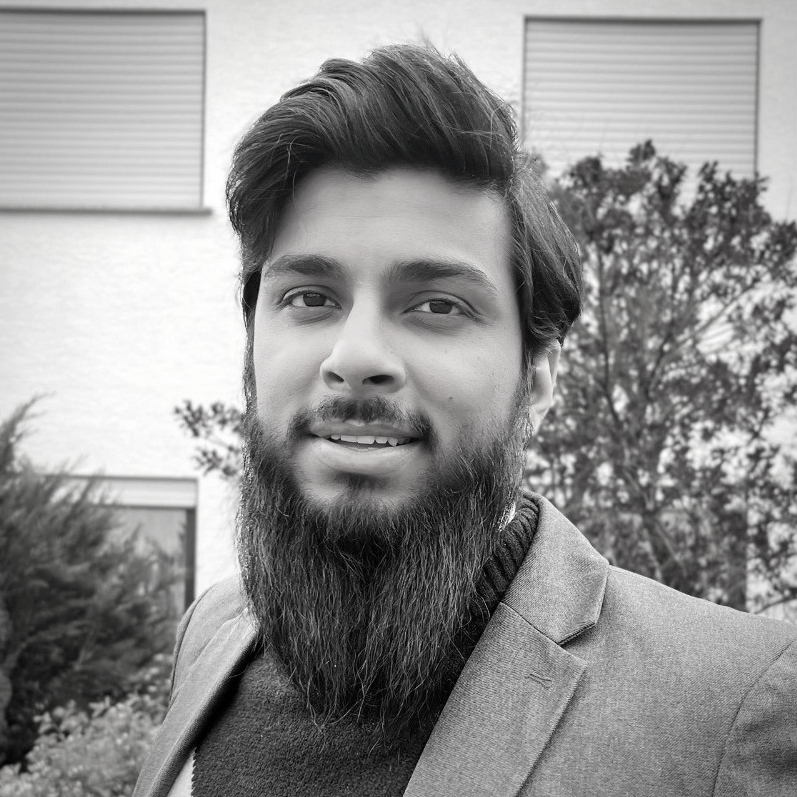}}]{Hannan Ejaz Keen}
is a Senior Researcher in the team of Autonomous Systems at XITASO GmbH. Prior to this role, he was a researcher at the Robotics Research Lab, RPTU Kaiserslautern-Landau, Germany, where he earned his Ph.D. in Autonomous Off-road Robotics. His research interests include sensor fusion, perception, and mapping. He holds an M.S. in Electrical Engineering from Lahore University of Management Sciences (LUMS) and a B.Sc. in Electrical Engineering from the University of Engineering and Technology (UET), Lahore, Pakistan.
\end{IEEEbiography}

\begin{IEEEbiography}[{\includegraphics[width=1in,height=1.25in,clip,keepaspectratio]{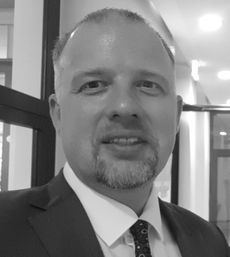}}]{Alexey Vinel}
is a professor at the Karlsruhe Institute of Technology (KIT), Germany. Previously he was a professor at the University of Passau, Germany. Since 2015, he has been a professor at Halmstad University, Sweden (now part-time). He received the Ph.D. degree from the Tampere University of Technology, Finland, in 2013. He has been the Senior Member of the IEEE since 2012. His areas of interests include vehicular communications and networking, cooperative automated and autonomous driving, future smart mobility solutions.
\end{IEEEbiography}

\end{document}

%% file: acronyms.tex
\newacronym{AD}{AD}{Autonomous Driving}
\newacronym{SLR}{SLR}{Systematic Literature Review}
\newacronym{CP}{CP}{Collaborative Perception}
\newacronym{CNN}{CNN}{Convolutional Neural Network}
\newacronym{ETSI}{ETSI}{European Telecommunications Standards Institute}
\newacronym{PDU}{PDU}{Packet Data Unit}
\newacronym{ITS}{ITS}{Intelligent Transportation System}
\newacronym{C-ITS}{C-ITS}{Cooperative-Intelligent Transportation Systems}

\newacronym{V2X}{V2X}{Vehicle-to-Everything}
\newacronym{V2V}{V2V}{Vehicle-to-Vehicle}
\newacronym{V2I}{V2I}{Vehicle-to-Infrastructure}
\newacronym{V2P}{V2P}{Vehicle-to-Pedestrian}
\newacronym{V2N}{V2N}{Vehicle-to-Network}
\newacronym{IoV}{IoV}{Internet of Vehicles}
\newacronym{LSTM}{LSTM}{Long Short-Term Memory}
\newacronym{MLP}{MLP}{Multilayer Perceptron}

\newacronym{FOV}{FOV}{Field of View}
\newacronym{ROIs}{ROIs}{regions of interest}
\newacronym{AVs}{AVs}{Autonomous Vehicles}
\newacronym{CAVs}{CAVs}{Connected Autonomous Vehicles}
\newacronym{CAV}{CAV}{Connected Autonomous Vehicle}
\newacronym{BEV}{BEV}{Bird’s Eye View}
\newacronym{GNN}{GNN}{Graph Neural Network}

\newacronym{COD}{COD}{Collaborative Object Detection}
\newacronym{COT}{COT}{Collaborative Object Tracking}
\newacronym{CMP}{CMP}{Collaborative Motion Prediction}
\newacronym{CSS}{CSS}{Collaborative Semantic Segmentation}
\newacronym{CLD}{CLD}{Collaborative Lane Detection}
\newacronym{CSC}{CSC}{Collaborative Scene Completion}

\newacronym{OD}{OD}{Object Detection}
\newacronym{OT}{OT}{Object Tracking}
\newacronym{MP}{MP}{Motion Prediction}
\newacronym{SS}{SS}{Semantic Segmentation}
\newacronym{LD}{LD}{Lane Detection}
\newacronym{MT}{MT}{Multi-Task}
\newacronym{TA}{TA}{Task-agnostic}

\newacronym{ICP}{ICP}{Iterative Closest Point}
\newacronym{HD Map}{HD Map}{High-definition Map}
\newacronym{PandP}{P\&P}{Perception and Prediction}
\newacronym{SLAM}{SLAM}{Simultaneous Localization and Mapping}

\newacronym{OEMs}{OEMs}{Original Equipment Manufacturers}
\newacronym{AI}{AI}{Artificial Intelligence}
\newacronym{ADAS}{ADAS}{Advanced Driver Assistance Systems}

\newacronym{CAM}{CAM}{Cooperative Awareness Message}
\newacronym{CPM}{CPM}{Collective Perception Message}

%% file: sections/01_introduction.tex
\section{Introduction} \label{sec:intro}

\begin{figure}[t]
\centering
\fbox{\includegraphics[width=0.9\linewidth]{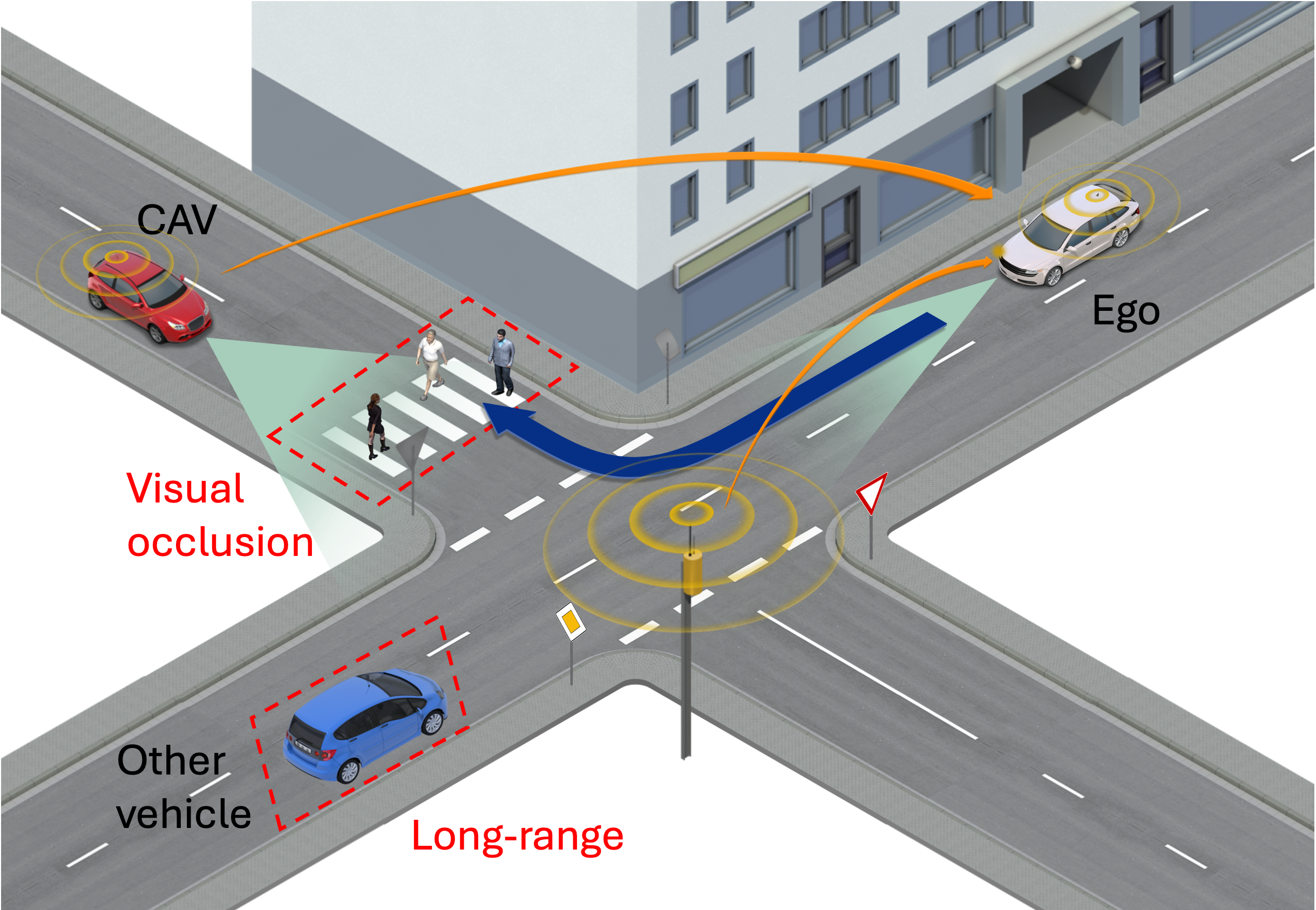}}
\caption{Illustration of a road traffic scenario for \acrfull{CP}: The green shaded areas represent the ego-vehicle’s (white) and the CAV’s (red) \gls{FOV}. The ego vehicle cannot perceive the pedestrians on its right due to the visual occlusion caused by a building, blocking its line of sight. Additionally, another vehicle (blue) on the opposite side of the intersection lies outside the ego vehicle’s perception range, presenting as the long-range problem. However, the CAV and infrastructure roadside unit can detect the pedestrians and the other vehicle, respectively, and share their observations with the ego vehicle, thereby enhancing its situational awareness.}
\label{fig:cp_schematic}
\end{figure}

\noindent \gls{AVs} are a crucial technology for intelligent transportation systems, offering the potential to significantly enhance road safety and transportation efficiency. With the emergence of \gls{V2V} and \gls{V2I} communication, \gls{CAVs} advance this potential by enabling data sharing not only among vehicles but also with traffic management systems, thereby adding new value to \gls{C-ITS}. A critical component of both \gls{AVs} and \gls{CAVs} is their perception capability, which involves using multiple sensors to recognize and interpret the driving environment, forming the foundation for subsequent planning and control operations. Perception tasks include 2D/3D object detection, semantic segmentation, object tracking, and motion prediction, among others. Driven by advances in artificial intelligence and multi-sensor fusion, the perception capabilities of individual vehicles have significantly improved. However, these capabilities are still limited by challenges such as visual occlusion and long-range detection, which are difficult to overcome with onboard sensors alone. These limitations can lead to reduced situational awareness, increase the risk of traffic accidents, and reduce the driving efficiency.

To address the limitations of individual vehicle perception, \gls{CP}\footnote{In the context of collaborative perception, the terms cooperative and collective perception are frequently used. However, in this paper, we specifically use the term Collaborative Perception to emphasize the dual aspects of information sharing and coordinated action among agents. In contrast, cooperative perception focuses on information sharing, while collective perception emphasizes the distributed nature of shared perception.} supported by \gls{V2V} and \gls{V2I} communication has gained significant attention \cite{FcooperFeatureBased-2019-chend}. In the context of \gls{CP}, where only vehicles and infrastructure are equipped with sensors, \gls{CP} utilizing both \gls{V2V} and \gls{V2I} is widely described as using \gls{V2X}. \footnote{We note that in communication technology, \gls{V2X} encompasses a broader scope, covering \gls{V2V}, \gls{V2I}, \gls{V2P} and \gls{V2N}.} As shown in Figure~\ref{fig:cp_schematic}, \gls{CP} allows for the sharing of sensor data between vehicles and infrastructure, thereby significantly extending the \acrfull{FOV} of individual vehicles to overcome challenges related to occlusion and long-range detection, which is critical for enhancing road safety and improving traffic efficiency across a wide range of use cases.

Initial investigations into \gls{CP} concentrated on the transmission of object-level information~\cite{Gunther2015} and aspects of the communication protocol design, such as message generation rules \cite{Thandavarayan_msg}, redundancy mitigation \cite{delooz_redundancy} and data congestion awareness~\cite{delooz_CP_DCCaware}, and culminated in the publication of communication standards in the standard development organizations (SDOs) ETSI~\cite{etsiCPM} and SAE~\cite{SAE_J3224}. As \gls{CP} evolved,  its scope broadened to include contributions from computer vision, with particular focus on the design of advanced perception algorithms and data fusion methods. Research has increasingly explored diverse data types for \gls{CP}, ranging from raw sensor data~\cite{JointPerceptionScheme-2022-ahmeda,RobustRealtimeMultivehicle-2023-zhangb,EdgeCooperNetworkAwareCooperative-2024-luoa}, intermediate neural features \cite{FcooperFeatureBased-2019-chend,BandwidthAdaptiveFeatureSharing-2020-marvastia,CooperativeLIDARObject-2020-marvastia}, to processed perception results \cite{DistributedDynamicMap-2021-zhanga,EnvironmentawareOptimizationTracktoTrack-2022-volka,CooperativePerceptionSystem-2023-songc}.

The different data types correspond to three principal paradigms of \gls{CP}: early cooperation, intermediate cooperation, and late cooperation. In early cooperation, network nodes exchange raw sensor data, which contain comprehensive environmental information but require substantial bandwidth for transmission. In contrast, late cooperation involves sharing processed perception results, which is the most bandwidth-efficient data format. However, this approach is vulnerable to errors introduced during earlier perception stages, such as sensor noise, object misclassification, and data synchronization issues, and is less resilient to pose inaccuracies~\cite{V2XViTVehicletoEverythingCooperative-2022-xub}. Intermediate cooperation is a viable solution to balance the trade-off between network bandwidth usage and accuracy. It requires less bandwidth than data-level fusion and is expected to offer higher accuracy than result-level fusion.

Each \gls{CP} method offers distinct advantages and disadvantages. Nonetheless, all types consistently outperform single-vehicle perception systems that lack collaboration. \gls{CP} has the potential to enhance perception accuracy and address blind spot issues. However, its practical implementation faces several significant challenges. Communication bandwidth is a significant constraint, restricting the amount of data that can be shared effectively \cite{When2comMultiAgentPerception-2020-liua,BandwidthAdaptiveFeatureSharing-2020-marvastia}. Localization errors further challenge data fusion by causing spatial misalignments \cite{LearningCommunicateCorrect-2021-vadivelub}, while time latency introduces temporal misalignments, undermining fusion accuracy~\cite{LatencyAwareCollaborativePerception-Lei-22}. Additionally, CP faces other critical challenges, including communication disruptions~\cite{InterruptionAwareCooperativePerception-2024-renc}, domain shifts \cite{DIV2XLearningDomainInvariant-2023-xiangb}, modality heterogeneity~\cite{HMViTHeteromodalVehicletoVehicle-2023-xiangb}, and susceptibility to adversarial attacks \cite{UsAdversariallyRobust-2023-lia}. Overcoming these barriers is crucial for scaling CP solutions and unlocking their full potential in advancing vehicular perception systems.

\subsection{Related Work} \label{intro_related_surveys}

\begin{table*}[h]
    \centering
    \caption{Summary of Surveys in Vehicular Collaborative Perception. Mod.: Modality, Co.: Collaborative type, OD: Object Detection, OT: Object Tracking, MP: Motion Prediction, SS: Semantic segmentation, LD: Lane detection, MT/TA: Multi-task.Task agnostic, LE: Localization error, TL: Time latency, CB: Communication bandwidth constraint, CI: Communication Interruption, DS: Domain Shift, Hetero.: Heterogeneous system, Adv.: Adversarial attack, DA: Dataset, ES: Evaluation scenarios, EM: Evaluation Metrics, AS: Ablation Study}
    \label{tab:survey_summary}
    \resizebox{\textwidth}{!}{%
    \begin{tabular}{l l l c c c c c c c c c c c c c c c c c c c c c}
        \toprule
        \multirow{2}{*}{Paper} & \multirow{2}{*}{Year} & \multirow{2}{*}{Publication} & \multirow{2}{*}{SLR} & 
        \multicolumn{8}{c}{The Taxonomy of Vehicular Collaborative Perception} & \multicolumn{7}{c}{Issues of Vehicular Collaborative Perception} & \multicolumn{4}{c}{Evaluation Method} \\
        \cmidrule(lr){5-12} \cmidrule(lr){13-19} \cmidrule(lr){20-23}
        & & & & Mod. & Co. & OD & OT & MP & SS & LD & MT/TA & LE & TL & CB & CI & DS & Hetero. & Adv. & DA & ES & EM & AS \\
        \midrule
        \cite{SurveyFrameworkCooperativePerceptionHeterogeneousSingletonHierarchicalCooperation-2022-bai} & 2022 & IEEE T-ITS & &  & $\checkmark$ & & & & & & & & & & & & & & $\checkmark$ & & & \\
        \cite{SurveyCooperativePerception-2022-caillotb} & 2022 & IEEE T-ITS & & & & $\checkmark$ & $\checkmark$ & & & & & & & & & & & & & & & \\
        \cite{CollaborativePerceptionAutonomous-2023-hanc} & 2023 & IEEE ITS & & & $\checkmark$ & & & & & & & $\checkmark$ & $\checkmark$ &  & $\checkmark$ &  & $\checkmark$ & $\checkmark$ & $\checkmark$ & & $\checkmark$ & \\
        \cite{VehicletoeverythingAutonomousDrivingSurveyCollaborativePerception-2023-liu} & 2023 & arXiv & & & $\checkmark$ & & & & & & & $\checkmark$ & $\checkmark$ & $\checkmark$ & & & & $\checkmark$ & $\checkmark$ & & & \\
        \cite{Huang2023V2XCP} & 2024 & arXiv & & & $\checkmark$ & & & & & & $\checkmark$ & $\checkmark$ & $\checkmark$ & $\checkmark$ & $\checkmark$ & $\checkmark$ & $\checkmark$ & $\checkmark$ & $\checkmark$ & & & \\
        \textbf{Ours} & 2024 & -- & $\checkmark$ & $\checkmark$ & $\checkmark$ & $\checkmark$ & $\checkmark$ & $\checkmark$ & $\checkmark$ & $\checkmark$ & $\checkmark$ & $\checkmark$ & $\checkmark$ & $\checkmark$ & $\checkmark$ & $\checkmark$ & $\checkmark$ & $\checkmark$ & $\checkmark$ & $\checkmark$ & $\checkmark$ & $\checkmark$ \\
        \bottomrule
    \end{tabular}%
    }
\end{table*}

\noindent Several narrative reviews on \gls{CP} have been published, each offering distinct perspectives on the field. For instance, Bai et al.~\cite{SurveyFrameworkCooperativePerceptionHeterogeneousSingletonHierarchicalCooperation-2022-bai} offer a high-level overview of the architecture and node structure of \gls{CP} systems, while Caillot~\cite{SurveyCooperativePerception-2022-caillotb} reviews \gls{CP}, with a focus on localization, object detection and tracking. In 2023, Han et al.~\cite{CollaborativePerceptionAutonomous-2023-hanc} explore \gls{CP} methods for both ideal scenarios and real-world applications, highlighting the gaps between current research and practical implementation. Liu et al.~\cite{VehicletoeverythingAutonomousDrivingSurveyCollaborativePerception-2023-liu} introduce issues of \gls{CP} while Huang et al. \cite{Huang2023V2XCP} propose a generic framework of \gls{CP}.

As summarized in Table \ref{tab:survey_summary}, all of these studies are narrative reviews and touch upon several aspects of \gls{CP}  but lack a transparent, comprehensive, and structured analysis of \gls{CP}, particularly from a computer vision perspective. They do not offer a detailed taxonomy of \gls{CP}  technologies or fully address the range of perception tasks that benefit from collaborative approaches. For instance, key tasks such as semantic segmentation, motion prediction, and lane detection remain unexamined in prior surveys. Additionally, the role of different sensing modalities in \gls{CP}  has not been systematically analyzed, leaving a critical gap in understanding camera-based \gls{CP}  or fusion-based \gls{CP} . Moreover, evaluation methodologies, which are essential for guiding the future development of \gls{CP}  technologies, are either absent or insufficiently discussed in previous reviews. This gap makes it difficult for readers to fully understand the range of \gls{CP} tasks and to quickly identify the specific focus of their own research within the field.

To address these shortcomings, this \gls{SLR} follow the the PRISMA
2020 guidelines and define five research questions as below:

\begin{itemize}
\item{{\bf{RQ1}}: How can collaborative perception be classified within a structured taxonomy?}
\item{{\bf{RQ2}}: Which methodological approaches are being used for evaluating collaborative perception?}
\item{{\bf{RQ3}}: Which scenarios are covered by evaluation approaches for collaborative perception?}
\item{{\bf{RQ4}}: Which metrics are used to measure the effectiveness of collaborative perception? }
\item{{\bf{RQ5}}: What are the challenges, opportunities, and risks of collaborative perception research?}
\end{itemize}

This \gls{SLR} selects relevant works based on predefined inclusion and exclusion criteria and extracts key data terms from the selected papers to address the research questions. Ultimately, this review evaluates the current state of \gls{CP} and highlights areas requiring further research.


\begin{figure*}[htb]
    \centering
    \includegraphics[width=\linewidth]{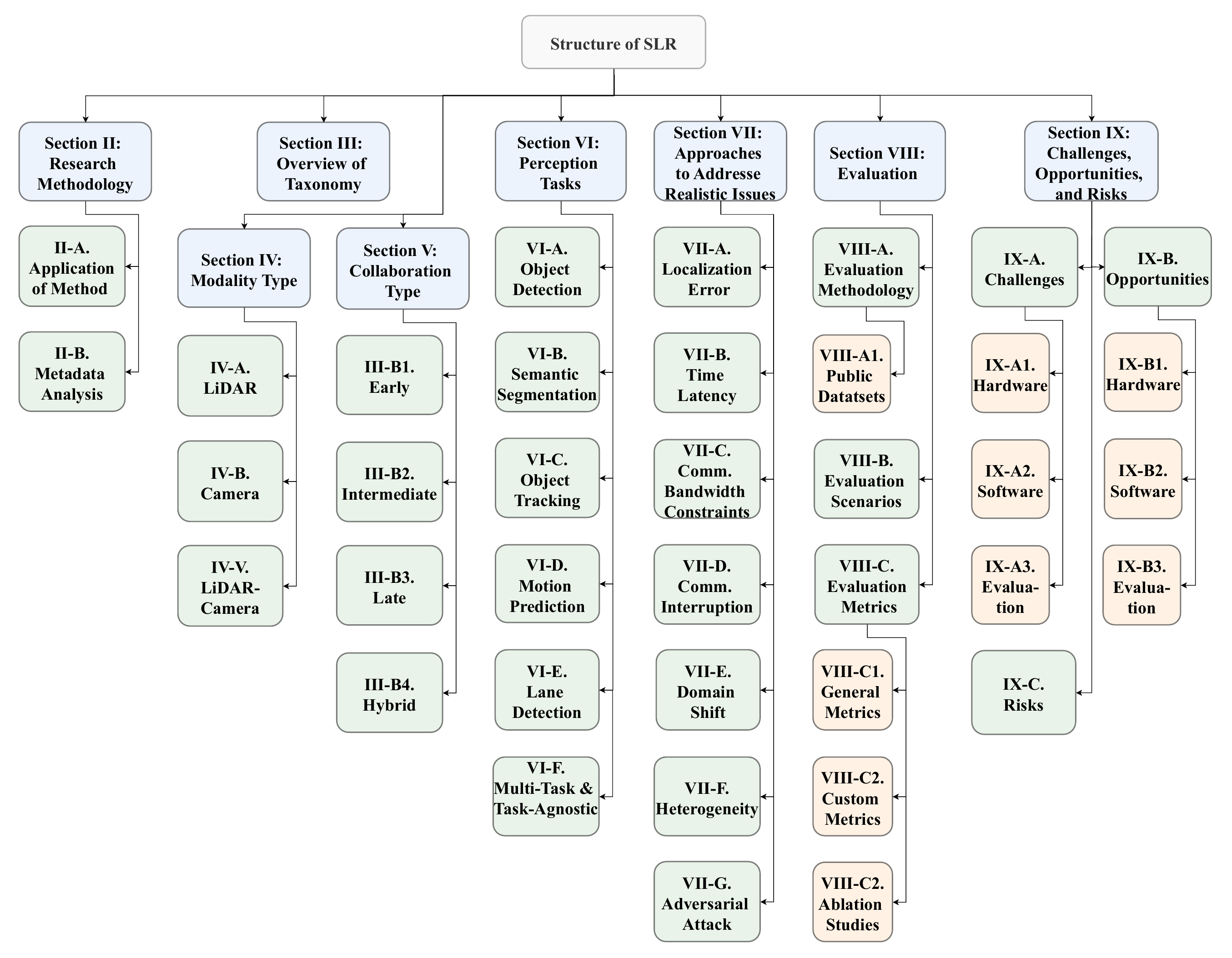}
    \caption{Organization of this \acrlong{SLR}.}
    \label{fig:organisation}
\end{figure*}

\subsection{Contributions} \label{intro_contri}
\noindent To minimize bias, enhance transparency, and ensure comprehensive coverage, we employ the methodology of \gls{SLR} and follow the PRISMA 2020 guidelines. This review examines 106 peer-reviewed papers that meet our selection criteria, offering a summary of existing research, and a comparative analysis of critical components in cooperative perception algorithms, highlighting remaining research gaps. The key contributions of this review are as follows:

\begin{itemize}
\item This systematic literature review distinguishes itself from existing narrative reviews by selecting relevant works in accordance with the PRISMA 2020 guidelines, ensuring transparency and reproducibility. At the conclusion of the study, five predefined research questions are addressed.

\item This review proposes a structured taxonomy for \acrlong{CP} technology, addressing the limitations of prior narrow classifications in existing surveys. The taxonomy categorizes solutions along modality, collaboration and task.
Furthermore, approaches to address real-world challenges in \gls{CP} for autonomous driving, such as localization errors, latency, communication issues, domain shifts, heterogeneous setups, and adversarial attacks, are systematically reviewed, categorized, and comparatively analyzed.

\item In contrast to the limited attention to evaluation methods in existing surveys, this review systematically examines the evaluation methodologies, performance metrics, and ablation studies employed in \gls{CP}. In particular, the \gls{CP} datasets are categorized and analyzed, distinguishing between synthetic and real-world datasets.

\item A comparative analysis is conducted to understand the advantages and disadvantages of different methods. Building upon this analysis, the study identifies future challenges, opportunities, and risks associated with \gls{CP} from various perspectives, including advancements in hardware and software for \gls{CP} and improvements in evaluation methods.

\end{itemize}

\subsection{Structure of Survey} \label{sec:intro_slr_structure}


The Sections~\ref{sec:taxonomy} to~\ref{sec:cp_issues}
address RQ1, beginning with an overview of a structured taxonomy in Section~\ref{sec:taxonomy}. Section~\ref{sec:modality} and~\ref{sec:collaboration-type} cover modality type and collaboration type, respectively, while Section~\ref{sec:perception_tasks} explores perception tasks addressed through multi-agent collaboration. Section \ref{sec:cp_issues} discusses the issues encountered in real-world applications and the existing solutions. Sections~\ref{sec:evaluation} addresses RQ2 to RQ4 and focuses on the evaluation methods of \gls{CP}, with particular emphasis on the available public datasets and evaluation metrics. Section~\ref{sec:challenges} addresses RQ5, highlighting the challenges, opportunities, and risks in CP research. Finally, Section~\ref{sec:conclusion} summarizes the findings of the review and provides conclusions. Figure~\ref{fig:organisation} provides a visual overview of the review's structure.

%% file: sections/02_research_methodology.tex
\section{Research Methodology}
\label{sec:research_methodolody}

\begin{figure}[!t]
\centering
\includegraphics[width=\linewidth]{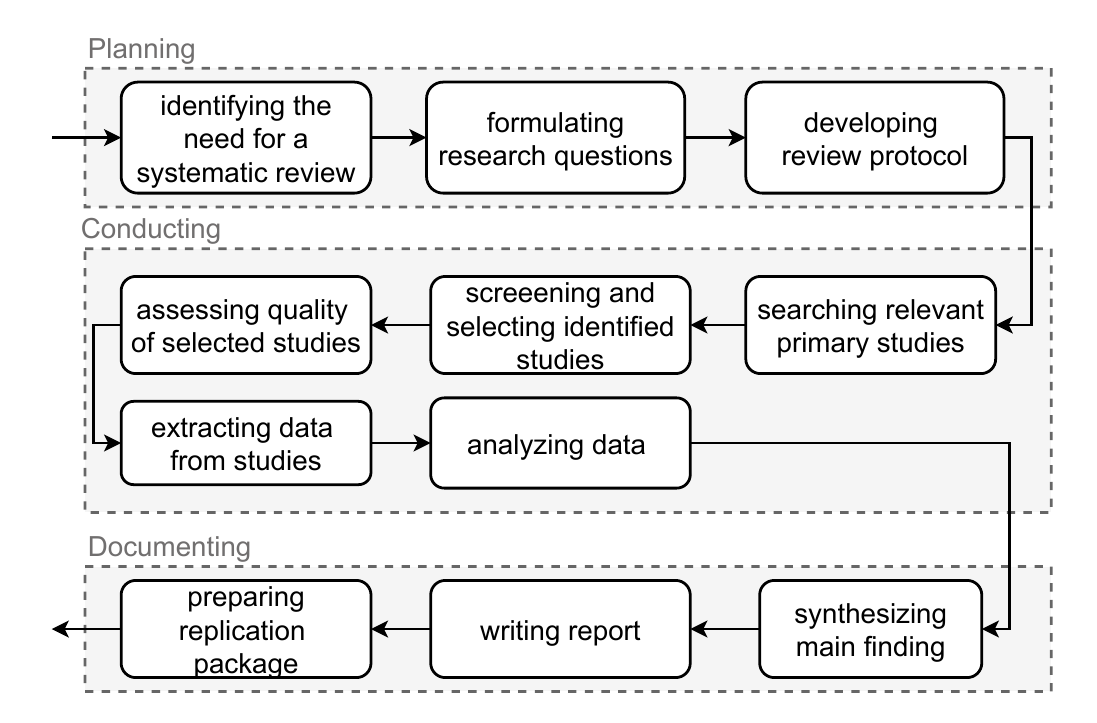}
\caption{The procedure of SLR in three stages: planning (review protocol development), conducting (screening and selection of articles), and documenting (synthesizing of findings).}
\label{fig:slr_procedure}
\end{figure}

\noindent A \acrfull{SLR} is a structured and methodical approach to reviewing and synthesizing existing research on a specific topic or research question. Unlike traditional narrative reviews, an \gls{SLR} follows a predefined protocol that includes a comprehensive search strategy, clear criterias for selecting studies, and rigorous methods for analyzing and synthesizing the findings. The aim is to minimize bias, ensure transparency, and provide a comprehensive overview of the current state of knowledge on the topic. Our research process is following the guideline of the PRISMA~2020 statement~\cite{Pagen71} and the methodology presented in Kitchenham et al.~\cite{KITCHENHAM20132049}, which serves as a transparent and uniform systematic review framework. Fig.~\ref{fig:slr_procedure} illustrates the general procedure of a \gls{SLR}, which consists of three phases: Planning, Conducting, and Documenting. Additionally, the primary reviewers have diverse backgrounds in AI, computer vision, robotics, human-robot interaction, and communication, ensuring a broad range of perspectives in the review process. The application of the method to \gls{CP} literature will be discussed in Section~\ref{research_methodolody_application}, while the metadata analysis will be described in Section~\ref{research_methodolody_meta}.

\subsection{Application of the SLR Method to Cooperative Perception Literature} \label{research_methodolody_application}
\noindent This section will outline the practical application of the above described process. The subsequent subsections will provide a detailed explanation of each step involved in the procedure.

\subsubsection{Definition of Review Protocol}
The review protocol establishes the methodological framework for this study and comprises four key components: search strategy, selection criteria, data extraction strategy, and quality assurance strategy. The search strategy specifies the approach for systematically identifying relevant literature, while the selection criteria outline the criteria for including or excluding studies. The quality assurance strategy ensures the reliability of the review by evaluating the quality of the included studies.

\begin{table}[htbp]
\centering
\caption{Search resources and Search String.}
\label{tab1}
\begin{tabular}{p{0.9\linewidth}}
\toprule
\textbf{Databases and Search Engine} \\ \midrule
IEEE Xplore, ACM Library, ScienceDirect, MDPI, Scopus, Google Scholar \\
\toprule
\textbf{Search String} \\ \midrule
(collaborative OR collective OR cooperative OR multi-agent) AND perception AND (V2X OR V2V OR V2I) \\ 
\bottomrule
\end{tabular}
\end{table}

\begin{table}[ht]
\centering
\caption{Selection Criteria}
\label{tab2}
\begin{tabular}{p{0.9\linewidth}}
\toprule
\textbf{Inclusion criteria} \\ \midrule

\begin{itemize}
    \item [1)] Primary studies that provide an explicit research character.  
    \item [2)] Studies published from 2019 to 2024 (March). 
    \item [3)] Studies that are classified as academic articles from conferences or journals and pre-print versions of articles that were clearly accepted by conferences or journals. 
    \item [4)] Studies that address (or evaluate) perception that is derived as joint effort between different entities.  
\end{itemize} \\ 

\toprule
\textbf{Exclusion criteria} \\ \midrule
\begin{itemize}
    \item [1)] Studies written in any language other than the English language.
    \item [2)] Grey literature that are preprints, blog posts, websites, newsletters, white papers, government documents, RSS feed, videos, podcasts and webinars, except the preprint versions of accepted papers by conferences and journals.
    \item [3)] Studies that are not available, and hence not analyzable (e.g., the full text of a scientific article is not accessible). 
    \item [4)] Duplicates of already included studies.  
    \item [5)] Studies that only address the single-entity perception from vehicle perspective or infrastructure perspective. 
    \item [6)] Studies that do not provide details about the fusion of perception data from different entities or the evaluation of perception results in road environments. 
    \item [7)] Studies that only focus on the communication  protocol design.
\end{itemize} \\ 
\bottomrule
\end{tabular}
\end{table}

\begin{itemize}
\item{{\bf{Search Strategy}}: The search strategy encompasses the selection of resources to be searched, the formulation of the search string, and the execution of the search procedure. In this study, several databases and one search engine were chosen as resources, as illustrated in Table~\ref{tab1}. The search string, provided in Table~\ref{tab1}, was applied across these databases and the search engine to gather relevant literature. To refine the search, appropriate filters were utilized for each resource. The main steps of the search procedure include: collecting relevant papers from each resource up to a defined upper limit (1,000 items), removing duplicates, applying the selection criteria to the collected papers, performing forward and backward snowballing \footnote{Snowballing is a technique for expanding a literature search by reviewing the references of selected papers (backward snowballing) and identifying papers that cite them (forward snowballing).} on the paper set, and finally, reapplying the selection criteria.}

\item{{\bf{Selection Criteria}}: The selection criteria include both inclusion and exclusion criteria, as detailed in Table \ref{tab2}. These criteria narrow the scope of the review to peer-reviewed academic articles published within the last five years, ensuring that the final set of papers is of high quality. Therefore, preprint papers without peer review are not included to ensure that the collected papers meet established academic standards.} Specifically, exclusion criterion 6, which pertains to the level of evaluation detail, further reinforces the quality of the selected articles. The criteria are also designed to maintain a specific focus on cooperative perception techniques, explicitly excluding studies on roadside perception or ego vehicle perception. An article is included in the final set only if it satisfies all the inclusion criteria and does not meet any of the exclusion criteria.

\item{{\bf{Data extraction strategy}}: The data extraction aims to gather all relevant information necessary to address the predefined research questions. Prior to commencing the process, the specific data term to be extracted from the articles will be clearly defined and formulated. Once the final paper set is determined, the extraction strategy will be reviewed and refined to ensure both comprehensiveness and the availability of the required data. The extracted data terms are detailed in Table \ref{tab3}, \ref{tab4} and \ref{tab5}, respectively.}

\item{{\bf{Quality assurance}}: The quality assurance process is designed to mitigate potential biases introduced by individual researchers by implementing multi-round reviews, cross-validation, and establishing consensus on key principles. The detailed quality assurance plan is outlined in Table~\ref{tab6}.}

\end{itemize}

\begin{table}[ht]
\centering
\caption{Data extraction term corresponding to research questions RQ1.}
\label{tab3}
\begin{tabular}{l|p{5cm}}
\toprule
\multicolumn{2}{c}{\textbf{RQ1: Taxonomy of \gls{CP}}} \\ \midrule
\textbf{Taxonomy} & Perception task, Modality/Sensor, Collaboration type, Entity type \\ \midrule
\textbf{Fusion Mechanisms} & Shared information, Information fusion mechanisms, Temporal alignment mechanisms, Spatial alignment mechanisms\\ \midrule
\textbf{Repository} & Repository accessibility \\ 
\bottomrule
\end{tabular}
\end{table}

\begin{table}[ht]
\centering
\caption{Data extraction term corresponding to research questions RQ2-4.}
\label{tab4}
\begin{tabular}{l|p{5cm}}
\toprule
\multicolumn{2}{c}{\textbf{RQ2-4: Evaluation of \gls{CP}}} \\ \midrule
\textbf{Methodology} & Evaluation approach, Dataset type, Real-world experiment setup, Simulation platforms and Tools \\ \hline

\textbf{Datasets} & Supported \gls{CP} tasks, V2X type, Number of CAVs, Sensor layout, Annotation, Localization of vehicle, Synchronization, Number of annotated frames, Maps, Location \\ \midrule

\textbf{Scenarios} & Environment type, Road type, Traffic scenarios, Weather, Time of the day, Visual occlusion, Accident \\ \midrule

\textbf{Metrics} & General metrics, Specific metrics, Ablation studies \\

\bottomrule
\end{tabular}
\end{table}

\begin{table}[ht]
\centering
\caption{Data extraction term corresponding to research questions RQ5.}
\label{tab5}
\begin{tabular}{l|p{5cm}}
\hline
\multicolumn{2}{c}{\textbf{RQ5: Challenges, opportunities, and risks of \gls{CP} research}} \\ \toprule
\textbf{Study} & Objectivies of the study, Contributions of the study, Main findings of the study, The limitation of the proposed approach, The future work of the study \\ 
\bottomrule
\end{tabular}
\end{table}

\begin{table}[ht]
\centering
\caption{Quality assurance plan.}
\label{tab6}
\begin{tabular}{p{0.9\linewidth}}
\hline
\textbf{Definition of review protocol} \\ \hline
\begin{itemize}
    \item [1)] The first author defines the review protocol, including definition of research questions, search strategy, selection criteria, data extraction strategy.  
    \item [2)] The other authors review the review protocol  
    \item [3)] Disagreements will be discussed until the consensus is reached  
\end{itemize} \\ \hline
\textbf{Random assessment of included/excluded publications and extracted data} \\ \hline
\begin{itemize}
    \item [1)] The first author conducts the selection/extraction process on the entire set. 
    \item [2)] The second author conducts the selection/extraction on a sampled subset (randomly 10\%, based on the amounts of papers). 
    \item [3)] The outcomes of the selection/extraction on the subset are compared, and any disagreements are forwarded to the third author for discussion among the three until a consensus is achieved. 
    \item [4)] If the percentage of incorrectly excluded articles exceeds 10\%, then it is necessary for the first author to re-examine all results considering the new consensus and return to the second step 
\end{itemize} \\ \hline
\end{tabular}
\end{table}

\begin{figure}[t]
    \centering
    \setlength{\fboxrule}{0.5pt}
    \fcolorbox{lightgray}{white}{
    \includegraphics[width=0.8\linewidth]{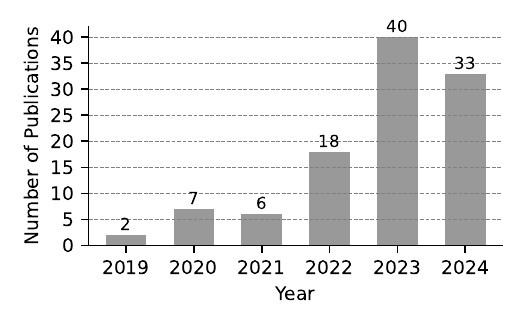}}
    \caption{Number of publications over the past five years}
    \label{fig:dist_year}
\end{figure}

\subsubsection{Search and Selection}
\begin{figure}[t]
    \centering
    \includegraphics[width=\linewidth]{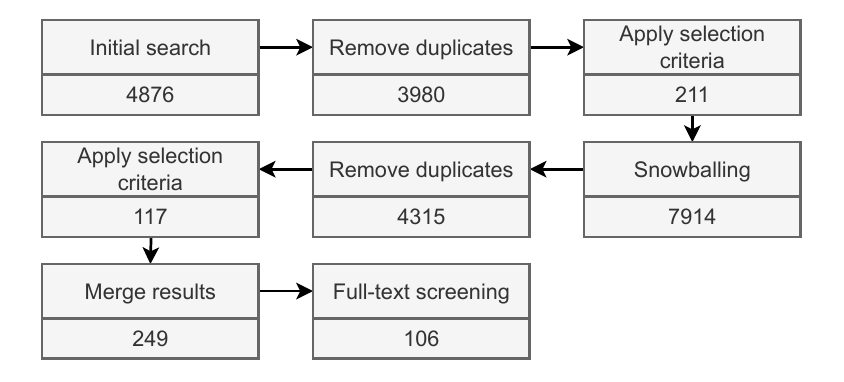}
    \caption{Procedure of the search and selection, starting from 4876 items, reduced to 3980 after duplicates were removed, 249 after screening and snowballing, and resulting in 106 studies included in the final review.}
    \label{fig:selection_procesure}
\end{figure}

By applying the search strategy, 3,980 articles were identified after duplicate removal. The subsequent selection process involved applying the inclusion and exclusion criteria to the titles, abstracts, conclusions, and overall structure of the articles, which narrowed the paper set to 211 articles. Forward and backward snowballing techniques were then conducted on this set to ensure that no relevant articles beyond the initial search were overlooked. The selection criteria were also applied to any articles identified through snowballing. To further validate the selection, the criteria were applied to the full text of all remaining articles. This comprehensive and rigorous process ultimately resulted in a final set of 106 articles. The detailed procedure is outlined in Figure \ref{fig:selection_procesure}.

\subsubsection{Data Extraction and Analysis}
The data extraction strategy was initially reviewed and then systematically applied to all selected articles. The extracted data were subsequently clustered, examined, summarized, and analyzed. Both quantitative and qualitative analyses were conducted to address the research questions. These analyses enabled a clear identification of the current state of research, existing gaps, and future research trends.

\subsection{Metadata Analysis} \label{research_methodolody_meta}
\noindent This section presents the metadata analysis conducted to identify research trends in cooperative perception. Figure~\ref{fig:dist_year} visualizes the number of publications from 2019 to 2024, showing a steady increase, with a pronounced surge in 2023. This trajectory reflects both the maturation of foundational research and the rapid expansion of real-world applications, supported by significant funding initiatives in intelligent transportation and autonomous driving. It also highlights that cooperative perception (\gls{CP}) is transitioning from an emerging topic to a consolidated research domain. Nevertheless, due to indexing delays (e.g., IEEE Xplore updates) and the cutoff date for data collection in March 2024, the actual number of recent publications is likely higher than reported, suggesting that this upward trend is even stronger than captured here.

\begin{table*}[t]
\centering
\caption{Summary of surveyed collaborative perception studies across multiple dimensions.}
\label{tab:cp_summary} 
\begin{tabular}{p{2.05cm}p{7cm}p{7cm}}
\toprule
\textbf{Dimension} & \textbf{Categories (with counts)} & \textbf{Key observations} \\ 
\midrule
Region & Asia (54), North America (38), Europe (13), Africa (1) 
& Research is predominantly conducted in Asia and North America. \\ \midrule
Publication venue (top 10) & ICRA (16), CVPR (8), IEEE T-IV (8), NeurIPS (8), ICCV (5), 
IEEE RA-L (5), IEEE ITSC (5), IEEE T-ITS (4), IEEE IoTJ (4), IEEE IV (4) 
& Publications are concentrated in leading robotics and vision venues, with fewer in transportation-focused outlets. \\ \midrule
Modality & LiDAR (63), Camera (13), LiDAR–Camera (12) 
& LiDAR-based approaches remain predominant, reflecting their reliability for accurate 3D detection. \\ \midrule
Collaboration & Early (6), Intermediate (71), Late (15), Hybrid (6) 
& Intermediate fusion is the predominant strategy, balancing bandwidth efficiency with information richness. \\ \midrule
Task & Object detection (78), Semantic segmentation (6), Object tracking (5), 
Motion prediction (3), Lane detection (2), Multi-task (3), Task-agnostic (8) 
& Object detection dominates the field, while other tasks remain comparatively underexplored. \\
\bottomrule
\end{tabular}
\end{table*}

\textbf{Regional distribution.} Table~\ref{tab:cp_summary} shows that Asia (54) and North America (38) dominate the research landscape. This concentration reflects the substantial investment in \gls{V2X} testbeds, 5G infrastructure, and large-scale smart mobility projects in countries such as China and the United States. Europe, while contributing fewer studies (13), plays an important role in standardization and cross-border research initiatives (e.g., ETSI standards for cooperative ITS). By contrast, contributions from other regions, including Africa (1), are minimal, underscoring the geographical imbalance of current \gls{CP} research and highlighting opportunities for more globally distributed investigations.

\textbf{Publication venues.} The majority of papers are published in premier robotics and computer vision venues, including ICRA (16), and CVPR (8) as summarized in Table~\ref{tab:cp_summary}. This distribution indicates that \gls{CP} has expanded beyond its original roots in communication and networking, and is increasingly recognized as a computer vision and AI challenge. The growing presence in CVPR, NeurIPS, and ICCV emphasizes the centrality of deep learning and visual perception methods, while robotics-oriented outlets such as ICRA and IEEE RA-L demonstrate the integration of \gls{CP} into embodied autonomous systems. The shift of publication venues therefore illustrates both methodological diversification and the convergence of perception, AI, and robotics communities around \gls{CP}.

\textbf{Modalities.} LiDAR-based approaches account for the majority of studies (63), with only 13 camera-only and 12 LiDAR–Camera fusion papers (Table~\ref{tab:cp_summary}). LiDAR’s predominance stems from its robustness in capturing precise 3D spatial geometry, which is critical for reliable detection in cluttered and occluded environments. Camera-based methods, by contrast, face inherent challenges in depth estimation and performance degradation under illumination changes. LiDAR–Camera fusion remains underrepresented despite its potential to combine complementary strengths (texture-rich visual data and precise depth). This limited adoption reflects the technical challenges of spatial–temporal calibration and multimodal fusion complexity. Moreover, the near absence of alternative modalities (e.g., radar, thermal, event cameras) underscores a critical research gap and signals promising directions for future multimodal \gls{CP}.

\textbf{Collaboration strategies.} Table~\ref{tab:cp_summary} further reveals that intermediate collaboration dominates (71 studies), while early (6), late (15), and hybrid (6) approaches are far less common. The predominance of intermediate fusion reflects its ability to balance bandwidth efficiency with perception accuracy by exchanging processed features rather than raw data or final outputs. The limited adoption of other schemes underscores the persistent technical barriers: early fusion faces prohibitive communication demands, late fusion suffers from information loss, and hybrid designs increase synchronization and integration complexity. These findings suggest that while intermediate collaboration is currently the most practical solution, advancing adaptive and flexible collaboration schemes will be crucial for future large-scale deployments.

\textbf{Tasks.} Finally, research efforts are heavily skewed toward object detection (78 studies), with relatively few works on semantic segmentation (6), object tracking (5), and motion prediction (3), as shown in Table~\ref{tab:cp_summary}. The focus on object detection underscores its central role as a prerequisite for higher-level reasoning, yet the limited attention to other tasks reveals important gaps. In particular, tracking and prediction are essential for safety-critical decision-making, and their underrepresentation highlights opportunities for future work. Similarly, multi-task and task-agnostic designs could support more integrated perception pipelines, but remain at an early stage of development.

Overall, this metadata analysis demonstrates that while cooperative perception has become a rapidly growing and increasingly multidisciplinary field, current research exhibits clear imbalances across regions, modalities, collaboration schemes, and perception tasks. These insights provide a foundation for identifying gaps and charting future research directions.



%% file: sections/03_taxonomy.tex
\section{Overview of Structured taxonomy (RQ1)} \label{sec:taxonomy}

\noindent \acrlong{CP} is a complex field of study with numerous subsets of sensors, collaboration methodologies and tasks. In this survey, we propose a taxonomy to classify the multitude of solutions available. We define the taxonomy based on the modality (sensor type), collaboration type, and perception task. Through the \gls{SLR}, we have identified a strong focus on two types of sensor, LiDAR and camera. While the usage of LiDAR as the data source is more abundant, there is also a significant presence of cameras and the combination of LiDAR and cameras in the surveyed work. Therefore, as it relates to the modality, we classify the work into LiDAR, Camera, or a combination of LiDAR-Camera.

\gls{CP} can be further classified by the collaboration type. Based on the level of the underlying data fusion algorithm, we classify work into Early, Intermediate, Late, and Hybrid Collaboration. Early, Intermediate, and Late Collaboration are self-explanatory as the data fusion inputs are shared among participants. Hybrid Collaboration refers to solutions that share data across multiple fusion levels. Furthermore, the subcategories within intermediate collaboration are outlined as follows: traditional feature fusion, attention-based feature fusion, and graph-based feature fusion.

In addition, we have identified several specific \gls{CP} tasks that further classify the solutions, including object detection, object tracking, motion prediction, semantic segmentation, lane detection, multi-task approaches, and task-agnostic methods.

We further analyze the the approaches used in the surveyed studies to address realistic issues. These issues are categorized as localization errors, time latency, communication bandwidth constraints, communication interruptions, domain shifts, heterogeneity, and adversarial attacks. For each issue, we provide the corresponding categories of approaches employed to address them.

Due to the challenges associated with conducting a fair experimental comparison, such as the lack of publicly available source code for many methods, this review primarily adopts a qualitative analysis approach.

\section{Modality Type (RQ1)} \label{sec:modality}
\noindent  In this section, we provide an in-depth examination of the different modalities in \gls{CP}. Our systematic review identifies three primary modalities in the reviewed literature: LiDAR, camera, and their combination.

\subsection{LiDAR}

LiDAR is an acronym for Light Detection and Ranging. It describes a class of sensors that determine ranges by targeting an object or surface with a laser and measuring the time for the reflected light to arrive at the receiver. The sensor used in vehicular perception performs a multi-point scan across the environment at high frequencies to accurately measure the distance from the sensor to objects. 
The \emph{channels} of a LiDAR sensor refer to the number of distinct laser beams emitted. It affects its resolution and field of view. For example, compared to a 16-channel LiDAR system, a 128-channel one captures more vertical slices of the surrounding environment.
By varying the number of channels and their configurations, LiDAR can achieve different resolutions, ranges, and levels of detail, suitable for various applications in perception.

Just as deep neural networks can extract features from images, they can also be used to extract features from LiDAR data. One intuitive method is point-based feature extraction: process the raw data and generate a sparse representation, aggregate the features of adjacent points, and extract the feature of each point. However, this method poses stringent hardware requirements and is not seen in our surveyed work. Currently, the main feature-extraction approaches are voxel-based and pillar-based.

Voxel-based methods first convert point clouds into a structured, regular grid of 3D cells called voxels. By dividing the 3D space into voxels, the network can leverage 3D or 2D convolutional neural networks for feature extraction, making detecting objects more efficient and structured.
The VoxelNet\cite{zhou2018voxelnet} is frequently used \cite{FcooperFeatureBased-2019-chend, CoFFCooperativeSpatial-2021-guoa, SlimFCPLightweightFeatureBasedCooperative-2022-guoa, V2VFormerVehicletoVehicleCooperative-2024-lina}, 
often with sparsely embedded convolutional layers applied to 3D voxel features to improve the efficiency of object detection
\cite{DistributedDynamicMap-2021-zhanga, SoftActorCriticBased-2022-xiea, CooperativePerceptionLearningBased-2023-liub}.

The effort to improve the backbone feature extractor network is still ongoing. Besides VoxelNet, different network architectures are also proposed\cite{DynamicFeatureSharing-2023-baia, CollectivePVRCNNNovel-2023-teufela}. Chen et al. \cite{HP3DV2VHighPrecision3D-2024-chena} propose to improve the LiDAR data feature extraction backbone. They construct voxel pillars on voxel feature maps and encode them to generate \gls{BEV} features, thereby addressing the issue of spatial feature interaction lacking in PointPillars \cite{lang2019pointpillars} methods and enhancing the semantic information of extracted features. A maximum pooling technique reduces dimensionality and generates pseudo images, skipping complex 3D convolutional computation. In the work of Ma et al. \cite{MACPEfficientModel-2024-maa}, each vehicle encodes point cloud features locally using a new feature encoder network with a module called ConAda.

The pillar-based method offers advantages in real-time performance due to its efficient handling of {3D} point cloud data. The pillar representation disregards partitioning along Z-axis and divides the 3D space into fixed size pillars. Intuitively the pillar is seen as an unbound voxel along the \mbox{Z-axis}. Pillar-based features are extracted through Deep Learning models inspired by PointNet~\cite{qi2017pointnet}. Since pillars are not partitioned along \mbox{Z-axis}, a pillar-based representation of a point cloud is seen as a \gls{BEV} image of multiple channels.

The pillar-based feature extractor often applied DNN on the BEV-form or raw LiDAR data. In the early phase of using this approach for LiDAR data in CP (about from 2020 to 2023), several different networks are proposed. Marvasti et al. propose such a network structure~\cite{BandwidthAdaptiveFeatureSharing-2020-marvastia, CooperativeLIDARObject-2020-marvastia}.
Luo et al. \cite{ComplementarityEnhancedRedundancyMinimizedCollaboration-2022-luob} adopt the MotionNet, quantizing the 3D points into regular voxels and representing the 3D voxel lattice as a 2D pseudo-image, with the height dimension corresponding to image channels. Qiao et al. \cite{AdaptiveFeatureFusion-2023-qiaob} use PointNet instead. The DiscoNet proposed by Li et al. \cite{LearningDistilledCollaboration-2021-lib} is later used by others \cite{RobustCollaborativePerception--rena, LatencyAwareCollaborativePerception-Lei-22}.

The most representative module for pillar-based feature extraction is PointPillars~\cite{lang2019pointpillars}. 
It employs a simplified version of PointNet within each pillar to extract features from the points. The point-wise features are then aggregated to create a single feature vector for each pillar. These pillar features are organized into a 2D grid, allowing leveraging 2D CNN for feature extraction.

PointPillars is widely used for LiDAR data feature extraction \cite{FTransformerPointCloud-2022-wanga, V2XViTVehicletoEverythingCooperative-2022-xub, AsynchronyRobustCollaborativePerception--weic, PillarGridDeepLearningBased-2022-baia, Collaborative3DObject-2023-wanga, FeaCoReachingRobust-2023-guc, How2commCommunicationEfficientCollaborationPragmatic--yanga, What2commCommunicationefficientCollaborative-2023-yange, LearningVehicletoVehicleCooperative-2023-lia, RobustCollaborative3D-2023-lub, RobustRealtimeMultivehicle-2023-zhangb, SpatioTemporalDomainAwareness-2023-yangb, DUSADecoupledUnsupervised-2023-kongc, EdgeCooperNetworkAwareCooperative-2024-luoa, HPLViTUnifiedPerception-2023-liuc, CenterCoopCenterBasedFeature-2024-zhou, InterruptionAwareCooperativePerception-2024-renc, PracticalCollaborativePerception-2024-daoa, RobustCollaborativePerception-2024-lei, S2RViTMultiAgentCooperative-2024-lib, Select2ColLeveragingSpatialTemporal-2024-liuc, RegionBasedHybridCollaborative-2024-liu}.
Some studies build on the PointPillars framework by developing structurally similar models that adapt its core principles without directly replicating it~\cite{PillarBasedCooperativePerception-2022-wanga,MKDCooperCooperative3D-2024-lia}.
Wang et al.~\cite{PAFNetPillarAttention-2024-wanga} retain the PointPillars architecture but enhance it by replacing the 2D backbone with a four-layer residual network and adding a spatial pyramid pooling module. This enhancements expand the model's input area and enable it to combine information from multiple scales.

Some of the most recent research efforts try to improve the feature extraction mechanism. 
Instead of using the standard backbones, Bai et al.~\cite{PillarAttentionEncoder-2024-bai} propose a new adaptive feature encoder named Pillar Attention Encoder, which extracts the feature data based on the attention mechanism and adaptively reduces the data amount for sharing based on the exact communication bandwidth.

\subsection{Camera}

Cameras are among the most widely utilized modalities in perception systems, valued for their ability to capture high-resolution visual data containing dense semantic information, which is essential for tasks such as object detection, lane detection, and scene understanding. Monocular and multi-view camera setups are the two most common configurations employed in visual perception systems. Camera-only 3D perception provides an economical alternative to LiDAR-based systems. However, accurately estimating depth remains challenging due to the lack of direct 3D measurements. Similarly, camera-only CP remains relatively under-explored, encountering challenges similar to those in single-vehicle camera-only 3D perception. Due to that there are a limited number of camera-based papers, we will introduce them separately in this part.

Hu et al. \cite{CollaborationHelpsCamera-2023-hub} introduce CoCa3D, the camera-only 3D detection improved by introducing multi-agent collaborations, while many previous work focus on network designs. The proposed CoCa3D method first enhances image-based single-agent depth estimation before the Collaborative detection feature learning module that enhances 3D detection. In the later phase, the \gls{BEV} features that may contain the most informative cues are exchanged and fused to get a better \gls{BEV} feature map.  

Huang et al. \cite{ActFormerScalableCollaborative-2024-huangc} aims to achieve scalable camera-based collaborative perception with a Transformer-based architecture. The image information of the vehicles is projected into features using a \gls{BEV} encoder backbone such as BEVFormer. The transformer is trained to take the \gls{BEV} feature of the ego-vehicle and the poses of a collaborator and its cameras as input, and it chooses which part of the collaborator's feature map is important and should be transmitted. 

Wang et al.~\cite{EMIFFEnhancedMultiscale-2024-wangc} propose to address the information loss and pose errors due to time asynchrony across cameras in image-based fusion. Thus, it proposes a new fusion network architecture. It contains an attention and channel masking mechanism to enhance infrastructure and vehicle features at scale, spatial, and channel levels to correct the pose error introduced by camera asynchrony. It also uses feature compression to improve transmission efficiency. The proposed structure uses ResNet-50 as a backbone and FPN as a 2D neck to extract image features. Its evaluation is based on the \mbox{DAIR-V2X} dataset.

Fan et al.~\cite{QUESTQueryStream-2023-fand} propose the query cooperation paradigm for cooperative perception tasks, which is more interpretable than scene-level feature cooperation. They then propose the transformer-based QUEST framework  utilizing VoVNetV2 \cite{lee2019energy} as the feature encoding backbone. Every query output from the decoder corresponds to a possible detected object, and the query will be shared if its confidence score meets the request agent's requirements. As the cross-agent queries arrive, they are utilized for query fusion and implementation. 

\begin{figure*}[ht]
    \centering
    \includegraphics[width=\linewidth]{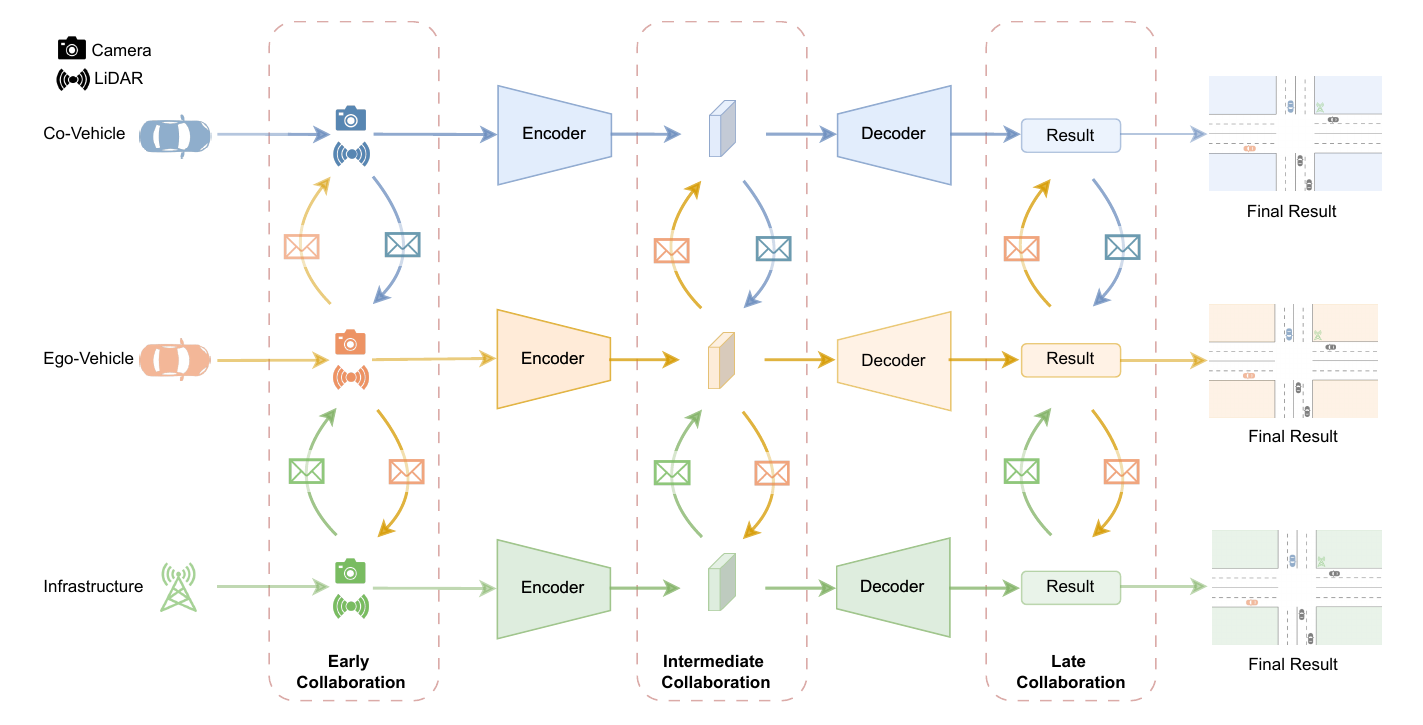}
    \caption{Illustration of collaboration type in \gls{CP}, showcasing Early Collaboration, Intermediate Collaboration, and Late Collaboration across Co-Vehicle, Ego-Vehicle, and Infrastructure, integrating Camera and LiDAR data. Early Collaboration involves sharing raw sensor data (e.g., images, LiDAR point clouds) for joint processing, while Intermediate Collaboration transmits extracted features (e.g., key points, feature maps). Late Collaboration shares final perception results (e.g., detected objects, trajectories).}
    \label{fig:cp_collaboration_type}
\end{figure*}

\subsection{LiDAR-Camera}

Most work on CP that utilizes both LiDAR and Camera sensors follows a simple paradigm. The proposed structure can use either LiDAR or Camera data as inputs because both types of sensor data will be turned into the same type of \gls{BEV} feature maps as a uniform intermediate representation for later processing. The work of Yin et al.~\cite{V2VFormerMultiModalVehicletoVehicle-2024-yina} is a typical example. It proposes V2VFormer++, where individual camera-LiDAR representation is incorporated with dynamic channel fusion (DCF) at \gls{BEV} space, and ego-centric \gls{BEV} maps from adjacent vehicles are aggregated by a global-local transformer module. The camera images are first cropped with a resolution of 520×520 pixels, fed into the ResNet-34 encoder for multi-scale feature extraction, and then processed by a sparse cross-attention view Transformer module.
PointPillars first processes the single-vehicle LiDAR data for point feature extraction, and a simple PointNet architecture is used for pillar feature extraction. Finally, a 2D CNN backbone is introduced to merge multi-resolution maps into a dense LiDAR \gls{BEV} feature. Many other work follow the same pattern
\cite{Where2commCommunicationEfficientCollaborative-2022-huc,MultimodalCooperative3D-2023-chia,HMViTHeteromodalVehicletoVehicle-2023-xiangb,ExtensibleFrameworkOpen-2024-luc}. These work may vary slightly in the backbone used, especially for processing camera data. For example, Zhou et al.~\cite{ViTFuseNetMultimodalFusion-2024-zhoua}  uses the Fast-SCNN network as the image feature map encoder, while some may use BEVFormer. Zhang et al. \cite{V2VFusionMultimodalFusion-2023-zhanga} provide a slightly different scenario where each agent is equipped with LiDAR and camera sensors. The work of Zhang et al.~\cite{MultiModalVirtualRealFusion-2022-zhanga} fuse LiDAR and RGB data through point cloud fusion, first converting RGB images into virtual point clouds and then combining them with real point clouds.

\subsection{Comparative Analysis}

One noticeable observation is that while there is extensive research on LiDAR, camera-based CP has only recently emerged, with relatively few papers exploring the use of cameras as a data source.
The reason could be the lack of depth perception of cameras, sensitivity to lighting and weather conditions, and heavy computational requirements in semantic understanding. Additionally, visual data from cameras raises privacy concerns under data protection laws, further impacting the deployment of camera-based systems. However, despite such differences, we can still observe that the problem to solve in both cases are similar, such as the limited bandwidth, lossy communication, temporal- and spatial-asynchrony, sensor and model heterogeneity, etc. They are still being actively investigated regardless of sensor types. 
Thus, the improvement in one area can also have an impact on the other. 
Besides, one trend we can observe is that research tends to use existing backbones and datasets, gradually convergent to a limited number of choices.

\section{Collaboration type (RQ1)} \label{sec:collaboration-type}
\noindent This section presents an in-depth review of the various collaboration types in \gls{CP}: Early, Intermediate, Late, and Hybrid.

\subsection{Early Collaboration}

\noindent In collaborative perception, early collaboration refers to the approach where raw sensor data (such as camera images, or LiDAR point clouds) from multiple vehicles are shared and fused early in the processing pipeline. This is done before any significant local processing or feature extraction is applied to the data. The fused data is then processed collectively to generate a unified perception of the environment. It allows for richer information exchange, as the original details in the sensor data are preserved. On the other hand, sharing raw sensor data, such as high-resolution camera images or dense LiDAR point clouds, requires significant communication bandwidth. Some examples of this approach exist where raw LiDAR data is shared among vehicles \cite{JointPerceptionScheme-2022-ahmeda, EdgeCooperNetworkAwareCooperative-2024-luoa, FastClusteringCooperative-2024-kuanga, RobustRealtimeMultivehicle-2023-zhangb} and one where infrastructure also participates \cite{EdgeCooperNetworkAwareCooperative-2024-luoa}. \cite{ViTFuseNetMultimodalFusion-2024-zhoua} differs from the others in that it also enables the sharing of raw camera data.

\subsection{Intermediate Collaboration}
\noindent In intermediate collaboration, neural network-generated features are distributed and merged to improve perception performance and conserve bandwidth. Based on their fusion mechanisms, these methods are categorized into three types: traditional feature fusion, attention-based feature fusion, and graph-based feature fusion. This section presents a comparative analysis of these intermediate fusion approaches.

\subsubsection{Traditional Feature Fusion}
Non-parametric operators such as summation, maximum, and average are commonly employed in neural network architectures to integrate information. These operators are particularly effective for merging features with spatial characteristics from different agents. For example, Marvasti et al. utilize non-parametric element-wise summation to fuse \gls{BEV} features from multiple sources \cite{BandwidthAdaptiveFeatureSharing-2020-marvastia}, ensuring comprehensive inclusion of available data. However, features with larger magnitudes may disproportionately affect the outcome, potentially overshadowing smaller yet significant inputs. Guo et al. \cite{SlimFCPLightweightFeatureBasedCooperative-2022-guoa} introduce a lightweight feature-based \gls{CP} framework employing the maxout operator, which excels in emphasizing the most critical features or activations while being robust against variations in the number of contributing agents. Despite its effectiveness, the maximum operator risks discarding valuable contextual information by focusing solely on the highest values. Non-parametric operations are favored for their computational efficiency and simplicity of implementation.
In contrast, parametric operators involve learnable parameters within the fusion module, such as convolution layers, offering a more adaptive approach to feature integration. Qiao et al. \cite{AdaptiveFeatureFusion-2023-qiaob} propose an adaptive feature fusion model that combines spatial and channel-wise feature fusion, leveraging both max and average pooling and trainable neural layers to enhance feature extraction selectively. Another prominent method is feature concatenation followed by a trainable neural layer, as demonstrated by Bai's feature fusion backbone \cite{DynamicFeatureSharing-2023-baia} using a dense CNN network to process concatenated features. This approach allows for the extraction of relevant information, significantly enhancing performance, though it may increase the feature dimensionality and computational demand. 

To conclude, traditional feature fusion techniques utilize reduction operators and often integrate trainable neural layers to extract the most relevant features effectively. This approach strives to balance performance improvement with computational efficiency, ensuring an optimal feature integration process.

\subsubsection{Attention-Based Feature Fusion}
\begin{figure}[t]
    \centering
    \subfloat[\label{fig:att_fusion}]{
        \includegraphics[width=0.47\linewidth]{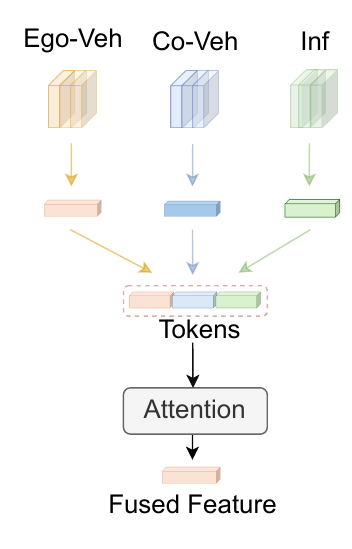}
    }
    \subfloat[\label{fig:graph_fusion}]{
        \includegraphics[width=0.47\linewidth]{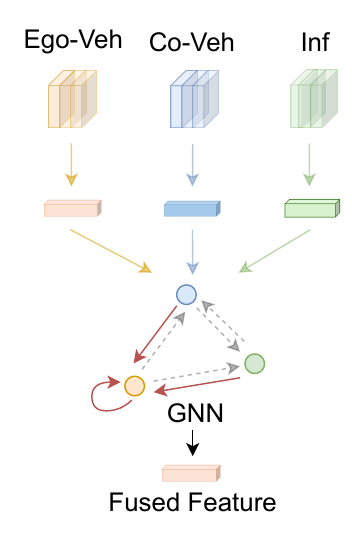}
    }
    \caption{Illustration of intermediate feature fusion: 
(a) Attention-based feature fusion, where features from ego-vehicle, cooperative vehicle, and infrastructure are transformed into tokens and aggregated using attention mechanisms; 
(b) Graph-based feature fusion, where features are represented as nodes and fused through message passing and node state updates in a GNN.}
    \label{fig:attention_fusion}
\end{figure}

The attention mechanism \cite{vaswani2017attention} is effective in capturing long-range dependencies and contextual relationships, making it highly suitable for feature weighting during fusion. For example, Wang et al. propose the F-Transformer \cite{FTransformerPointCloud-2022-wanga}, a point cloud fusion transformer that employs only Transformer encoder to fuse features from different views. As illustrated in Fig.~\ref{fig:att_fusion}, features from multiple entities are represented as tokens and forwarded to the attention module, where contextual relationships are learned to produce a fused feature representation. Unlike conventional Transformers, it omits position embeddings since the spatial arrangement of views is arbitrary, and there is no inherent ordering relationship between features from multiple perspectives. This design enhances robustness by preventing the model from making erroneous spatial assumptions and instead allowing it to focus on learning meaningful feature correlations across views. Xu et al. introduce the V2X-ViT \cite{V2XViTVehicletoEverythingCooperative-2022-xub}, designed to fuse information across on-road agents efficiently. It enhances self-attention by incorporating an additional weight matrix tailored to the type of the source and target agents. For example, agents are categorized as either vehicles or infrastructure, and the weight matrix dynamically adjusts to optimize collaboration based on their type. Hu et al. \cite{Where2commCommunicationEfficientCollaborative-2022-huc} introduce a spatial confidence-aware attentive fusion, where a spatial confidence map identifies perceptual uncertainty across different areas, serving as a basis for attention learning. This method prioritizes features with higher confidence during fusion, enhancing reliability. Lu et al. \cite{RobustCollaborative3D-2023-lub} propose a robust multiscale attentive fusion to mitigate noise from spatial misalignment. This method leverages features at different scales: finer scales provide detailed semantic information, while coarser scales offer robustness against spatial noise, thus maintaining semantic density and enhancing overall robustness. Yang et al. \cite{How2commCommunicationEfficientCollaborationPragmatic--yanga} address temporal noise using the spatial-temporal collaboration transformer (STCFormer), which features decoupled spatial and temporal cross-attention. STCFormer follows the architecture of a vanilla transformer but incorporates three customized modules: temporal cross-attention, decoupled spatial attention, and adaptive late fusion. The temporal cross-attention captures historical context across agents to enhance the representation of the current frame, mitigating point cloud sparsity caused by fast-moving objects. The decoupled spatial attention fuses spatial features from multiple agents, while the adaptive late fusion module integrates spatial features using weight maps. With these customized modules, STCFormer achieves robust detection performance even in dynamic environments. Despite its effectiveness, the computational complexity of attention mechanisms $O(N^2)$ poses scalability challenges. To address this, Yang et al. \cite{SpatioTemporalDomainAwareness-2023-yangb} utilized a deformable cross-attention module that selectively focuses on informative locations, significantly reducing computational demands and memory usage. Unlike standard attention, which assigns weights to all elements in the feature space, deformable attention selectively attends to a sparse set of informative locations, improving computational efficiency and scalability. 

LiDAR-based features inherently possess spatial characteristics suitable for per-location fusion via attention. Expanding cooperative perception to camera sensors, the \gls{BEV} feature is commonly used. However, due to the inherent uncertainty in depth estimation, visual \gls{BEV} features are less reliable than LiDAR features. To mitigate spatial misalignment, Huang et al. \cite{ActFormerScalableCollaborative-2024-huangc} propose a camera-based collaborative \gls{BEV} feature fusion using selective deformable attention, which fuses features based on an interest score threshold, emphasizing relevant and significant features for ego’s perception. The interest score is generated by a simple network that processes the BEV features as input, and during inference, only those features with scores above a threshold of 1 are selected.

In conclusion, Attention mechanisms and their variants play a pivotal role in collaborative feature fusion, enhancing feature integration across channel, spatial, and temporal dimensions. Techniques like confidence mapping and deformable attention are employed to improve fusion robustness and effectiveness further.

\subsubsection{Graph-Based Feature Fusion}
Graph structures are practical tools to represent complex relationships among data elements, where features are modeled as nodes interconnected by edges. These edges depict interactions between features. As shown in Fig.~\ref{fig:graph_fusion}, multi-agent collaboration can be conceptualized as a graph where nodes represent individual agents and edges represent inter-agent collaborations. \gls{GNN} \cite{scarselli2008graph} are well suited for processing such graph-structured data, enabling effective message passing and node state updates that facilitate information aggregation and propagation across the network. For instance, Wang et al. \cite{V2VNetVehicletoVehicleCommunication-2020-wangb} introduce a spatially-aware \gls{GNN} where each agent maintains a local graph with nodes holding state representations. These states are updated via a trainable neural network such as ConvGRU, which processes edge-weighted feature maps from all nodes to output updated node representations. This method incorporates historical context, enhancing temporal alignment and enabling joint object detection and motion prediction. Li et al. \cite{LearningDistilledCollaboration-2021-lib} propose a collaboration graph with trainable edge weights reflecting the collaboration strength between agents. These spatially and temporally aware weights allow agents to identify regions requiring collaboration dynamically. Xiang et al. \cite{HMViTHeteromodalVehicletoVehicle-2023-xiangb} further evolve this concept by introducing the H3GAT, a heterogeneous 3D graph attention model that integrates attention mechanisms with GNNs. This model captures local and global interactions, preserving detail and providing a comprehensive context. Liu et al. \cite{HPLViTUnifiedPerception-2023-liuc} employ a multiscale graph-attention technique to extract more comprehensive semantic information across different levels of granularity, enhancing feature integration. 

In conclusion, GNNs represent a sophisticated approach to modeling multi-agent collaboration. GNN fusion with attention mechanisms enables a nuanced capture of local and global contexts, facilitating a more detailed and integrated feature analysis.

\subsection{Late Collaboration}

\noindent Unlike the previous two approaches, each agent processes its sensor data independently in late collaboration and extracts relevant results or information. The processed information (often in the form of high-level results such as object-level data) is then shared with other agents or a central system, and the fusion happens at a higher, more abstract level.
Since only high-level, compact information is shared, late collaboration requires much less communication bandwidth compared to the other two approaches. Besides, it allows different agents to have varying sensor capabilities and still collaborate since only the abstract results are shared. However, some detail or precision may be lost during local processing.

The late collaboration approach is potentially modality agnostic, and in specific papers LiDAR data are often used as example\cite{CooperativePerceptionLearningBased-2023-liub},\cite{DistributedDynamicMap-2021-zhanga}, \cite{ModelAgnosticMultiAgentPerception-2023-xub}. 
Yu et al. \cite{MultistageFusionApproach-2022-yua} utilize both LiDAR and camera sensors.

A simple approach in late fusion is to use detected bounding boxes from multiple vehicles and weight them according to the detection confidence, such as Non-Maximum Suppression (NMS).  Another way of late fusion is to use an adapted Kalman filter. The collectively perceived tracks are considered as
measurements and integrated into the local environmental model. In general, late fusion approaches utilizes less information than the previous two approaches, and existing work's focus is mostly on improving fusion accuracy, given constraints such as heterogeneous density, low quality object proposals, overconfidence, etc.

According to the data exchanged among vehicles for cooperative object detection, the most common form is the detected object list.
Zhang et al. develop a three-stage fusion scheme: partitioning local objects, generating global objects, and eliminating overlapped boxes \cite{DistributedDynamicMap-2021-zhanga}. Yu et al propose a detection boxes fusion network for the late fusion, the inputs of which are vehicle-side and road-side boxes. This network performs coordination transform, filtering, object match, and combination~\cite{MultistageFusionApproach-2022-yua}. In \cite{UsAdversariallyRobust-2023-lia}, the authors assumes the existence of attackers and  propose an approach where each vehicle samples a subset of teammates and compares the results with and without the sampled teammates. Only after a consensus is verified, indicating no attackers among the participants, that the vehicle can output the perceptual results. Sampling ensures the scalability of this solution. 
Teufel et al. propose to incorporate collectively detected objects to enhance the local perception capabilities \cite{CollectivePVRCNNNovel-2023-teufela}. 
Song et al. uses optimal transport theory to correct inaccurate vehicle location and heading measurements using only object-level bounding boxes~\cite{CooperativePerceptionSystem-2023-songc}. 
Xu et al. propose mechanisms that considers confidence scores and mitigate the misalignment in box aggregation \cite{ModelAgnosticMultiAgentPerception-2023-xub}.
In \cite{EnvironmentawareOptimizationTracktoTrack-2022-volka}, the aim is to check perceived information for its trustworthiness and validity, so other information, such as covariance information, is also exchanged.

\subsection{Hybrid Collaboration}

Yuan et al. \cite{KeypointsBasedDeepFeature-2022-yuanc} combine late and intermediate collaboration. The fusion step combines multiple types of information: object box proposals (as in late collaboration), sensor pose, selected key point coordinates, and selected features (instead of all deep features as in intermediate collaboration). The aim is to reduce the redundancy of shared deep features to decrease the communication overhead.

Wang et al. \cite{PillarBasedCooperativePerception-2022-wanga} employ a two-stage fusion approach. In the first stage, an edge device collects and fuses the encoded Pillar features from the LiDAR data of all cooperative vehicles to generate a list of detected objects. This object list is then transmitted to the ego vehicle, which performs a late fusion by combining it with its own object list predictions.

Dao et al. \cite{PracticalCollaborativePerception-2024-daoa} propose a "late-early" collaboration framework for V2X cooperative perception. Here, objects detected by each connected agent at a past time closest to the present are broadcast. Detected objects shared between agents are propagated to the present timestamp using their velocities, computed by pooling point-wise scene flow. These propagated objects are then fused with the point cloud collected by the ego vehicle at the current time to enhance its perception. This work relaxes the assumption of inter-agent synchronization to agents sharing a shared time reference (e.g., GPS time) and acknowledges that agents produce detections at different rates. As a result, exchanged detections always have older timestamps than the timestamp of the query made by the ego vehicle, thus risking a misalignment between exchanged detections and their associated ground truth. To resolve this issue, the method simultaneously predicts both the velocities and locations of objects by pooling point-wise scene flow, effectively correcting for temporal discrepancies.

Liu et al. \cite{RegionBasedHybridCollaborative-2024-liu} also combine intermediate and late collaboration approaches. In the proposed fusion scheme, LiDAR data is divided into two types according to the overlapping area between the detection ranges of vehicles. For the overlapping area, intermediate collaboration is applied by sharing and fusing the features from different vehicles. For the non-overlapping area, late collaboration is conducted by generating and sharing the local detection result with an economic bandwidth. 

Xie et al. \cite{SoftActorCriticBased-2022-xiea} combine all fusion approaches. This framework enables vehicles to partition each point cloud frame into three parts: raw, feature, and object data, and exchange the data with other vehicles. To address spatial alignment issues, the receiving vehicle transforms these data levels from the sender's local coordinate system into its own. This transformation is achieved by constructing a matrix using additional information such as LiDAR sensor poses and GPS/IMU readings.

\subsection{Comparative Analysis}
\noindent Through a comprehensive review of collaboration types in \gls{CP} approaches, each level offers distinct advantages and challenges. Early collaboration, while providing the richest information from various agents, demands substantial bandwidth, and methods to address time latency at the raw data level remain underexplored. In contrast, late collaboration is bandwidth-efficient but sacrifices significant scene semantic context, resulting in decreased performance and robustness against noise. Intermediate collaboration balances efficiency and accuracy, enhancing noise robustness within the system. To optimize further, hybrid collaboration allows dynamic combinations of early, intermediate, or late collaboration based on accuracy demands. However, implementing hybrid frameworks is complex, mainly due to the challenges of managing heterogeneous data sources.

\section{Perception Tasks (RQ1)} \label{sec:perception_tasks}
\noindent There are various critical perception tasks that can benefit from a collaborative approach, including object detection, object tracking, motion prediction, semantic segmentation, and lane detection. \gls{OD} identifies and locates objects within a sensor frame, establishing a foundation for further perception processes. \gls{OT} involves monitoring the dynamic status of an object across multiple frames, while \gls{MP} aims to forecast the future movements or intentions of an object. \gls{SS} plays a crucial role in scene understanding, helping CAVs identify drivable areas and provide essential information for subsequent tasks. \gls{LD} is integral to determining road boundaries and lane markings, enabling CAVs to comprehend the geometry of the road network. This section provides a comprehensive overview of collaborative methods used in detection, tracking, motion prediction, semantic segmentation, and lane detection. Additionally, it introduces concepts of \gls{MT} and \gls{TA} pipelines, which are pivotal in enhancing the efficiency and accuracy of vehicle perception systems. 


Furthermore, the subcategories for each perception task are provided, with classifications based on representation formats. For example, \gls{OD} and \gls{SS} are categorized into 2D, 3D, and \gls{BEV} representations. \gls{MT} is divided into trajectory and \gls{BEV} map representations, while \gls{LD} is classified into curve-model and \gls{BEV} map representations. \gls{OT} is further categorized into tracking with \gls{COD} and tracking without \gls{COD}.



\subsection{Collaborative Object Detection}
\noindent Object detection is a fundamental perception task that focuses on identifying and locating relevant objects from raw sensor data. Typically, object detection results are presented as bounding boxes, each labeled with the corresponding object category. These bounding boxes can vary in representation: they may appear in 2D, \gls{BEV}, or 3D formats. 2D bounding boxes, often used in camera-based 2D object detection, capture object on image plane. \gls{BEV} representation disregards height and emphasizes the spatial layout of dynamic objects on the road plane, which is often sufficient for downstream tasks such as planning. The 3D format includes height and \mbox{z-axis} position, offering a more comprehensive view of the scene. This section discusses \gls{COD} across these different representations, with 3D being the most prevalent form in COD applications. All papers on \gls{COD} that meet our criteria are summarized in Tables \ref{tab:cod_1} and \ref{tab:cod_2}.

\subsubsection{2D}
Collaborative 2D object detection focuses on recognizing individual objects across multiple viewpoints on the image plane, which is particularly challenging. For instance, Anouar et al. \cite{NovelMultiviewPedestrian-2020-benkhalifaa}  propose a multi-view pedestrian detection approach that proceeds through a sequence of steps: monocular detection, geometric transformation, detection matching, and detection fusion. Similarly, Diego et al. \cite{BandwidthConstrainedCooperative-2022-mareza}  introduce a general COD framework, CP Faster-RCNN, designed to detect both vehicles and pedestrians. This framework extracts features from multiple viewpoints and uses an alignment module to warp them, followed by feature fusion to generate detection results. Mao et al. \cite{MoRFFMultiViewObject-2023-maoa} present MoRFF, a multi-view object detection pipeline that reduces communication costs by matching deep features rather than image data.

\subsubsection{BEV \& 3D}
\gls{BEV} and 3D bounding boxes are widely used to represent dynamic objects in autonomous driving applications. The \gls{BEV} representation simplifies the 3D bounding box by disregarding the height dimension, making it especially useful in camera-based pipelines that utilize \gls{BEV} features for detection. For instance, Hu et~al.~\cite{CollaborationHelpsCamera-2023-hub} present CoCa3D, a camera-only \gls{CP} pipeline that extracts \gls{BEV} features through a depth estimation module and voxel transformation module, subsequently decoding these features to predict object locations. Similarly, LiDAR-based pipelines can also leverage \gls{BEV} features by collapsing 3D voxel feature into a \gls{BEV} format, which avoids computationally demanding 3D convolutions. Wei et al. \cite{AsynchronyRobustCollaborativePerception--weic} introduce CoBEVFlow which utilze \gls{BEV} features to predict detection result as well as predict the flow of \gls{BEV} boxes. \gls{BEV} features are also advantageous in LiDAR-camera pipelines, as they facilitate the alignment and fusion of multi-modal data. For example, Yin et al. \cite{V2VFormerMultiModalVehicletoVehicle-2024-yina} present V2VFormer++, a multi-modal detection pipeline that first fuses \gls{BEV} features from LiDAR and camera data at the entity level and then combines the multi-modal features across entities in the CP fusion step, resulting in a streamlined and unified fusion process in \gls{BEV} space.  In addition to \gls{BEV}, 3D bounding boxes are widely used in LiDAR-only pipelines and occasionally in camera-only approaches. For instance, Wang et al. \cite{EMIFFEnhancedMultiscale-2024-wangc} introduce EMIFF, a camera-based pipeline that directly employs 3D voxel features to estimate the 3D position and dimensions of objects.

\subsection{Collaborative Semantic Segmentation}
\noindent Semantic segmentation is a process designed to assign a semantic class label to every pixel in an image or every point in a LiDAR scan. This technique offers a granular understanding of scenes, going beyond object detection that typically uses bounding boxes to localize objects. Semantic segmentation facilitates the precise delineation of object boundaries and enables the identification of multiple instances within the same scene. However, visual occlusions can create areas where semantic labels cannot be accurately predicted. Through V2X collaboration, CAVs can extend their \gls{FOV} and supplement the semantic labels of occluded areas, thus achieving a more comprehensive understanding of their surroundings. This section summarizes and categorizes \gls{CSS} approaches based on their representation format. All papers on \gls{CSS} that meet the selection criteria are listed in Table \ref{tab:semantic_segmentation}.

\begin{table*}[t]
\centering
\caption{ Overview of the methods for \acrfull{CSS}. V: Vehicle, I: Infrastructure, UAV: Unmanned Aerial Vehicle, Raw: Raw data fusion, Trad Feat: Traditional Feature Fusion, Atten Feat: Attention Feature Fusion. }
\label{tab:semantic_segmentation}
\begin{tabular}{c|cccccccc}
\toprule
 Method & Publication & Year & Modality & Agents & \textbf{Representation} & Scheme & Fusion & Code \\ \midrule
 When2com\cite{When2comMultiAgentPerception-2020-liua} & CVPR & 2020 & Camera & UAV & \textbf{2D} & Intermediate & Trad Feat & \cmark \\
Who2com\cite{Who2comCollaborativePerception-2020-liua} & ICRA & 2020 & Camera & UAV & \textbf{2D} & Intermediate & Trad Feat & \cmark \\ 
MASH\cite{OvercomingObstructionsBandwidthLimited-2021-glasera} & IROS & 2021 & Camera & UAV & \textbf{2D} & Intermediate & Atten Feat & \xmark \\ \midrule
GenBEV\cite{GeneratingEvidentialBEV-2023-yuanb} & ISPRS & 2023 & LiDAR & V & \textbf{BEV} & Early & Raw &  \cmark \\
CoBEVT\cite{CoBEVTCooperativeBird-2023-xuc} & CoRL & 2023 & Camera & V & \textbf{BEV} & Intermediate & Trad Feat & \cmark \\ \midrule
VICSS\cite{LiDARSemanticSegmentation-2023-liua} & VTC & 2023 & LiDAR & V,I & \textbf{3D} & Intermediate & Atten Feat & \xmark \\ 
CoHFF\cite{CollaborativeSemanticOccupancy-2024-songb} & CVPR & 2024 & Camera & V & \textbf{3D} & Intermediate & Atten Feat & \cmark \\
\bottomrule
\end{tabular}
\end{table*}

\subsubsection{2D} \label{sec:segmentation_2d}
2D semantic segmentation directly labels pixels within the 2D image plane. For instance, Liu et al. \cite{Who2comCollaborativePerception-2020-liua} introduce the Who2com framework, a pioneering collaborative approach to 2D semantic segmentation. This framework utilizes observations from multiple agents, including RGB images, aligned dense depth maps, and poses, to produce a 2D semantic segmentation mask for each agent.  Additionally, Liu's  subsequent When2com approach achieves improved performance with reduced bandwidth requirements \cite{When2comMultiAgentPerception-2020-liua}. In 2021, Glaser et al. \cite{OvercomingObstructionsBandwidthLimited-2021-glasera} introduce a novel pipeline that operates solely on raw image data, showing superior performance particularly in scenarios with image occlusions. This method employs an attention mechanism to identify visually similar patches across different perspectives, a crucial step when depth and pose information are absent.

\subsubsection{BEV} \label{sec:segmentation_bev}
\gls{BEV} semantic segmentaiton involves creating the top-down semantic map of the environment around a vehicle. In 2023, Yuan et al. \cite{GeneratingEvidentialBEV-2023-yuanb} present GenBEV, the first BEV collaborative segmentation approach based on LiDAR. In this model, 3D voxel features, extracted by a backbone network, are projected onto a \gls{BEV} map and processed by a task-specific head to segment both static road elements and dynamic objects. For camera-based \gls{BEV} segmentation, 2D image feature is typically converted into a top-down perspective by depth estimation. For instance,  Xu et al. \cite{CoBEVTCooperativeBird-2023-xuc} present the CoBEVT, a framework that enables collaborative generation of BEV map predictions. \gls{CAVs} extract \gls{BEV} features using the SinBEVT module and shares them with others. Received features are transformed to match the receiving vehicle’s coordinate system using the FuseBEVT module, which integrates fused axial attention (FAX) to efficiently manage local-global interactions. Local attention resolves pixel correspondence on occluded objects, while global attention assimilates contextual information such as road topology and traffic density.

\subsubsection{3D} \label{sec:segmentation_3d}
3D semantic segmentation provides a more detailed understanding of the environment by incorporating not only road-plane information but also the height and spatial dimensions of objects. For instance, Liu et al. \cite{LiDARSemanticSegmentation-2023-liua} introduce the first vehicle-infrastructure \gls{CSS} framework. This innovative approach begins by transforming the point cloud data from infrastructure sensors into the vehicle's coordinate system, followed by a feature extraction process. The extracted features are then compressed and transmitted to the vehicle. Upon reception, these features are divided into two subsets based on whether they fall inside or outside the overlapping \gls{FOV}. Each subset is processed separately to extract valuable information, then recombined and concatenated with the vehicle's own data. The integrated vehicle-infrastructure features are subsequently fed into a \gls{MLP} to predict the class labels of the points. Experiments conducted on a synthetic dataset demonstrate that the framework outperforms several classical single-vehicle LiDAR semantic segmentation algorithms, showcasing its enhanced performance and utility. Besides LiDAR, RGB cameras also support 3D semantic segmentation by labeling occupied voxels semantically. Song et al. \cite{CollaborativeSemanticOccupancy-2024-songb} present CoHFF framework, the first to explore collaborative semantic occupancy prediction. It consists of four modules: occupancy prediction, semantic segmentation, V2X feature fusion, and task feature fusion. Initial RGB data is processed for depth estimation and then transformed into a voxel representation, supplemented by a 3D occupancy encoder. The semantic segmentation task net maps RGB-derived 2D semantic features onto the 3D space using deformable cross-attention. These features are projected onto orthogonal planes, optimizing bandwidth usage. V2X feature fusion updates these features with input from various agents, enhancing the perception beyond the ego vehicle’s observations. The task-fusion module combines multi-agent features to reconstruct a comprehensive semantic occupancy grid, effectively mitigating issues caused by visual occlusion.

\subsection{Collaborative Object Tracking}
\noindent Object tracking involves locating and following object trajectories across sequences of video frames or point cloud data. Accurate tracking enables determination of an object's position, velocity, and acceleration, collectively understood as its motion status. Challenges in object tracking, such as dynamic changes in appearance, occlusions, and complex motion patterns, necessitate robust algorithms for continuous and precise tracking. Multi-view collaboration is a promising solution to address occlusions and maintain continuous tracking. This section categorizes collaborative object tracking into two approaches: tracking with \acrfull{COD} and tracking without COD. Tracking with COD integrates closely with collaborative detection outcomes, enhancing subsequent perception processes. Alternatively, tracking without COD offers flexibility by fusing perception results from multiple agents independently. Both methods predominantly utilize Kalman filters and their variants for tracking and incorporate uncertainty propagation to refine their tracking processes. All papers on \gls{COT} that meet the selection criteria are summarized in Table \ref{tab:object_tracking}.

\begin{table*}[t]
\centering
\caption{Overview of the methods for collaborative object tracking (COT). V: Vehicle, I: Infrastructure, Obj: Object, COD: Collaborative object detection, Atten Feat: Attention Feature Fusion, Obj Fusion: Object-level Fusion.}
\label{tab:object_tracking}
\begin{tabular}{c|ccccccccc}
\toprule
Method & Publication & Year & Modality & Entity & Scheme & Shared data & \textbf{Tracker} & Fusion & Code \\ \midrule
Track-by-det\cite{3DMultiObjectTracking-2023-sub} & IV & 2023 & Agnostic & V,I & NA & Obj & \textbf{with COD} & NA &  \xmark \\
HYDRO-3D\cite{HYDRO3DHybridObject-2023-menga} & T-IV & 2023 & LiDAR & V,I & Intermediate & Feature & \textbf{with COD}  & Atten Feat &  \xmark \\
FFTrack\cite{V2XSeqLargeScaleSequential-2023-yub} & CVPR & 2023 & LiDAR & V,I & Intermediate & Feature & \textbf{with COD} & Atten Feat & \xmark \\
MOT-CUP\cite{CollaborativeMultiObjectTracking-2024-sua}  & RA-L & 2024 & Agnostic & V & NA & Obj & \textbf{with COD} & NA &  \xmark \\ \midrule
DMSTrack\cite{Probabilistic3DMultiObject-2024-chiuc} & ICRA & 2024 & Agnostic & V & Late & Obj & \textbf{without COD} & Obj Fusion & \cmark \\
\bottomrule
\end{tabular}
\end{table*}

\subsubsection{Tracking with COD}
Tracking with COD involves performing tracking based on results from collaborative object detection. For instance, Su et al. \cite{3DMultiObjectTracking-2023-sub} propose a 3D multi-object tracking (3D-MOT) framework that utilizes results from collaborative detection. The process begins with the tracker receiving collaborative detection results, followed by the estimation of object states at the next frame using a Kalman filter. The states are then matched to update the tracked object’s status and initialize any new objects detected. This approach significantly reduces false negatives and positives compared to individual 3D-MOT setups. Additionally, Su et al.  \cite{CollaborativeMultiObjectTracking-2024-sua} introduce a method to address uncertainty in detection, termed MOT-CUP. This framework quantifies uncertainty using conformal prediction, assuming a Gaussian distribution, which is incorporated into a Standard Deviation-based Kalman Filter (SDKF) for enhanced prediction accuracy.

\subsubsection{Tracking without COD}
Tracking without COD utilizes lists of detected objects from multiple agents to enable cooperative tracking. For instance, Chiu et al. \cite{Probabilistic3DMultiObject-2024-chiuc} present DMSTrack framework, a differentiable multi-sensor Kalman filter facilitates 3D multi-object tracking. Uniquely, this framework decentralizes the prediction of object state covariances, allowing each vehicle to independently predict uncertainties associated with its detections. These detected object states, along with their predicted uncertainties, are then transformed from local to global coordinate systems before being shared with neighboring vehicles. Once integrated, these data inform the Kalman filter’s prediction and update stages, allowing for continuous and robust tracking by effectively managing the detection uncertainties from various agents.

\subsection{Collaborative Motion Prediction}
\noindent Motion prediction involves forecasting the future states of moving objects using historical data, a critical capability for autonomous navigation. Accurate predictions of dynamic entities’ trajectories allow systems to make right decisions, thereby enhancing safety and operational efficiency. The task becomes increasingly complex in environments with multiple interacting agents due to the nonlinear and unpredictable nature of agent interactions.

Collaborative motion prediction leverages the collective intelligence of multiple observing agents, integrating diverse data sources to enhance the accuracy and robustness of predictions. This cooperative approach not only mitigates the effects of individual sensor occlusions but also provides a more reliable prediction framework compared to isolated mechanisms. Motion prediction can be descirbed as forecasting the trajectory of bounding boxes or forcasting the \gls{BEV} map. All papers on \gls{CMP} that meet the selection criteria are summarized in Table \ref{tab:motion_prediction}.

\begin{table*}[t]
\centering
\caption{Overview of the method for collaborative motion prediction (CMP). V: Vehicle, I: Infrastructure, Raw: Raw data fusion, Trad Feat: Traditional Feature Fusion, Atten Feat: Attention Feature Fusion, Obj Fusion: Object-level Fusion, Graph: Graph-based Fusion.}
\label{tab:motion_prediction}
\begin{tabular}{c|cccccccc}
\toprule
 Method & Publication & Year & Modality & Entity & Scheme & \textbf{Representation} & Fusion & Code \\ \midrule
V2VNet\cite{V2VNetVehicletoVehicleCommunication-2020-wangb} & ECCV & 2020 & LiDAR & V & Intermediate & \textbf{Trajectory} & Graph & \cmark \\ 
V2VNet-Robust\cite{LearningCommunicateCorrect-2021-vadivelub} & CoRL & 2021 & LiDAR & V & Intermediate & \textbf{Trajectory} & Hybrid(Atten Feat, Graph) &  \xmark \\
Late-early\cite{PracticalCollaborativePerception-2024-daoa}& IEEE T-ITS & 2024 & LiDAR & V,I & Hybrid & \textbf{Trajectory} & Hybrid(Raw,Obj) & \cmark \\ \midrule
BEV-V2X\cite{BEVV2XCooperativeBirdsEyeView-2023-changa} & IEEE T-IV & 2023 & Camera & V,I & Intermediate & \textbf{BEV Map} & Atten Feat &  \xmark \\
V2XFormer\cite{DeepAccidentMotionAccident-2024-wanga} & AAAI & 2024 & Camera & V,I & Intermediate & \textbf{BEV Map} & Trad Feat & \cmark \\
\bottomrule
\end{tabular}
\end{table*}

\subsubsection{Trajectory}
Dynamic objects in environment can be represented through the bounding boxes with attritubes such as position and shape. In this case, motion prediction means to predict a sequence of future position of the bounding boxes, known as the trajectory. For instance, Wang et al. \cite{V2VNetVehicletoVehicleCommunication-2020-wangb} introduce V2VNet, a pioneering collaborative framework designed for simultaneous perception and prediction, termed \gls{PandP}. This approach not only enhances performance but also increases computational efficiency compared to traditional two-step processes. V2VNet extends individual \gls{PandP} capabilities by integrating \gls{V2V} communication. The model captures multi-scale historical data using Inception-like convolutional blocks \cite{Ding_2021_CVPR} for accurate forecasting. After integrating data across different agents, the combined feature map is processed through dual networks that deliver detection and motion forecasting outcomes. In 2021, Vadivelu et al. \cite{LearningCommunicateCorrect-2021-vadivelub} enhance V2VNet by addressing pose errors, thus improving accuracy. Additionally, Dao et al. \cite{PracticalCollaborativePerception-2024-daoa} present a LiDAR-based method for scene flow prediction, called Aligner, which can be adapted for motion forecasting. Aligner predicts the movement of point-wise features extracted from LiDAR point clouds, achieving precise scene flow predictions.

\subsubsection{BEV Map}
\gls{BEV} map can naturally combine the static road map and dynamic object map together, which benefits the motion prediction of objects. Wang et al. \cite{DeepAccidentMotionAccident-2024-wanga} introduce a camera-based framework, V2XFormer, which builds upon the capabilities of BEVerse\cite{zhang2022beverse}. V2XFormer utilizes the Swin-Transformer\cite{liu2021swin} to extract \gls{BEV} features and incorporates a multi-task head that simultaneously addresses detection and motion prediction tasks. This model introduces the V2XFusion module, which integrates \gls{BEV} features from multiple vehicles, enhancing collaborative perception capabilities. Chang et al. \cite{BEVV2XCooperativeBirdsEyeView-2023-changa} introduce BEV-V2X, a pioneering framework for cooperative prediction of \gls{BEV} occupancy grid maps. This framework represents dynamic objects and road structures within the \gls{BEV} occupancy grid on the map, capturing the dynamics of the scene over time. BEV-V2X leverages historical and current \gls{BEV} map data to forecast future \gls{BEV} maps within a three-second timeframe.

\subsection{Collaborative Lane Detection}
\noindent Lane detection is a critical component for \gls{ADAS} and automated driving (AD), as it provides essential information for path planning and vehicle control. \gls{HD Map}, though effective, is expensive to create, maintain, and scale. This makes real-time lane detection and online \gls{HD Map} learning increasingly important. However, lane detection, like other perception tasks, faces challenges such as visual occlusion and limited perception range, particularly in urban intersections with dense traffic, where multi-agent collaboration offers a potential solution. This section categorizes collaborative lane detection into two main approaches: curve-model-based methods and \gls{BEV}-map-based methods. Lane information can be represented using curve models, which are more data-efficient and require less bandwidth, or \gls{BEV} segmentation, which provides pixel-level detail with higher resolution and greater robustness to noise. While both approaches offer substantial potential, they remain under-explored and require deeper investigation. All papers on \gls{CLD} that meet the selection criteria are summarized in Table \ref{tab:lane_detection}.

\begin{table*}[t]
\centering
\caption{ Overview of the methods for \gls{CLD}. V: Vehicle, Trad Feat: Traditional Feature Fusion.}
\label{tab:lane_detection}
\begin{tabular}{c|cccccccc}
\toprule
Method & Publication & Year & Modality & Entity & Scheme & \textbf{Representation} & Fusion & Code \\ \midrule
Co-mapping\cite{CooperativeRoadGeometry-2020-sakra} & IEEE CAVS & 2020 & Camera & V & Late & \textbf{Curve-model} & Kalman filter &  \xmark \\
CoLD Fusion\cite{CoLDFusionRealtime-2023-gamerdingera} & IEEE IV & 2023 & Agnostic & V & Late & \textbf{Curve-model} & Spline-based Fusion &  \xmark \\ \midrule
LaCPF\cite{EnhancingLaneDetection-2024-jahna} & ROBOT AUTON SYST & 2024 & Agnostic & V & Late & \textbf{BEV map} & Trad Feat & \xmark \\
\bottomrule
\end{tabular}
\end{table*}

\subsubsection{Curve-Model-Based}
Curve-model-based methods represent the lane information as mathematical curves, enabling efficient data sharing and processing. For example, Sakr et al. \cite{CooperativeRoadGeometry-2020-sakra} propose a cooperative road geometry estimation framework, where sensor-rich vehicles share perceived road information with other vehicles. The road is divided into multiple connected segments, with each segment described by a clothoid-based model that uses parameters such as position, initial curvature, and curvature change rate. These parameters can be transmitted via V2X communication to extend the perception range. However, this approach does not account for fusing local lane detection data. To address this, Gamerdinger et al. \cite{CoLDFusionRealtime-2023-gamerdingera} introduce convoy fusion and spline fusion methods, which handle scenarios with and without overlapping lanes, respectively. Convoy fusion uses a weighted mean to merge lane data, assuming that closer lane detection is more accurate. For non-overlapping segments, spline fusion reconstructs the road between visible segments to provide a complete lane model.

\subsubsection{BEV-Map-Based}
While these methods represent the road using segmented curves, Jahn et al. \cite{EnhancingLaneDetection-2024-jahna} propose LaCPF, a different approach with the lightweight collaborative lane detection framework. In this method, roads are represented as static \gls{BEV} maps, transforming lane detection into a \gls{BEV} segmentation task. Each vehicle generates its own \gls{BEV} road segmentation, which is shared via V2X with neighboring vehicles. After aligning all local \gls{BEV} data into the same coordinate system, a fusion process using an encoder-decoder architecture combines the data into a comprehensive segmentation result.

Lane information can be represented through either curve models or \gls{BEV} segmentation. Curve models are more data-efficient and require less bandwidth, while \gls{BEV} segmentation provides pixel-level detail, offering higher resolution and greater robustness to noise. Both approaches have significant potential but remain underexplored.

\subsection{Multi-Task and Task-Agnostic}
\noindent Autonomous vehicle navigation requires addressing various perception tasks, from object detection to semantic segmentation. Traditionally, these tasks are performed independently, consuming significant computational resources. To optimize resource usage and enhance performance across multiple tasks simultaneously, researchers have proposed multi-task learning pipelines to address multiple perception tasks. All papers on multi-task and task-agnostic method that meet the selection criteria are summarized in Table \ref{tab:multi_task_and_task_agnostic}. For example, V2XFormer \cite{DeepAccidentMotionAccident-2024-wanga} simultaneously performs object detection, motion prediction, and accident prediction. Similarly, CoBEVT \cite{CoBEVTCooperativeBird-2023-xuc} handles both object detection and semantic segmentation in parallel. V2VNet \cite{V2VNetVehicletoVehicleCommunication-2020-wangb} is also able to conduct object detection and motion prediction at the same time.

\begin{table*}[t]
\centering
\caption{ Overview of methods for multi-task pipeline and task-agnostic pipeline. OD: Object detection, OT: Object tracking, MP: Motion prediction, AP: Accident prediction, SS: Semantic segmentation, V: Vehicle, I: Infrastructure, Raw: Raw data fusion, Trad Feat: Traditional Feature Fusion, Atten Feat: Attention Feature Fusion, Obj Fusion: Object-level Fusion, Graph: Graph-based Fusion.. }
\label{tab:multi_task_and_task_agnostic}
\begin{tabular}{c|cccccccc}
\toprule
 Method & Publication & Year & Modality & Entity & Scheme & Fusion & Task & Code \\ \midrule
 V2VNet \cite{V2VNetVehicletoVehicleCommunication-2020-wangb} & ECCV & 2020 & LiDAR & V & Intermediate & Graph & OD,MP &  \xmark \\
Robust V2VNet\cite{LearningCommunicateCorrect-2021-vadivelub} & CoRL & 2021 & LiDAR & V & Intermediate & Atten Feat, Graph & OD,MP & \xmark \\
BEV-V2X\cite{BEVV2XCooperativeBirdsEyeView-2023-changa} & IEEE T-IV & 2023 & Agnostic & V, I & Intermediate & Atten Feat & SS,MP & \xmark \\
HYDRO-3D\cite{HYDRO3DHybridObject-2023-menga} & IEEE T-IV & 2023 & LiDAR & V & Intermediate & Atten Feat & OD,OT & \xmark \\
FF-Tracking\cite{V2XSeqLargeScaleSequential-2023-yub} & CVPR & 2023 & LiDAR, Camera & V, I & Intermediate & Trad Feat & OD,OT & \cmark \\
CoBEVT\cite{CoBEVTCooperativeBird-2023-xuc} & CoRL & 2023 & Camera & V & Intermediate & Trad Feat & OD,SS & \cmark \\
V2XFormer\cite{DeepAccidentMotionAccident-2024-wanga} & AAAI & 2024 & LiDAR, Camera & V, I & Intermediate & Trad Feat & OD,MP,AP & \cmark \\
Late-early\cite{PracticalCollaborativePerception-2024-daoa} & IEEE T-ITS & 2024 & Camera & V, I & Hybrid & Hybrid(Raw,Obj) & OD,MP & \cmark \\ \midrule
STAR\cite{MultiRobotSceneCompletion--lib} & CoRL & 2022 & LiDAR & V & Intermediate & Trad Feat & Task-agnostic & \cmark \\
Core\cite{CoreCooperativeReconstruction-2023-wanga} & ICCV & 2023 & LiDAR & V & Intermediate & Trad Feat & Task-agnostic & \cmark \\
\bottomrule
\end{tabular}
\end{table*}

However, multi-task learning alone cannot fully address task heterogeneity issues. To tackle this challenge, researchers have proposed task-agnostic frameworks, such as \gls{CSC}, which can support various downstream perception tasks. In 2022, Li et al. \cite{MultiRobotSceneCompletion--lib} introduce STAR, a multi-agent scene completion framework where each agent learns to reconstruct the complete scene as viewed by all agents. STAR employs a spatial-temporal autoencoder architecture with a vision transformer (ViT) backbone to extract scene features. These features from various agents are aggregated with pose awareness and then processed by a decoder to predict the complete view. STAR demonstrates compatibility with single-agent perception models, allowing for integration without additional training. This approach significantly benefits scenarios with visual occlusion. In contrast to STAR, which conducts downstream tasks on completed scene representations, Wang et al. \cite{CoreCooperativeReconstruction-2023-wanga} propose CORE, a novel cooperative reconstruction framework. CORE performs downstream tasks directly on collaborative features, using reconstruction as additional guidance to develop a powerful encoder and fusion module. This approach generates informative intermediate representations that are then processed by task-specific decoders for various purposes, such as detection or segmentation. CORE has shown superior performance in both 3D object detection and \gls{BEV} semantic segmentation tasks while maintaining bandwidth efficiency. In conclusion, scene completion can serve as a guideline for feature learning, benefiting various downstream tasks. It can also be combined with single-agent perception models to enhance accuracy across different perception tasks.

\section{Approaches to address realistic issues (RQ1)} \label{sec:cp_issues}
\noindent In the initial stages of research, the focus on \gls{CP} primarily focused on the collaboration process and fusion strategies under ideal conditions, often relying on unrealistic assumptions such as precise localization and ideal communication conditions. However, CP algorithms encounter numerous challenges when deployed in real-world scenarios. This section summarizes these practical issues and their corresponding solutions.

\subsection{Localization Errors}

\begin{figure}[t]
    \centering
    \includegraphics[width=\linewidth]{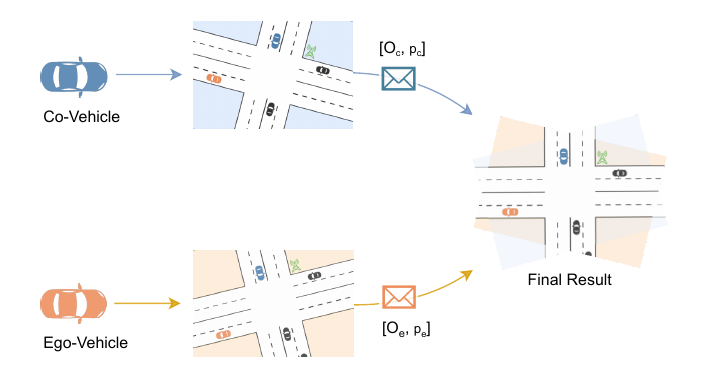}
    \caption{Illustration of the localization error issue: The cooperative vehicle transmits its sensing data and pose to the ego vehicle. The ego vehicle corrects the relative pose before alignment using consensus derived from the sensing data. Subsequently, the sensing data from multiple agents are fused based on the corrected pose.}
    \label{fig:pose_error}
\end{figure}

\noindent Accurate spatial alignment is essential for effective data fusion among different agents. However, errors in localization can lead to data misalignment, significantly impacting perception accuracy. To tackle this issue, researchers focus on correcting the relative pose before alignment \cite{JointPerceptionScheme-2022-ahmeda,FastClusteringCooperative-2024-kuanga,LearningCommunicateCorrect-2021-vadivelub,BEVV2XCooperativeBirdsEyeView-2023-changa,FeaCoReachingRobust-2023-guc,MoRFFMultiViewObject-2023-maoa,KeypointsBasedDeepFeature-2022-yuanc,CooperativePerceptionSystem-2023-songc,RobustCollaborative3D-2023-lub,RobustCollaborativePerception-2024-lei}, the process as shown in Figure~\ref{fig:pose_error}. Approaches to address localization errors are summarized in Table \ref{tab:method_addressing_pose_error}. The various approaches to address this problem can be categorized into three levels: raw-sensor, object, and feature levels. These methods are comparatively analyzed below.

\begin{table*}[t]
\centering
\caption{Overview of the methods for addressing pose error.\\V: Vehicle, I: Infrastructure, Raw: Raw sensor data, Feat: Feature, Obj: Object-level data.}
\label{tab:method_addressing_pose_error}
\begin{tabular}{c|ccccccc}
\toprule
Method & Publication & Year & Modality & Entity & \textbf{Data} & Pose correction approach & Code \\ \midrule
JointPerception\cite{JointPerceptionScheme-2022-ahmeda} & IEEE Sensors & 2022 & LiDAR & V & \textbf{Raw} & ICP Point cloud registration &  \xmark \\ 
FastClustering\cite{FastClusteringCooperative-2024-kuanga} & Cogn. Comput. & 2024 & LiDAR & V & \textbf{Raw} & ICP Point cloud registration &  \xmark \\ \midrule
Robust V2VNet\cite{LearningCommunicateCorrect-2021-vadivelub} & CoRL & 2021 & LiDAR & V & \textbf{Feat} & Markov random field &  \xmark \\ 
BEV-V2X\cite{BEVV2XCooperativeBirdsEyeView-2023-changa} & IEEE T-IV & 2023 & Agnostic & V,I & \textbf{Feat} & Global spatial aware attention &  \xmark \\ 
FeaCo\cite{FeaCoReachingRobust-2023-guc} & ACM MM & 2023 & LiDAR & V & \textbf{Feat} & Proposal Centers Matching & \cmark \\ 
MoRFF\cite{MoRFFMultiViewObject-2023-maoa} & IEEE VTC & 2023 & Camera & V & \textbf{Feat} & Multi-view feature matching &  \xmark \\ 
FPV-RCNN\cite{KeypointsBasedDeepFeature-2022-yuanc} & IEEE RA-L & 2024 & LiDAR & V & \textbf{Feat} & Semantic keypoint feature matching & \cmark \\ \midrule
Co-perception\cite{CooperativePerceptionSystem-2023-songc} & IEEE IV & 2023 & Agnostic & V & \textbf{Obj} & optimal transport theory &  \xmark \\ 
CoAlign\cite{RobustCollaborative3D-2023-lub} & ICRA & 2023 & LiDAR & V & \textbf{Obj} & Agent-Object Pose Graph Optimization & \cmark \\ 
FreeAlign\cite{RobustCollaborativePerception-2024-lei} & ICRA & 2024 & LiDAR & V & \textbf{Obj} & Graph matching & \cmark \\ 
\bottomrule
\end{tabular}
\end{table*}

\subsubsection{Raw-Sensor Level}
To achieve accurate relative positioning between cooperative vehicles, Ahmed et al.\cite{JointPerceptionScheme-2022-ahmeda} introduce a joint perception scheme that utilizes compressed point clouds. This approach employs point-to-plane \gls{ICP} registration to determine the optimal transformation matrix between the point clouds of the ego vehicle and the sender. This matrix is then used to achieve spatial alignment of the point clouds. While raw-sensor data level corrections typically provide precise pose estimations, they require the transmission of point cloud data, which consumes substantial bandwidth.

\subsubsection{Object-Level}
Song et al. \cite{CooperativePerceptionSystem-2023-songc} introduce the application of optimal transport theory to correct inaccurate vehicle locations and headings using only object-level bounding boxes. The pose correction process involves two stages. Given the local pose estimations of the ego vehicle and a cooperative vehicle, along with noisy measurements of perceived objects, the first step is to identify the co-visible region and associate the corresponding objects. Subsequently, an accurate transformation matrix $F$ is estimated by optimizing the following problem:
\begin{equation}
\label{ex:1}
\min_{\mathcal{F}} \sum_{(i,j) \in \mathcal{M}} \| x_i - \mathcal{F}(y_j) \|^2
\end{equation}
$x$ represents the position vector of objects perceived by the ego vehicle (similarly, $y$ for the cooperative vehicle), with $i$,$j$ denoting associated object pairs that represent the same physical target.

Similarly, Lu et al.\cite{RobustCollaborative3D-2023-lub} introduce another optimization-based approach, CoAlign, commonly used in Simultaneous Localization and Mapping (SLAM) algorithms, to correct the relative pose over various timeframes once the close-loop of the pose graph is identified. CoAlign introduces an agent-object pose graph to represent the relationships between agents and objects, aiming for consistency in the object's pose from different viewpoints. This consistency is pursued by formulating and minimizing a pose consistency error optimization problem. This method not only corrects the agents' pose but also enhances the positional accuracy of perceived objects. However, pose-graph optimization depends on a good initial guess, limiting its effectiveness in the presence of large noise.

Both optimization-based methods may be constrained by an underperforming object association step, which relies on prior knowledge of the pose. To overcome this limitation, Lei et al.  \cite{RobustCollaborativePerception-2024-lei} propose a spatial alignment approach, FreeAlign, which utilizes geometric consistency of a shared object map to associate objects without prior pose knowledge. Geometric consistency implies that co-visible regions should have a similar distribution of objects, with consistent geometric characteristics between object pairs. A graph model, where nodes represent objects and edges represent relative distances, can be used to depict this relationship. By identifying the most similar graph between two agents, corresponding nodes in the two graphs represent associated objects. Subsequently, FreeAlign employs RANSAC \cite{10.1145/358669.358692} to calculate the relative pose between these object maps.

While object-level pose correction is more communication-efficient, it is generally less accurate than methods using raw-sensor data due to higher noise levels in processed object-level data.

\subsubsection{Feature-Level}
To balance the performance of pose correction with communication efficiency, feature-level approaches have been developed. Vadivelu et al. \cite{LearningCommunicateCorrect-2021-vadivelub} introduce a pose regression module that estimates the relative pose through end-to-end learning, further refined by a Markov Random Field \cite{WANG20131610}. This approach has proven to enhance both object detection and motion forecasting tasks. Additionally, Chang et al. \cite{BEVV2XCooperativeBirdsEyeView-2023-changa} develop a method that incorporates global spatially-aware attention to improve spatial alignment. This technique utilizes prior map information to compare with current \gls{BEV} segmentation results, achieving more precise global positioning.

Feature matching is the most commonly used method at this level. For example, Gu et al. \cite{FeaCoReachingRobust-2023-guc} introduce FeaCo, which utilizes a robust feature-level proposal centers matching technique to calculate an accurate transformation matrix. This matching process, inspired by \gls{ICP}, minimizes the distance between original proposal centers from the ego vehicle and the transformed proposal centers from cooperative vehicles to derive the rotation matrix and translation vector. Similarly, MoRFF \cite{MoRFFMultiViewObject-2023-maoa} and FPV-RCNN \cite{KeypointsBasedDeepFeature-2022-yuanc} both employ feature keypoint matching to rectify the relative pose, enhancing pose alignment. While feature matching is straightforward to implement, its accuracy is limited by the spatial resolution of the features.

\subsubsection*{Discussion}
Pose correction methods vary significantly in their trade-offs among accuracy, bandwidth consumption, and robustness to noise. Approaches based on raw-sensor data typically offer the highest accuracy but require substantial bandwidth. In contrast, object-level methods are more efficient in terms of communication and computation but are less robust against noisy conditions. Feature-level approaches represent a middle ground, balancing accuracy and efficiency. The choice of a pose correction approach generally depends on the type of collaboration involved. When feature-level data is shared across agents, frameworks tend to utilize this same-level data for pose alignment, eliminating the need for external data sources. It allows for a more streamlined integration and efficient processing within the \gls{CP} system.

\subsection{Time Latency}
\noindent Effective data fusion in \gls{CAVs} necessitates that data be temporally aligned. In practice, \gls{CAV}s synchronize their clocks with Coordinated Universal Time (UTC) primarily through Global Navigation Satellite System (GNSS) signals, while the Network Time Protocol (NTP) may serve as a fallback when GNSS is unavailable. Achieving perfect alignment, however, remains challenging due to factors such as communication delays, interruptions, heterogeneous processing times, and varying data rates across vehicles, all of which can introduce temporal misalignment. To mitigate these issues, three levels of data are utilized: object-level data, feature-level data, and occupancy-level data.

\subsubsection{Object-Level}
The object-level approach is frequently applied in late collaboration scenarios to adjust for the movement of dynamic objects across different time frames. This method uses motion models to predict the position of an object at the current timestamp based on its previous data frame. For example, Su et al. \cite{3DMultiObjectTracking-2023-sub} employ the Constant Acceleration (CA) motion model to predict future positions of objects, allowing for more accurate data synchronization and integration in \gls{CAVs}.

\begin{table*}[t]
\centering
\caption{Overview of methods for addressing latency at the feature level. V: Vehicle, I: Infrastructure.}
\label{tab:method_addressing_latency}
\begin{tabular}{c|cccccc}
\toprule
Method & Publication & Year & Modality & Entity & Approach & Code \\ \midrule
SyncNet\cite{LatencyAwareCollaborativePerception-Lei-22} & ECCV & 2022 & LiDAR & V & Time-series prediction & \cmark \\
UMC\cite{UMCUnifiedBandwidthefficient-2023-wangb} & ICCV & 2023 & LiDAR & V & Time-series prediction & \cmark \\ \midrule
CoBEVFlow\cite{AsynchronyRobustCollaborativePerception--weic} & NeurIPS & 2023 & LiDAR & V & BEV/ROI flow prediction & \cmark \\ \midrule
FFNet\cite{FlowBasedFeatureFusion-2023-yue} & NeurIPS & 2023 & LiDAR & V,I & Feature flow prediction & \cmark \\
How2comm\cite{How2commCommunicationEfficientCollaborationPragmatic--yanga} & NeurIPS & 2023 & LiDAR & V & Feature flow prediction & \cmark \\
FF-Tracking\cite{V2XSeqLargeScaleSequential-2023-yub} & CVPR & 2023 & LiDAR, Camera & V,I & Feature flow prediction & \cmark \\
V2X-INCOP\cite{InterruptionAwareCooperativePerception-2024-renc} & IEEE T-IV & 2024 & LiDAR & V,I & Feature flow prediction &  \xmark \\
\bottomrule
\end{tabular}
\end{table*}

\subsubsection{Feature-Level}
Feature-level approaches have gained significant attention in addressing temporal alignment challenges for intermediate collaboration in \gls{CAVs}. The overview of feature-level approaches to address time latency is shown in Table \ref{tab:method_addressing_latency}. One notable approach is the latency-aware collaborative perception system introduced by Lei et al. \cite{LatencyAwareCollaborativePerception-Lei-22}. This system employs SyncNet, a latency compensation module utilizing \gls{LSTM} networks to estimate real-time features for collaboration. SyncNet has demonstrated effectiveness in enhancing intermediate collaboration, particularly in high-latency scenarios. However, real-time feature prediction can be computationally intensive. An alternative method, proposed by Wei et al. \cite{AsynchronyRobustCollaborativePerception--weic}, focuses on \gls{BEV} flow prediction within the \gls{CP} framework. This approach, called CoBEVFlow, generates spatial \gls{ROIs} based on received perceptual feature maps. By associating correlated ROIs across message sequences, it calculates motion vectors and estimates object positions at specific timestamps. The resulting \gls{BEV} flow map is used to adjust the spatial position of features, ensuring temporal alignment with ego features for efficient aggregation. Yu et al. \cite{FlowBasedFeatureFusion-2023-yue} introduce another technique called Feature Flow Net (FFNet), which employs feature flow prediction. This method describes feature changes over time, enabling direct prediction of aligned features at the current timestamp of collaborating vehicles. Similarly, the How2comm \cite{How2commCommunicationEfficientCollaborationPragmatic--yanga} framework utilizes feature flow prediction but refines it with a scale matrix. This scaling of predicted features has been shown to enhance temporal alignment effectiveness.

These feature-level approaches offer promising solutions for addressing temporal alignment challenges in \gls{CP} systems. By focusing on feature prediction, BEV-ROI-flow prediction, and feature-flow prediction, researchers are developing more robust and efficient methods for \gls{CAVs} to share and process perceptual information.

\subsubsection{Occupancy-Level}
Occupancy-Grid representation have emerged as a promising solution for 3D scene understanding, offering a more comprehensive depiction of dynamic environments compared to traditional object-level or feature-level methods. Zhang et al. \cite{RobustRealtimeMultivehicle-2023-zhangb} introduce the concept of occupancy flow prediction for temporal alignment in CP. This approach utilizes occupancy maps, which provide a more effective representation of the environment than raw point clouds and offer more detailed information compared to neural features. By predicting the flow of occupancy over time, these systems can better account for the dynamic nature of traffic scenarios and compensate for localization discrepancies between collaborating vehicles.

\subsubsection*{Discussion}
Object-level approaches offer efficiency and ease of implementation, making them attractive for real-time applications. These methods typically involve sharing high-level information such as object positions and velocities. However, their effectiveness is heavily dependent on the accuracy of upstream tasks, including object tracking and motion estimation. In scenarios with significant time latency, the propagation of errors from these tasks can lead to reduced overall system performance. To mitigate such challenges, feature-level approaches offer enhanced robustness by estimating environmental features with greater temporal precision. These methods often involve sophisticated prediction algorithms that can compensate for temporal mis-alignments in data from multiple vehicles. While more complex than object-level methods, feature-level approaches offer a better balance between computational efficiency and latency mitigation. Occupancy-level approaches, particularly those employing occupancy flow prediction, deliver the most comprehensive representation of the environment by modeling dynamic occupancy states over time. These methods provide detailed environmental information and offer significant benefits for temporal alignment.

\subsection{Communication Bandwidth Constraints}
\noindent In any system requiring communication, bandwidth can become a bottleneck when multiple entities participate and actively contribute to sharing data. In \gls{CP}, several entities (vehicles or infrastructure) collect, share, and aggregate perception data. In Europe, the \gls{ETSI} has specified functional requirements for collective awareness and cooperative perception applications. These specifications establish well-recognized networking constraints that must be considered in the design of CP algorithms. The vehicular network is ad hoc, participants establish communication in a self-organizing manner, and safety applications, such as \gls{CP}, generate messages within the limits of a \gls{PDU}. The PDU size is defined by the access layer protocol, which imposes bandwidth limitations and the requirement that safety messages such as \gls{CAM} and \gls{CPM} must fit into a single \gls{PDU}, as they are broadcast only once without retransmission or forwarding.

In practice, the bandwidth available for vehicular communication is highly constrained. For instance, IEEE 802.11p/DSRC provides a theoretical data rate of up to 27~Mbps in a 10~MHz channel, but the effective throughput in dense traffic environments is typically below 10~Mbps due to protocol overhead and channel contention~\cite{Kenney2011DedicatedSC,8031051}. Similarly, LTE-V2X operating in a 10~MHz channel can achieve around 15~Mbps under ideal conditions, yet its capacity decreases substantially as the number of vehicles increases~\cite{gonzalez2018analytical}. These limitations render the direct transmission of raw sensor data, such as LiDAR point clouds or high-resolution camera frames, largely infeasible in real-world deployments. This motivates the need for more efficient communication strategies that focus on transmitting intermediate features, or final perception outputs rather than raw data.

To mitigate these constraints, recent research has explored approaches such as data selection to filter transmitted information, data compression to reduce message size, and cooperator selection to restrict the number of entities participating in the CP algorithm. Table~\ref{tab:method_addressing_communication_issue} provides an overview of representative methods that address communication bandwidth challenges in \gls{CP}.

\begin{table*}[t]
\centering
\caption{Overview of methods for addressing communication efficiency. V: Vehicle, I: Infrastructure.}
\label{tab:method_addressing_communication_issue}
\begin{tabular}{c|cccccc}
\toprule
Method & Publication & Year & Modality & Entity & Approach for comm. efficiency & Code \\ \midrule
F-Transformer\cite{FTransformerPointCloud-2022-wanga} & IEEE SEC & 2019 & LiDAR & V & Data selection & \xmark \\
MASH\cite{OvercomingObstructionsBandwidthLimited-2021-glasera} & IROS & 2021 & Camera & UAV & Data selection & \xmark \\
Where2comm\cite{Where2commCommunicationEfficientCollaborative-2022-huc} & NeurIPS & 2022 & LiDAR, Camera & V & Data selection, Cooperator selection & \cmark \\
CoCa3D\cite{CollaborationHelpsCamera-2023-hub} & CVPR & 2023 & Camera & V,I & Data selection & \cmark \\
DFS\cite{DynamicFeatureSharing-2023-baia} & IEEE ITSC & 2023 & LiDAR & V,I & Data selection &  \xmark \\
What2comm\cite{What2commCommunicationefficientCollaborative-2023-yange} & ACM MM & 2023 & LiDAR & V,I & Data selection &  \xmark \\
How2comm\cite{How2commCommunicationEfficientCollaborationPragmatic--yanga} & NeurIPS & 2023 & LiDAR & V & Data selection, Data compression & \cmark \\
FPV-RCNN\cite{KeypointsBasedDeepFeature-2022-yuanc} & IEEE RA-L & 2024 & LiDAR & V & Data selection & \cmark \\
EdgeCooper\cite{EdgeCooperNetworkAwareCooperative-2024-luoa} & IEEE JSAC & 2024 & LiDAR & V,I & Data selection &  \xmark \\
SemanticComm\cite{SemanticCommunicationCooperative-2024-shenga} & J. Franklin Inst. & 2024 & LiDAR & V & Data selection &  \xmark \\
PillarAttention\cite{PillarAttentionEncoder-2024-bai} & IEEE IoT-J & 2024 & LiDAR & V,I & Data selection & \xmark \\
CenterCoop\cite{CenterCoopCenterBasedFeature-2024-zhou} & IEEE RA-L & 2024 & LiDAR & V,I & Data selection &  \xmark \\ \midrule
AFS-COD\cite{BandwidthAdaptiveFeatureSharing-2020-marvastia} & IEEE CAVS & 2020 & LiDAR & V & Data compression & \xmark \\
Slim-FCP\cite{SlimFCPLightweightFeatureBasedCooperative-2022-guoa} & IEEE IoT-J & 2022 & LiDAR & V & Data compression & \xmark \\ \midrule
When2com\cite{When2comMultiAgentPerception-2020-liua} & CVPR & 2020 & Camera & UAV & Cooperator selection & \cmark \\
Who2com\cite{Who2comCollaborativePerception-2020-liua} & ICRA & 2020 & Camera & UAV & Cooperator selection & \cmark \\
Co3D\cite{Collaborative3DObject-2023-wanga} & IEEE T-ITS & 2023 & LiDAR & V,I & Cooperator selection & \xmark \\
\bottomrule
\end{tabular}
\end{table*}

\subsubsection{Data Selection}
Data selection becomes necessary when, in the \gls{ETSI} \gls{ITS} scenario, there are limitations on how much data can or should be sent. Perception algorithms output data with varying accuracy, which can lead to filtering data based on the accuracy or confidence in the quality of the perception \cite{etsiCPM}. Luo et al. \cite{EdgeCooperNetworkAwareCooperative-2024-luoa} propose a Voxelization-based strategy for LiDAR data where the detection model groups samples into voxels. When transmission is required, the number of points from a voxel is limited while still being able to represent an object. Wang et al. \cite{FTransformerPointCloud-2022-wanga} employ data selection as a two-step process with negotiation and transmission. This method divides the view (perception field of view) into sections called pillars. Pillars with occlusion or partial occlusion require auxiliary information requested through the negotiation step. The relevant pillars are sent in the transmission step. Yang et al. \cite{What2commCommunicationefficientCollaborative-2023-yange} utilize a request-response methodology for cooperation where the ego vehicle broadcasts a request based on a filtered importance map generated from the feature map. Neighboring vehicles respond based on specificity and consistency constraints contained in the request.

\subsubsection{Data Compression}
Another means of reducing bandwidth is through data compression. This can be achieved by employing compression algorithms that reduce the binary representation of the same data and can be decompressed at the receiver. However, this adds overhead in both operations. Some work utilizes a change in the encoding of data, such as Slim-FCP \cite{SlimFCPLightweightFeatureBasedCooperative-2022-guoa}, where feature maps are reduced with a negligible degradation to the recall performance of the CP algorithm. Marvasti et al. \cite{BandwidthAdaptiveFeatureSharing-2020-marvastia} utilize a \gls{CNN} encoder solution to transform the feature map into a lower dimension. This compressed feature map is transmitted along with GPS information to other cooperative entities. Compression through transformation implies a trade-off between feature map accuracy (post-decompression) and transmission bandwidth requirements. 

\subsubsection{Cooperator Selection}
Cooperator selection restricts the number of entities in the vehicular network participating in the \gls{CP} algorithm. This way, the number of messages being transmitted can be reduced. However, the choice of participating entities is not trivial, since designing effective selection metrics is challenging, including determining the factors that should be incorporated into these metrics. Wang et al. \cite{Collaborative3DObject-2023-wanga} utilize a scoring system among participants that share feature maps encoded into query features using \gls{CNN}. Each participant then scores these query features to select their communication targets. Liu et al. \cite{When2comMultiAgentPerception-2020-liua,Who2comCollaborativePerception-2020-liua} present a three-stage handshake among entities to establish a group of participants to communicate. Participants calculate a matching score based on the correlation between two entities, which represents the amount of information one entity can provide for the other.

\subsubsection*{Discussion}
The three strategies - data selection, data compression, and cooperator selection - offer distinct methods for mitigating communication bandwidth constraints in \gls{CP}. Data selection focuses on transmitting the most relevant information, optimizing bandwidth but risking incomplete perception if criteria are overly restrictive. Data compression achieves bandwidth efficiency through compact representations but introduces computational costs and potential loss of fidelity. Cooperator selection reduces the communication load by limiting participants, though the exclusion of key entities due to suboptimal metrics can compromise effectiveness. Combining these approaches could provide a balanced solution, leveraging their strengths to address bandwidth limitations comprehensively.

\subsection{Communication Interruptions}

\noindent Ad-hoc networks, such as the vehicular network, are prone to communication issues that lower the effectiveness of data transmission. In some cases, packets may fail to arrive due to collision, which we call communication interruption. One solution to this issue is proposed by Ren et al. \cite{InterruptionAwareCooperativePerception-2024-renc}, where missing data is estimated by prediction from a previous frame. In such cases where historical data from a known entity is available, missing frames can be estimated.

\subsection{Domain Shifts}
\noindent \gls{CP} frameworks for \gls{CAVs} face significant challenges due to domain shift, a problem often under-explored in the field. This section examines the approaches to address different types of domain shift caused by training data, sensor characteristics, and the transition from simulation to real-world environments (Sim2Real). An overview can be seen in Table~\ref{tab:method_addressing_domain_shift}.

\begin{table*}[t]
\centering
\caption{Overview of methods for addressing domain shift. V: Vehicle, I: Infrastructure.}
\label{tab:method_addressing_domain_shift}
\begin{tabular}{c|ccccccc}
\toprule
Method & Publication & Year & Modality & Entity & Domain gap & Approach for bridging gap & Code \\ \midrule
FDA\cite{BreakingDataSilos-2024-lic} & ICRA & 2024 & LiDAR & V,I & Dataset domain & Learnable Feature Compensation &  \xmark \\ \midrule
DI-V2X\cite{DIV2XLearningDomainInvariant-2023-xiangb} & AAAI & 2023 & LiDAR & V,I & LiDAR sensor domain & Domain invariant distillation framework & \cmark \\
HPL-ViT\cite{HPLViTUnifiedPerception-2023-liuc} & ICRA & 2024 & LiDAR & V & LiDAR sensor domain & Heterogeneous Graph-attention &  \xmark \\ \midrule
DUSA\cite{DUSADecoupledUnsupervised-2023-kongc} & ACM MM & 2023 & LiDAR & V,I & Sim2Real domain & Sim/Real-invariant features & \cmark \\
S2R-ViT\cite{S2RViTMultiAgentCooperative-2024-lib} & ICRA & 2024 & LiDAR & V & Sim2Real domain & Domain invariant feature learning &  \xmark \\
\bottomrule
\end{tabular}
\end{table*}

\subsubsection{Domain Shift Caused by Training Data}
Collaborative perception systems in CAVs often involve vehicles from different manufacturers, each employing its own perception pipeline. Even if these vehicles utilize the same neural network architecture for feature extraction, variations in their training data can still lead to inconsistencies in the extracted features. To address this challenge, Li et al. \cite{BreakingDataSilos-2024-lic} propose the Feature Distribution-aware Aggregation (FDA) framework. The FDA framework incorporates a Learnable Feature Compensation (LFC) module, designed as an encoder-decoder architecture with skip connections, to predict and adjust residual discrepancies in the shared features. By applying this residual compensation, the shared features are enhanced before being fed into the fusion module. The FDA framework has been shown to effectively restore detection performance, even in the presence of distribution gaps, demonstrating its efficiency in maintaining reliable perception.

\subsubsection{Domain Shift Caused by Sensor}
CAVs from different manufacturers may be equipped with varying LiDAR sensors, which introduces inherent domain gaps in the raw sensor data. To address this issue, Li et al. \cite{DIV2XLearningDomainInvariant-2023-xiangb} propose the DI-V2X framework for Vehicle-Infrastructure Collaborative 3D Object Detection. DI-V2X is designed to learn domain-invariant representations using a distillation-based approach. First, the Domain Mixing Instance Augmentation (DMA) module creates a domain-mixing 3D instance bank for both teacher and student models during training, ensuring better alignment in data representation. Following this, the Progressive Domain-Invariant Distillation (PDD) module encourages student models across different domains to progressively learn domain-invariant feature representations from the teacher model. Additionally, a Domain-Adaptive Attention Framework (DAF) is used to further close the domain gap by incorporating calibration-aware, domain-adaptive attention.

In contrast to the domain-invariant approach, Liu et al. \cite{HPLViTUnifiedPerception-2023-liuc} explore the use of heterogeneous graph-attention mechanisms to fuse features from different agents, each with domain-specific characteristics. In this method, vehicles equipped with different types of LiDAR are treated as heterogeneous collaborators, represented as distinct nodes in a graph. The cooperative interactions between these heterogeneous nodes are modeled as weighted edges, where the weights reflect different fusion strategies for effective collaboration across domain gaps.

\subsubsection{Domain Shift Caused by Sim2Real}
\gls{CP} models require a large amount of labeled real-world data for training. However, collecting and annotating this data is both challenging and costly. As a result, synthetic data has gained attention due to its ease of production and cost-effectiveness. Despite these advantages, there is a significant domain gap between simulated environments and the real world, particularly in terms of appearance and content realism. This gap often leads to poor performance when models trained on simulated data are evaluated on real-world data.

To address this issue, Kong et al. \cite{DUSADecoupledUnsupervised-2023-kongc} introduce the DUSA framework for \gls{CP}. DUSA employs a Location-Adaptive Sim2Real Adapter (LSA) module to selectively aggregate features from critical locations on the feature map. It then aligns the features between simulated and real-world data using a sim/real discriminator in an adversarial training process. The aligned features are subsequently fed into the fusion module, ensuring \gls{CP} remains unaffected by the Sim2Real gap. Similarly, Li et al. propose the S2R-ViT framework \cite{S2RViTMultiAgentCooperative-2024-lib}, which uses domain discriminators to extract domain-invariant features from both simulation and real-world environments. Unlike other methods, S2R-ViT not only inputs features from individual agents into the discriminator before fusion but also applies the discriminator to the fused features, enhancing feature generalization and improving model performance in real-world scenarios.

\subsubsection*{Discussion}
Domain shifts can be categorized by their severity, ranging from low to high: dataset distribution, sensor characteristics, and Sim2Real discrepancies. Solutions for addressing domain shift include heterogeneous fusion, feature compensation, and domain-invariant feature learning. Heterogeneous fusion involves combining features with weights without fully eliminating the domain shift, making it less effective for larger gaps such as Sim2Real. In contrast, feature compensation and domain-invariant feature learning both aim to minimize domain gaps by generating more consistent features before fusion. Domain-invariant features can be achieved through cross-domain knowledge distillation and adversarial training, effectively bridging the gap and enhancing model performance.

\subsection{Heterogeneity}
\noindent Heterogeneity within \gls{CP} systems presents a significant challenge, primarily caused by differences in sensors and perception models across agents. CAVs on the road are often manufactured by various companies, leading to differences in sensor types and data processing models across vehicles from different \gls{OEMs}. This section provides a summary of the approaches used to address both model heterogeneity and modality heterogeneity within CP systems, as listed in Table \ref{tab:method_addressing_hetero}.

\begin{table*}[t]
\centering
\caption{Overview of methods for addressing the problem of heterogeneity. V: Vehicle, I: Infrastructure.}
\label{tab:method_addressing_hetero}
\begin{tabular}{c|cccccp{5cm}c}
\toprule
Method & Publication & Year & Modality & Entity & Heterogeneity & Approach for addressing Hetero. & Code \\ \midrule
MPDA\cite{BridgingDomainGap-2023-xua} & ICRA & 2023 & LiDAR & V,I & Model & \raggedright Cross-Domain Transformer: unify the feature patterns from different agents & \cmark \\ \midrule
HGAN\cite{MultiModalVirtualRealFusion-2022-zhanga} & IEEE PAAP & 2022 & LiDAR, Camera & V,I & Modality & \raggedright Data format alignment: virtual 3D points from RGB & \xmark \\
HM-ViT\cite{HMViTHeteromodalVehicletoVehicle-2023-xiangb} & ICCV & 2023 & LiDAR, Camera & V & Modality & \raggedright Feature interaction: 3D Graph Attention & \cmark \\ 
HEAL\cite{ExtensibleFrameworkOpen-2024-luc} & ICRA & 2024 & Agnostic & V & Modality & \raggedright Feature alignment: Backward alignment mechanism & \cmark \\ 
\bottomrule
\end{tabular}
\end{table*}

\subsubsection{Model Heterogeneity}
Current CP frameworks leverage deep neural network features to balance perception accuracy and communication bandwidth. However, these frameworks typically assume that all CAVs use identical neural networks, which is not always feasible in real-world scenarios. When features are transmitted from different models, a significant domain gap can emerge, leading to a decline in performance within CP systems.

To address this issue, Xu et al. \cite{BridgingDomainGap-2023-xua} introduce the Multi-agent Perception Domain Adaptation (MPDA) framework, a plug-in module designed to work with most existing systems while preserving confidentiality. MPDA includes a learnable feature resizer to align features across multiple dimensions and a sparse cross-domain transformer for domain adaptation. A domain classifier is then used to distinguish whether the features originate from the source or target domain. Through adversarial training, the sparse cross-domain transformer learns to produce domain-invariant features. Although MPDA has shown to improve performance in heterogeneous environments, it still struggles to fully resolve significant performance drops.

\subsubsection{Modality Heterogeneity}
Most existing work focuses on homogeneous systems where \gls{CAVs} are equipped with identical sensor types, an assumption that is unrealistic for real-world applications and significantly limits the scalability of collaboration.

To address modality heterogeneity, Zhang et al. \cite{MultiModalVirtualRealFusion-2022-zhanga} introduce the Multi-Modal Virtual-Real Fusion Transformer (MVRF) for collaborative perception. MVRF enables cross-modality cooperation between LiDAR and RGB cameras by generating virtual points from RGB images and incorporating them with LiDAR data. 

In contrast to this data alignment approach, Xiang et al. \cite{HMViTHeteromodalVehicletoVehicle-2023-xiangb} propose the hetero-modal Vision-Transformer (HM-ViT) for collaborative perception, which utilizes Heterogeneous 3D Graph Attention. HM-ViT separately extracts \gls{BEV} features from LiDAR and camera streams, treating the features as distinct nodes on a collaborative graph. A 3D graph attention mechanism is then applied to learn cross-modality interactions, and the updated features are fed into separate heads for final predictions from each modality.

In addition to learning cross-modality interactions, Lu et al. \cite{ExtensibleFrameworkOpen-2024-luc} introduce HEAL, an extensible framework for open heterogeneous \gls{CP}. HEAL addresses heterogeneity by aligning features in a unified space using a multi-scale, foreground-aware Pyramid Fusion network. To integrate new agents with previously unseen models or sensor modalities, only the encoder part of the architecture on new agents needs retraining. This step aligns the new agents' \gls{BEV} feature space with the unified space, offering low training costs and making the solution scalable for open heterogeneity scenarios.

\subsubsection*{Discussion}
There are four primary approaches to addressing heterogeneity in \gls{CP} systems. The first is to account for heterogeneity in the fusion process by learning cross-heterogeneity interactions. Another approach is to align the data format or feature space, allowing for homogeneous fusion. Additionally, rather than focusing on alignment, one can enable fusion by learning domain-invariant features across heterogeneous agents.

\subsection{Adversarial Attacks}
\noindent \gls{CP} enhances scene understanding but is particularly vulnerable to adversarial attacks. Ensuring the safety of CAVs requires protecting them from such threats. While adversarial attacks have been extensively studied in the communications field, they have not been deeply explored within the context of \gls{CP} frameworks. 

The first study to address adversarial attacks in this domain is ROBOSAC \cite{UsAdversariallyRobust-2023-lia}, a general sampling-based framework for adversarially robust \gls{CP}.
ROBOSAC aims to achieve consensus among co-vehicles during collaboration, preventing significant deviations from individual perceptions. Its workflow involves several steps. First, a vehicle samples a subset of its teammates and compares the results with and without the sampled teammates. Next, it verifies the consensus across results to ensure no attackers are present. Finally, the vehicle produces a collaborative perception result. The key advantage of ROBOSAC is that it does not require prior knowledge of specific attack patterns, allowing it to be generalized to new types of adversarial attacks. ROBOSAC has been shown to significantly enhance the robustness of \gls{CP} while maintaining high perception accuracy under attack.

Despite this progress, robust design against adversarial attacks in \gls{CP} remains an under-explored area, requiring further investigation in the future.

%% file: sections/04_evaluation.tex
\section{Evaluation Methods for Collaborative Perception (RQ2-4)} \label{sec:evaluation}
Evaluation methods are a critical aspect of research on \gls{CP}, complementing the development of \gls{CP} approaches. This section provides an overview of the current evaluation methodologies employed in the surveyed studies. Section \ref{sec:eval_method} describes the evaluation methodologies in detail, while Section \ref{sec:eval_scenarios} focuses on the evaluation scenarios. Additionally, Section \ref{sec:eval_metrics} presents the metrics and ablation studies used to assess \gls{CP} approaches.



\subsection{Evaluation Methodology} \label{sec:eval_method}
\noindent To evaluate new algorithms for \gls{CP}, various methodologies are employed. This section provides an overview of the approaches used in the surveyed papers. Real-world datasets and synthetic datasets are discussed in Sections \ref{sec:real_world_dataset} and \ref{sec:synthetic_dataset}, respectively. Additionally, real-world experiments and simulation-based evaluations are summarized in Sections \ref{sec:real_world_exp} and \ref{sec:sim_exp}.

\subsubsection{Real World Datasets} \label{sec:real_world_dataset}
Table \ref{tab:real_world_ds} provides a summary of publicly available datasets for \gls{CP}. These datasets predominantly focus on vehicle-to-infrastructure (V2I) collaboration, with limited attention given to vehicle-to-vehicle (V2V) interactions. The absence of datasets that integrate both \gls{V2I} and \gls{V2V} collaborations highlights a significant gap in existing resources, underscoring the need for more comprehensive datasets to advance \gls{CP} research.

LiDAR emerges as the most frequently used sensor in these datasets, often complemented by RGB cameras to enhance visual perception. However, the exclusion of additional modalities, such as infrared cameras or radars, limits their utility in handling complex scenarios, particularly in adverse environmental conditions. Furthermore, the scale of these datasets remains small compared to single-entity perception datasets like NuScenes \cite{NuScenes} and Waymo \cite{waymo}, which feature over 200,000 frames. This disparity is further compounded by limited scenario diversity, as most datasets are constrained to daytime and clear weather conditions. To address these shortcomings, future datasets should incorporate a wider variety of scenarios, including nighttime and adverse weather, to better reflect real-world challenges in \gls{CP}.

The diversity in annotated object classes across these datasets also reveals notable inconsistencies. For example, DAIR-V2X \cite{DAIRV2XLargeScaleDataset-2022-yua} includes annotations for 10 distinct object classes, whereas others, such as \cite{HoloVICLargescaleDataset-2024-mac, LUCOOPLeibnizUniversity-2023-axmanna}, focus on fundamental categories like pedestrians, cyclists, cars, and trucks. Some datasets adopt a task-specific approach, such as \cite{NovelMultiviewPedestrian-2020-benkhalifaa}, which is exclusively dedicated to pedestrian detection. Although object detection is a central feature, support for advanced tasks like object tracking is limited, and motion prediction remains under-represented, highlighting an imbalance in task coverage.

Table \ref{tab:real_world_ds_2} delves deeper into \gls{V2X} configurations, examining critical aspects such as the number of connected vehicles, localization methods, and time synchronization protocols. Most datasets involve three or fewer \gls{CAVs}, as exemplified by LUCOOP \cite{LUCOOPLeibnizUniversity-2023-axmanna}. Time synchronization is generally asynchronous, with a maximum latency of 50 milliseconds between entities. To ensure accurate ground-truth localization, hybrid localization approaches are commonly employed, combining multiple techniques such as \gls{HD Map} and Real Time Kinematic (RTK) to minimize positional errors. These precise localization methods play a pivotal role in enhancing \gls{CP} system performance and fostering effective collaboration between vehicles and infrastructure.

\begin{table*}[t]
\centering
\caption{Overview of all publicly available real world datasets for \gls{CP} that include infrastructure perspective. L indicates LiDAR and RGB denotes camera sensor in the modalities. For the tasks the datasets include object detection (OD) -- 3D if not indicated otherwise --, object tracking (OT), motion prediction (MP) and domain adaption (DA). }
\begin{tabular}{c|cccccccc}
\toprule
Dataset      & Year & Collaboration & Modalities & Task     & Location & \# classes & \# co-frames & Diversity        \\ \midrule
T\&J \cite{FcooperFeatureBased-2019-chend}                  & 2019 & V2V           & L          & OD         & USA      & NA          & 100       & Not described    \\
I2V-MVPD \cite{NovelMultiviewPedestrian-2020-benkhalifaa}   & 2020 & V2I           & RGB        & OD (2D)    & Tunisia  & 1          & 4.7k      & Weather          \\
DAIR-V2X-C \cite{DAIRV2XLargeScaleDataset-2022-yua}         & 2022 & V2I           & L \& RGB   & OD         & China    & 10         & 13k       & Weather, daytime \\
V2V4Real \cite{V2V4RealRealWorldLargeScale-2023-xua}        & 2023 & V2V           & L \& RGB   & OD, OT, DA & USA      & 5          & 10k       & Scenario         \\
DAIR-V2X-Seq \cite{V2XSeqLargeScaleSequential-2023-yub}     & 2023 & V2I           & L \& RGB   & OD, OT, MP & China    & 10         & 7.5k      & Daytime          \\
LUCOOP \cite{LUCOOPLeibnizUniversity-2023-axmanna}          & 2023 & V2V           & L          & OD, MP     & Germany  & 4          & 13.5k     & Weather, daytime \\
HoloVIC \cite{HoloVICLargescaleDataset-2024-mac}            & 2024 & V2I           & L \& RGB   & OD, OT     & China    & 3          & 100k       & Scenarios        \\
TUMTrafV2X \cite{TUMTrafV2XCooperative-2024-zimmerc}        & 2024 & V2I           & L \& RGB   & OD, OT     & Germany  & 8          & 1k        & Daytime          \\
\bottomrule
\end{tabular}
\label{tab:real_world_ds}
\end{table*}

\begin{table}[htbp]
\centering
\caption{Additional technical information for the publicly available real world datasets. A ``-'' indicates the absence of information in the referenced publication.}
\label{tab:real_world_ds_2}
\begin{tabular}{c|ccc}
\toprule
Dataset      & CAVs & Localization & Synchronisation \\ \midrule
T\&J \cite{FcooperFeatureBased-2019-chend}                  & 2    & NA            & NA \\ 
I2V-MVPD \cite{NovelMultiviewPedestrian-2020-benkhalifaa}   & 1    & GPS          & async ($\sim$30ms)                                          \\
DAIR-V2X-C \cite{DAIRV2XLargeScaleDataset-2022-yua}         & 1    & Hybrid       & \makecell{ async (10$\sim$30ms)\\+ sync (\textless{}10ms)}  \\
V2V4Real \cite{V2V4RealRealWorldLargeScale-2023-xua}        & 2    & Hybrid       & async (\textless{}50ms)                                     \\
DAIR-V2X-Seq \cite{V2XSeqLargeScaleSequential-2023-yub}     & 1    & Hybrid       & \makecell{async + \\ sync (\textless{}10ms)}                \\
LUCOOP \cite{LUCOOPLeibnizUniversity-2023-axmanna}          & 3    & Hybrid       & NA                                                           \\
HoloVIC \cite{HoloVICLargescaleDataset-2024-mac}            & 1    & RTK/INS      & sync (\textless{}10ms)                                      \\
TUMTrafV2X \cite{TUMTrafV2XCooperative-2024-zimmerc}        & 1    & RTK/INS      & NA                                                           \\
\bottomrule
\end{tabular}
\end{table}

\subsubsection{Synthetic Datasets} \label{sec:synthetic_dataset}
\Cref{tab:synth_ds} summarizes the synthetic datasets used in \gls{CP}. These datasets are predominantly generated using CARLA, often paired with frameworks like OpenCDA \cite{9564825}, which integrates CARLA with SUMO for traffic simulation. The use of simulation significantly reduces the effort required to create scenarios involving multiple \gls{CAVs} compared to real-world settings. Unlike real-world datasets, synthetic datasets frequently support both \gls{V2V} and \gls{V2I} communication, making them highly versatile for \gls{CP} research. Additionally, many synthetic datasets enable cooperation among more than three \gls{CAVs}, further enhancing their applicability. These datasets also exhibit greater diversity in sensor modalities, incorporating LiDAR and RGB cameras, with some extending to include depth information.

Synthetic datasets exhibit greater diversity in the range of tasks they support. While object detection remains the primary focus, many datasets extend their scope to include tasks such as semantic segmentation and accident prediction. However, the scenario diversity is often constrained by a reliance on pre-existing maps from CARLA, which limits geographic variety and reduces the ability to replicate a wide range of real-world conditions accurately.

A significant characteristic of synthetic datasets is their reliance on idealized system conditions. As outlined in \Cref{tab:synth_ds_2}, most datasets assume perfect time synchronization between connected entities, with the exception of \cite{AsynchronyRobustCollaborativePerception--weic}. Additionally, simulators provide precise ground-truth localization, resulting in error-free localization performance. While these conditions simplify evaluation, they may not fully reflect the challenges of real-world scenarios.

Despite these limitations, synthetic datasets are often designed to replicate real-world conditions to better evaluate \gls{CP} system performance. For instance, \cite{DeepAccidentMotionAccident-2024-wanga} explores the impact of time latency and pose errors, demonstrating that increasing the number of \gls{CAVs} enhances robustness against such issues. Similarly, \cite{OPV2VOpenBenchmark-2022-xub} reveals a positive correlation between the number of\gls{CAVs} and the average precision of object detection, with performance improvements plateauing at around four \gls{CAVs}. These studies highlight the value of synthetic datasets in examining trade-offs and identifying limitations in \gls{CP} systems.

\begin{table*}[t]
\centering
\caption{General overview of synthetic datasets for \gls{CP}. L indicates LiDAR as modality. The datasets support object detection (OD), object tracking (OT), semantic segmentation (SS), motion prediction (MP) and accident prediction (AP). }
\begin{tabular}{c|cccccccc}
\toprule
Dataset                                                         & Year & Collaboration & Modalities & Task               & \# classes & \# co-frames & Source        & Diversity        \\ \midrule
CODD \cite{AdaptiveFeatureFusion-2023-qiaob}                    & 2021 & \makecell{V2V       }& L          & OD                 & 2          & 13k       & CARLA         & Scenario         \\
V2X-Sim1.0 \cite{LearningDistilledCollaboration-2021-lib}       & 2021 & \makecell{V2V       }& L \& RGBD  & OD                 & 3          & 10k       & CARLA \& Sumo & Scenario         \\
OPV2V \cite{OPV2VOpenBenchmark-2022-xub}                        & 2022 & \makecell{V2V       }& L \& RGB   & OD                 & 1          & 12k       & OpenCDA       & Scenario         \\
V2X-Sim 2.0 \cite{V2XSimMultiAgentCollaborative-2022-lie}       & 2022 & \makecell{V2V \& V2I}& L \& RGBD& OD, OT, SS         & 3          & 10k       & CARLA \& Sumo & Scenario         \\
V2XSet \cite{V2XViTVehicletoEverythingCooperative-2022-xub}     & 2022 & \makecell{V2V \& V2I}& L        & OD                 & 1          & 12k       & OpenCDA       & Scenario         \\
IRV2V \cite{AsynchronyRobustCollaborativePerception--weic}      & 2023 & \makecell{V2V       }& L \& RGB   & OD                 & NA          & 8k        & CARLA         & Scenario         \\
OPV2V-H \cite{ExtensibleFrameworkOpen-2024-luc}                 & 2024 & \makecell{V2V       }& L \& RGBD  & OD                 & 1          & 10k       & OpenCDA       & Scenario         \\
Semantic OPV2V \cite{CollaborativeSemanticOccupancy-2024-songb} & 2024 & \makecell{V2V       }& L \& RGBD  & SS                 & 12         & 10k       & OpenCDA       & Scenario         \\
DeepAccident \cite{DeepAccidentMotionAccident-2024-wanga}       & 2024 & \makecell{V2V \& V2I}& L \& RGB & OD, OT, SS, MP, AP & 2          & 57k       & CARLA         & Weather, daytime \\
\bottomrule
\end{tabular}
\label{tab:synth_ds}
\end{table*}

\begin{table}[t]
\centering
\caption{Overview of more technical aspects of the synthetic datasets. Default synchronization means perfect synchronization}
\label{tab:synth_ds_2}
\begin{tabular}{c|ccc}
\toprule
Dataset                                                         & \# CAVs & Synchronization & GNSS/IMU \\ \midrule
CODD \cite{AdaptiveFeatureFusion-2023-qiaob}                    & 2       & Default         & No       \\
V2X-Sim1.0 \cite{LearningDistilledCollaboration-2021-lib}       & 2-5     & Default         & Yes      \\
OPV2V \cite{OPV2VOpenBenchmark-2022-xub}                        & 2-7     & Default         & Yes      \\
V2X-Sim 2.0 \cite{V2XSimMultiAgentCollaborative-2022-lie}       & 2-5     & Default         & Yes      \\
V2XSet \cite{V2XViTVehicletoEverythingCooperative-2022-xub}     & 2-7     & Default         & Yes      \\
IRV2V \cite{AsynchronyRobustCollaborativePerception--weic}      & 2-5     & Asynchronous    & Yes      \\
OPV2V-H \cite{ExtensibleFrameworkOpen-2024-luc}                 & 2-7     & Default         & Yes      \\
Semantic OPV2V \cite{CollaborativeSemanticOccupancy-2024-songb} & 2-7     & Default         & Yes      \\
DeepAccident \cite{DeepAccidentMotionAccident-2024-wanga}       & 4       & Default         & No       \\
\bottomrule
\end{tabular}
\end{table}

\subsubsection{Real World Experiment} \label{sec:real_world_exp}
As discussed in the previous section, existing real-world datasets have significant limitations. When new algorithms offer advantages that cannot be effectively demonstrated using these datasets, dedicated experiments become essential. However, real-world experiments demand considerable time and financial resources, making them far less common than simulation-based evaluations.

For example, Sakr et al. \cite{CooperativeRoadGeometry-2020-sakra} conduct an experiment where a legacy vehicle follows a sensor-rich vehicle that transmits road geometry information. The aim is to estimate the road geometry ahead of the legacy vehicle using the data provided by the sensor-rich vehicle. Similarly, Li et al. \cite{MKDCooperCooperative3D-2024-lia} design experiments involving two vehicles equipped with LiDARs and cameras. Their study demonstrates two scenarios where \gls{V2V} perception outperforms single-entity perception, particularly in detecting distant objects beyond the range of LiDAR sensors. Additionally, they show how \gls{V2V} communication effectively reduces positioning errors in various road scenarios. Xie et al. \cite{SoftActorCriticBased-2022-xiea} adopt a different approach by conducting real-world experiments with two vehicles equipped with LiDARs and cameras. These vehicles collect data across three representative V2V scenarios, facilitating the validation of their algorithm under real-world conditions.

In general, real-world experiments remain rare due to the significant resources required. Most of these studies focus on \gls{V2V} cooperation, as it is easier to design and continues to be the most extensively researched form of \gls{V2X} collaboration.

\subsubsection{Simulation Experiment} \label{sec:sim_exp}
In simulation experiments, trends similar to those observed in real-world evaluations are evident. Table \ref{tab:simulation_experiment} summarizes studies employing simulation-based experiments, which either generate new datasets or adapt existing ones to evaluate specific approaches, as demonstrated in \cite{EnvironmentawareOptimizationTracktoTrack-2022-volka}.

As with real-world tests, \gls{V2V} communication remains the dominant approach, preferred over \gls{V2I} or combined \gls{V2I} and \gls{V2V} methods. Object detection continues to be the most frequently studied perception task \cite{BandwidthAdaptiveFeatureSharing-2020-marvastia,JointPerceptionScheme-2022-ahmeda,PillarBasedCooperativePerception-2022-wanga,DynamicFeatureSharing-2023-baia,CollectivePVRCNNNovel-2023-teufela,MultimodalCooperative3D-2023-chia,HPLViTUnifiedPerception-2023-liuc,VINetLightweightScalable-2023-baia,EdgeCooperNetworkAwareCooperative-2024-luoa,FastClusteringCooperative-2024-kuanga,VINetLightweightScalable-2023-baia,EdgeCooperNetworkAwareCooperative-2024-luoa,FastClusteringCooperative-2024-kuanga}. Some studies create tailored datasets to address specific requirements, such as semantic segmentation \cite{LiDARSemanticSegmentation-2023-liua,OvercomingObstructionsBandwidthLimited-2021-glasera} or lane detection \cite{CoLDFusionRealtime-2023-gamerdingera}.

CARLA\footnote{\url{https://carla.org}} is the most widely used simulator for \gls{CP} research, frequently utilized in customized configurations. AirSim\footnote{\url{https://microsoft.github.io/AirSim}} and Gazebo\footnote{\url{https://gazebosim.org}} are also commonly employed. Among these, only one study incorporated a network simulator to model realistic network data traffic and its impact on perception performance~\cite{EdgeCooperNetworkAwareCooperative-2024-luoa}.

Simulation studies often explore specific aspects of \gls{CP}. A recurring focus is determining the optimal number of cooperating vehicles to maximize performance~\cite{EdgeCooperNetworkAwareCooperative-2024-luoa,EnvironmentawareOptimizationTracktoTrack-2022-volka,BEVV2XCooperativeBirdsEyeView-2023-changa}. Kuang et al.~\cite{FastClusteringCooperative-2024-kuanga} investigate scenarios where V2V cooperation significantly outperforms single-vehicle perception. Similarly, Liu et al.~\cite{HPLViTUnifiedPerception-2023-liuc} analyze the effects of homogeneous versus heterogeneous sensor configurations on \gls{CP} performance across various conditions.

Another line of research examines federated learning for \gls{CP}. For instance, Zhang et al.~\cite{DistributedDynamicMap-2021-zhanga} implement a dynamic map fusion algorithm using federated learning to recover objects missed by individual systems, demonstrating its potential to enhance \gls{CP} performance.

\begin{table}[htbp]
\centering
\caption{Overview of simulation experiments. Supported tasks are object detection (OD), object tracking (OT), semantic segmentation (SS), motion prediction (MP), lane detection (LD), map fusion (MF). A ``-'' indicates the absence of information in the referenced publication.}
\label{tab:simulation_experiment}
\begin{tabular}{c|cccc}
\toprule
Reference                                                    & Year & V2X          & Simulator                  & Task \\ \midrule
\cite{BandwidthAdaptiveFeatureSharing-2020-marvastia}        & 2020 & V2V          & Volony (CARLA based)       & OD   \\
\cite{DistributedDynamicMap-2021-zhanga}                     & 2021 & V2V          & CARLA                      & MF   \\
\cite{OvercomingObstructionsBandwidthLimited-2021-glasera}   & 2021 & V2V (drones) & AirSim                     & SS   \\
\cite{EnvironmentawareOptimizationTracktoTrack-2022-volka}   & 2022 & V2V          & Based on other dataset     & OT   \\
\cite{JointPerceptionScheme-2022-ahmeda}                     & 2022 & V2V          & CARLA                      & OD   \\
\cite{PillarBasedCooperativePerception-2022-wanga}           & 2022 & V2V          & Gazebo                     & OD   \\
\cite{DynamicFeatureSharing-2023-baia}                       & 2023 & V2V \& V2I   & CARTI                      & OD   \\
\cite{LiDARSemanticSegmentation-2023-liua}                   & 2023 & V2I          & CARLA                      & SS   \\
\cite{BEVV2XCooperativeBirdsEyeView-2023-changa}             & 2023 & V2V \& V2I   & NA                          & MP   \\
\cite{CoLDFusionRealtime-2023-gamerdingera}                  & 2023 & V2V          & CARLA \& Resist            & LD   \\
\cite{CollectivePVRCNNNovel-2023-teufela}                    & 2023 & V2V          & CARLA \& Resist            & OD   \\
\cite{MultimodalCooperative3D-2023-chia}                     & 2023 & V2V          & CARLA                      & OD   \\
\cite{HPLViTUnifiedPerception-2023-liuc}                     & 2023 & V2V          & OpenCDA                    & OD   \\
\cite{VINetLightweightScalable-2023-baia}                    & 2023 & V2V \& V2I   & CARLA                      & OD   \\
\cite{EdgeCooperNetworkAwareCooperative-2024-luoa}           & 2024 & V2V \& V2I   & CARLA, Sumo, NS3           & OD   \\
\cite{FastClusteringCooperative-2024-kuanga}                 & 2024 & V2V          & OpenCDA                    & OD   \\ 
\bottomrule
\end{tabular}
\end{table}

\subsection{Evaluation Scenarios} \label{sec:eval_scenarios}

\subsubsection{Environmental Settings}
The methodologies for evaluating \gls{CP} algorithms, including datasets and experiments, were introduced in the previous section. This section examines the specific scenarios used for algorithm evaluation. In both real-world and simulation studies, environments are typically categorized into three primary types: urban, rural, and highway, as outlined in \Cref{tab:evaluation_scenarios}. Simulation studies offer a wider range of scenarios compared to real-world evaluations, largely due to the extensive use of CARLA and its pre-defined maps. However, this reliance on CARLA maps introduces limitations in the diversity of road environments and the assessment of specific road features, such as intersections. Although many studies specify which CARLA map is utilized, detailed information about the types of intersections or the specific routes examined is often absent. Commonly evaluated road configurations, including cross intersections and straight road segments, are summarized in Table \ref{tab:road_scenarios}.

\begin{table*}[t]
\centering
\caption{Overview of the evaluation environments for real-world and simulation scenarios.}
\label{tab:evaluation_scenarios}
\begin{tabular}{cl|p{4.8cm}|p{3,5cm}|p{3cm}}
\toprule
\multicolumn{2}{c|}{\textbf{Scenario}} & \textbf{Urban} & \textbf{Rural} & \textbf{Highway} \\ \midrule
\multirow{3}{*}{\textbf{Real-world}} 
& V2V & 
\cite{LUCOOPLeibnizUniversity-2023-axmanna}, \cite{V2V4RealRealWorldLargeScale-2023-xua}, \cite{FcooperFeatureBased-2019-chend} & 
\ NA  & 
\ NA  \\
& V2I & 
\cite{NovelMultiviewPedestrian-2020-benkhalifaa}, \cite{DAIRV2XLargeScaleDataset-2022-yua}, \cite{HoloVICLargescaleDataset-2024-mac}, \cite{TUMTrafV2XCooperative-2024-zimmerc}, \cite{V2XSeqLargeScaleSequential-2023-yub}, \cite{MKDCooperCooperative3D-2024-lia}, \cite{SoftActorCriticBased-2022-xiea} & 
\ NA & 
\ NA  \\
& Both &
 \ NA  & 
 \ NA  & 
 \ NA \\ 
\midrule
\multirow{3}{*}{\textbf{Simulation}} 
& V2V & 
\cite{OPV2VOpenBenchmark-2022-xub}, \cite{ExtensibleFrameworkOpen-2024-luc}, \cite{CollaborativeSemanticOccupancy-2024-songb}, \cite{LearningDistilledCollaboration-2021-lib}, \cite{AsynchronyRobustCollaborativePerception--weic}, \cite{AdaptiveFeatureFusion-2023-qiaob} & 
\cite{OPV2VOpenBenchmark-2022-xub}, \cite{ExtensibleFrameworkOpen-2024-luc}, \cite{CollaborativeSemanticOccupancy-2024-songb}, \cite{LearningDistilledCollaboration-2021-lib}, \cite{AdaptiveFeatureFusion-2023-qiaob} & 
\cite{OPV2VOpenBenchmark-2022-xub}, \cite{ExtensibleFrameworkOpen-2024-luc}, \cite{CollaborativeSemanticOccupancy-2024-songb}, \cite{LearningDistilledCollaboration-2021-lib} \\
& V2I & 
\ NA & 
\ NA & 
\ NA \\
& Both & 
\cite{DeepAccidentMotionAccident-2024-wanga}, \cite{V2XSimMultiAgentCollaborative-2022-lie}, \cite{V2XViTVehicletoEverythingCooperative-2022-xub} & 
\cite{V2XSimMultiAgentCollaborative-2022-lie}, \cite{V2XViTVehicletoEverythingCooperative-2022-xub} & 
\cite{V2XSimMultiAgentCollaborative-2022-lie}, \cite{V2XViTVehicletoEverythingCooperative-2022-xub} \\ 
\bottomrule
\end{tabular}
\end{table*}

\begin{table}[t]
\centering
\caption{Overview of Road Environment for Evaluation}
\label{tab:road_scenarios}
\begin{tabular}{l|p{4cm}}
\toprule
\textbf{Road Environment} & \textbf{References} \\ \midrule
Roundabouts & \cite{NovelMultiviewPedestrian-2020-benkhalifaa} \\
Straight roads with curves & \cite{NovelMultiviewPedestrian-2020-benkhalifaa, HoloVICLargescaleDataset-2024-mac, LUCOOPLeibnizUniversity-2023-axmanna, V2V4RealRealWorldLargeScale-2023-xua, V2XSeqLargeScaleSequential-2023-yub, MKDCooperCooperative3D-2024-lia} \\ 
Cross intersection & \cite{DAIRV2XLargeScaleDataset-2022-yua, HoloVICLargescaleDataset-2024-mac, TUMTrafV2XCooperative-2024-zimmerc, V2XSeqLargeScaleSequential-2023-yub, MKDCooperCooperative3D-2024-lia, SoftActorCriticBased-2022-xiea} \\ 
T-Junction & \cite{LUCOOPLeibnizUniversity-2023-axmanna, V2XSeqLargeScaleSequential-2023-yub} \\ 
Parking lots & \cite{SoftActorCriticBased-2022-xiea, FcooperFeatureBased-2019-chend} \\ 
\bottomrule
\end{tabular}
\end{table}

\subsubsection{Daytime and Weather}
Robust evaluation of \gls{CP} algorithms requires testing under diverse conditions to assess their performance in challenging scenarios, such as low-light environments or adverse weather conditions. Many real-world datasets incorporate both daytime and nighttime data \cite{LUCOOPLeibnizUniversity-2023-axmanna, TUMTrafV2XCooperative-2024-zimmerc, V2XSeqLargeScaleSequential-2023-yub}, although not all studies provide explicit documentation of these conditions \cite{HoloVICLargescaleDataset-2024-mac, V2V4RealRealWorldLargeScale-2023-xua}. Regarding weather diversity, detailed descriptions are frequently omitted. Among real-world datasets, DAIR-V2X \cite{DAIRV2XLargeScaleDataset-2022-yua} stands out for its inclusion of varying weather and lighting conditions, establishing it as the most comprehensive dataset in this regard.

In contrast, synthetic datasets offer complete control over environmental parameters such as weather and time of day. However, these conditions are rarely detailed in the associated studies. An exception is the DeepAccident dataset \cite{DeepAccidentMotionAccident-2024-wanga}, which explicitly provides variations in weather conditions (e.g., clear, rainy, cloudy, wet) and times of day (e.g., noon, sunset, night). This level of specification enhances its utility for evaluating CP algorithms under diverse environmental settings.

\subsection{Evaluation Metrics} \label{sec:eval_metrics}
\noindent To quantitatively assess perception performance, various metrics are applied to specific tasks such as object detection, tracking, and motion prediction, depending on the evaluation objectives. Unified evaluation metrics are crucial for benchmarking different algorithms, enabling comparative analysis of their performance, and supporting the continuous improvement of these algorithms.

This section reviews and summarizes the metrics used for evaluating \gls{CP}. These metrics are categorized into two groups: general evaluation metrics, which are adapted from single-entity perception tasks, and custom metrics designed for \gls{CP}. Additionally, this section provides a summary of the ablation studies conducted in the reviewed papers. These studies offer insights into the evaluation process, highlighting common factors that impact \gls{CP} and how they influence performance. This understanding aids researchers in designing more practical and robust \gls{CP} frameworks.

\subsubsection{General Evaluation Metrics for Perception Tasks}
The general evaluation metrics for different perception tasks are summarized in the Table \ref{tab:eval_metrics}. These metrics are adapted from single-entity perception and are widely accepted by researchers. In some cases, evaluation results are divided into different groups based on detection difficulty levels, such as easy, medium, and difficult, as seen in the KITTI dataset \cite{6248074}, which considers factors like occlusion level and object size. Evaluation results can also be categorized by object type, such as cars, cyclists, and pedestrians. This categorization helps researchers better understand the strengths and limitations of different approaches.


\begin{table}[t]
\centering
\caption{Overview of general evaluation metrics for collaborative perception tasks}
\label{tab:eval_metrics}
\begin{tabular}{c|p{5cm}}
\toprule
Task & Metrics \\ \midrule
Object Detection & \textbf{mean Average Precision (mAP)}, Average Precision (AP),  Recall, Precision, Average Recall (AR), mean Average Orientation Similarity (mAOS) \\ \midrule
Object Tracking & \textbf{AMOTA}, \textbf{AMOTP}, sAMOTA, MOTA, MOTP, HOTA, Recall, FP, FN, ID F1 Score, Mostly Tracked Trajectories (MT), Mostly Lost Trajectories (ML), Negative Log Likelihood (NLL), Continuous Ranked Probability Score (CRPS) \\ \midrule
Prediction & \textbf{minADE}, \textbf{minFDE}, Video Panoptic Quality (VPQ), Accident Prediction Accuracy(APA), L2 Displacement Error, End-Point Error (EPE), strict/ relaxed accuracy (AccS/ AccR), outlier ratio (ROutliers), and Missing Rate (MR) \\ \midrule
Semantic Segmentation & \textbf{Mean Intersection over Union (mIoU)}, Intersection over Union (IoU)  \\ \midrule
Lane Detection & \textbf{Mean Squared Error MSE}, Maximum Error, Root Mean Square Error (RMSE), Mean Absolute Error (MAE), Intersection over Union (IoU) \\ 
\bottomrule
\end{tabular}
\end{table}

\subsubsection{Custom Evaluation Metrics for Collaborative Perception}
Traditional evaluation metrics used for single-entity perception do not adequately represent the performance of cooperative perception (\gls{CP}). Since \gls{CP} primarily aims to address visual occlusion problems and serves as a supplement to single-entity perception, it requires distinct evaluation criteria. Moreover, \gls{CP} is significantly constrained by communication resources. Therefore, communication factors should be incorporated into the design of evaluation metrics. In this subsection, we summarize custom metrics designed for \gls{CP} in Table \ref{tab:eval_metrics_specific}. These metrics are classified into the following three types.

\begin{table*}[htbp]
\centering
\caption{Overview of custom evaluation metrics for collaborative perception tasks}
\label{tab:eval_metrics_specific}
\begin{tabular}{p{3cm}|p{1.5cm}p{8cm}} 
\toprule
\textbf{Aspect} & \textbf{Metrics} & \textbf{Description} \\ \midrule
\multirow{2}{*}{\parbox{3cm}{Communication}} 
    & Average message size
    & Measures the communication cost in units such as Byte, KB, MB, or Mbps. \\ \midrule

\multirow{3}{*}{\parbox{3cm}{Communication and perception improvement}} 
    & BIS
    & Bandwidth Improvement Score (BIS): A ratio of relative improvement in overall accuracy over bandwidth usage. Smaller bandwidth usage and larger improvements in overall accuracy lead to higher scores. \\ \cline{2-3}
    & AIB
    & Average-precision-improvement-to-bandwidth-usage (AIB): evaluate the trade-off between detection performance improvement and bandwidth usage of the proposed framework. \\ \cline{2-3}
    & RB Ratio 
    & Recall/Bandwidth (RB) Ratio: Evaluates bandwidth efficiency relative to recall performance. \\ \midrule

\parbox{3cm}{Perception improvement} 
    & Marginal Gain 
    & Measures the the performance increase when an additional agent joins the collaboration. \\ \midrule

\multirow{4}{*}{\parbox{3cm}{Perception performance of invisible agents}} 
    & ARSV 
    & Average Recall of agents visible from Single-Vehicle View. \\ \cline{2-3}
    & ARCV 
    & Average Recall of agents invisible from Single-Vehicle View but visible from Collaborative-View. \\ \cline{2-3}
    & ARCI 
    & Average Recall of Completely-Invisible agents. \\ \cline{2-3}
    & ARTC 
    & Average Recall of agents visible previously but occluded at present \\ \bottomrule
\end{tabular}
\end{table*}

\begin{itemize}
\item{{\bf{Communication}}: Compared to single-entity perception, cooperative perception requires additional communication resources. Evaluating the communication demands of \gls{CP} algorithms is crucial for assessing their efficiency and scalability. For example, metrics such as average message size are commonly used to measure the communication costs associated with \gls{CP}.}

\item{{\bf{Perception}}: \gls{CP} aims to address visual occlusion problems, making it crucial to have metrics that assess how effectively \gls{CP} resolves these issues. Wang et al. \cite{UMCUnifiedBandwidthefficient-2023-wangb} introduce the Average Recall of Collaborative View (ARCV) metric, which measures the average recall of agents that are invisible from a single-vehicle perspective but become detectable through collaborative perception. In addition to uncovering occluded agents, \gls{CP} can enhance the perception of agents already visible to a single vehicle by incorporating additional information. To quantify this enhancement, Luo et al. \cite{ComplementarityEnhancedRedundancyMinimizedCollaboration-2022-luob} propose the marginal gain metric, defined as the performance improvement when an additional agent joins the collaboration. It is important to note that the marginal gain tends to diminish as more observing agents are added.}

\item{{\bf{Ratio between Communication and Perception}}: There is an inherent trade-off between communication cost and collaborative perception (\gls{CP}) performance. Reducing communication costs can constrain \gls{CP} effectiveness by limiting the amount of shared information among agents. Researchers are exploring how to balance these factors to develop efficient and effective \gls{CP} approaches. For instance, Liu et al. \cite{Who2comCollaborativePerception-2020-liua} introduce the Bandwidth Improvement Score (BIS), defined as the ratio of the relative improvement in overall accuracy to the bandwidth usage. A higher BIS indicates a more favorable balance, lower bandwidth cost coupled with greater improvement in perception performance.}

\end{itemize}

The custom metrics for collaborative perception place greater emphasis on improving the detection of both visible and previously invisible objects from the ego vehicle's viewpoint. However, these perception improvement metrics have not been widely accepted in the research community. Most studies predominantly utilize evaluation metrics adopted from single-entity perception, with the exception of studies \cite{UMCUnifiedBandwidthefficient-2023-wangb,Who2comCollaborativePerception-2020-liua,Collaborative3DObject-2023-wanga,SlimFCPLightweightFeatureBasedCooperative-2022-guoa}, which employ custom metrics for \gls{CP}. Communication cost metrics are also occasionally considered when evaluating the efficiency of collaborative perception methods.

\subsubsection{Ablation Studies}
Ablation studies are crucial for evaluating the robustness and scalability of \gls{CP} systems under various conditions. They help identify how different factors affect \gls{CP} performance, enabling researchers to optimize system design. In this section, we categorize and discuss ablation studies focusing on communication, localization, system scale, visual occlusion, adversarial attacks, and other relevant factors.

\begin{table}[t]
\centering
\caption{Ablation studies grouped by Aspect}
\begin{tabular}{p{2.5cm}|p{5cm}}
\toprule
\textbf{Aspect} & \textbf{Ablation studies} \\ \midrule
Communication
    & Bandwidth, Latency, Packet drop rate, Communication noise (SNR in dB), Probability of interruption, Compression rate \\ \midrule
Localization
    & Pose error, Position error, Heading error \\ \midrule
Scale of system
    & Number of CAVs, CAV rate, Ratio of Lidar/camera-equipped agents \\ \midrule
Visual occlusion
    & Occlusion level \\ \midrule
Adversarial attack & Attack ratio \\ \midrule
Others 
    & Object distance, Traffic density, Object speed, Ego vehicle velocity and acceleration, Number of cameras dropped \\
\bottomrule
\end{tabular}
\end{table}

\begin{itemize}
\item{{\bf{Communication}}: Communication poses significant challenges for \gls{CP} in real-world applications. Vehicle-to-everything (V2X) communication introduces practical constraints such as bandwidth limitations, data compression requirements, latency, noise, and interruptions. Most research includes ablation studies addressing these communication issues \cite{Who2comCollaborativePerception-2020-liua,JointPerceptionScheme-2022-ahmeda,Where2commCommunicationEfficientCollaborative-2022-huc,CooperativePerceptionSystem-2023-songc,CollaborationHelpsCamera-2023-hub,Collaborative3DObject-2023-wanga,CooperativePerceptionLearningBased-2023-liub,LearningVehicletoVehicleCooperative-2023-lia,RobustRealtimeMultivehicle-2023-zhangb}. It has been proven that latency and noise can significantly degrade \gls{CP} performance, while interruptions have the most severe impact. Designing communication-aware approaches is essential for enhancing the scalability and effectiveness of \gls{CP} applications.}

\item{{\bf{Localization}}: Localization errors heavily affect the performance of \gls{CP} systems. To ensure algorithms are viable in real-world settings, it is crucial to measure their robustness against positioning errors. Most studies validate algorithm performance under varying positioning errors, typically ranging from 0 to 1 meter \cite{Who2comCollaborativePerception-2020-liua,CooperativePerceptionSystem-2023-songc,FeaCoReachingRobust-2023-guc,ExtensibleFrameworkOpen-2024-luc,EMIFFEnhancedMultiscale-2024-wangc,HP3DV2VHighPrecision3D-2024-chena,QUESTQueryStream-2023-fand,RegionBasedHybridCollaborative-2024-liu,V2VFormerMultiModalVehicletoVehicle-2024-yina,RobustCollaborativePerception--rena}. The absence of significant performance degradation under these conditions demonstrates the algorithm's reliability in practical applications.}

\item{{\bf{Scale of system}}: The performance of \gls{CP} varies with the number and types of CAVs involved. Ablation studies are helpful in determining the optimal configuration of cooperative systems. Research has shown that \gls{CP} achieves the best results when 4 to 6 agents participate in the collaborative system; adding more agents does not further increase perception accuracy \cite{AdaptiveFeatureFusion-2023-qiaob,PracticalCollaborativePerception-2024-daoa,RegionBasedHybridCollaborative-2024-liu,V2VFormerVehicletoVehicleCooperative-2024-lina}. Additionally, the types of CAVs, such as those equipped with LiDAR or cameras, also influence \gls{CP} performance \cite{HMViTHeteromodalVehicletoVehicle-2023-xiangb}. Validating \gls{CP} under different ratios of LiDAR-equipped and camera-equipped CAVs is important to ensure robustness.}

\item{{\bf{Visual occlusion}}: Verifying \gls{CP}'s reliability in detecting occluded objects requires ablation studies that consider different levels of occlusion. These studies demonstrate the effectiveness of \gls{CP} in addressing visual occlusion and indicate how well the system performs under such conditions \cite{DistributedDynamicMap-2021-zhanga,CollaborativeMultiObjectTracking-2024-sua}.}

\item{{\bf{Adversarial attack}}: Robustness against adversarial attacks is a critical aspect of \gls{CP} systems. Ablation studies focusing on attack scenarios verify whether \gls{CP} can maintain reliability under various adversarial conditions \cite{UsAdversariallyRobust-2023-lia,RobustCollaborativePerception-2024-lei}. Ensuring resilience to such attacks is vital for the safe deployment of \gls{CP} systems.}

\item{{\bf{Others}}: Additional factors can affect \gls{CP} performance, such as traffic density, vehicle velocity and speed, and sensor dropout. Conducting diverse ablation studies under different scenarios ensures the system's reliability in real-world 
usage~\cite{RobustRealtimeMultivehicle-2023-zhangb, CoBEVTCooperativeBird-2023-xuc}. 
By validating \gls{CP} performance across these variables, researchers can develop more robust and adaptable systems.}

\end{itemize}

Various ablation studies have been conducted to assess the reliability and robustness of \gls{CP} systems. However, performing comprehensive ablation studies is time-consuming and resource-intensive. Researchers should prioritize validating factors that are most pertinent to the specific problems their work aims to address. To simplify the evaluation process, an automated evaluation framework is needed.

While ablation studies are valuable for measuring \gls{CP} performance, they may not always yield accurate results due to the interplay of multiple influencing factors, such as the number of \gls{CAVs} and communication bandwidth. To ensure reliable validation, it is important to conduct online evaluations using simulations or real-world experiments. These methods can address the interdependencies of various factors, filling the gap left by traditional ablation studies.

%% file: sections/05_challenges_opportunities_risks.tex
\section{Challenges, opportunities, and risks (RQ5)} \label{sec:challenges}
\noindent \acrlong{CP} holds significant potential to extend the perception range of individual vehicles and address critical scenarios caused by occlusion. However, implementing this technology in real-world applications faces numerous challenges. Based on the comprehensive analysis of \gls{CP}, this section introduces the challenges, opportunities, and risks associated with \gls{CP} research.

We examine the challenges and opportunities from three perspectives: hardware, software, and evaluation methods. The risks in \gls{CP} research are summarized concerning application gaps, reproducibility, and evaluation. Each aspect provides insight into the current state of \gls{CP} and highlights areas for future improvement.

\subsection{Challenges}

\subsubsection{Hardware}
\gls{CAVs} employ a variety of sensors, each with its advantages and limitations. These vehicles are typically equipped with multiple sensors, such as LiDAR and cameras, to navigate diverse driving scenarios effectively.  Integrating these sensors enhances the capabilities of multi-modality in \gls{CP}, allowing vehicles to perceive their environment more comprehensively and share multi-modal information with nearby vehicles. However, achieving precise time synchronization and calibration among multiple sensors is challenging. Multi-modal \gls{CP} methods rely on accurate spatial and temporal alignment from different sensors, but factors like sensor drift and environmental variability make consistent precision challenging to maintain. This inconsistency can hinder the full potential of multi-modal approaches. To fully harness the advantages of multi-modality, it is essential to develop efficient calibration methods for multi-sensor systems, not only on the vehicles themselves but also within the supporting infrastructure \cite{9732063}. Addressing these calibration challenges will enhance the reliability and effectiveness of \gls{CP} in real-world applications.

\subsubsection{Software}
While hardware challenges such as sensor calibration are significant, various software challenges also exist and are the main focus of this literature review. In the following sections, we discuss these software challenges from two critical aspects, communication and perception, which together form the core technologies of cooperative perception.

\begin{itemize}
\item{{\bf{Communication}}: V2X communication enables data transmission between entities but comes with certain constraints. As discussed in Section \ref{sec:cp_issues}, the communication challenges involved in \gls{CP} are significant. Bandwidth limitations and communication range constraints are primary considerations when designing a \gls{CP} framework. To prevent network congestion, the framework must minimize bandwidth demands. In addition to communication efficiency, the robustness of the \gls{CP} framework against latency, data loss, and interruptions is crucial for maintaining reliable perception. By addressing these real-world communication factors, we can effectively implement \gls{CP} technology in practical applications, enhancing perception in critical scenarios and improving the safety of \gls{CAVs}. However, only one study, V2X-INCOP~\cite{InterruptionAwareCooperativePerception-2024-renc}, has specifically addressed communication interruptions. Research on robust \gls{CP} under realistic communication conditions remains significantly under-explored. In addition to addressing communication constraints, the standardization of \gls{V2X} protocols presents significant challenges for early and intermediate collaboration approaches. Currently, standardized \gls{CP} exclusively support late collaboration. Exploring effective methods to transmit raw sensor data and intermediate features within the framework of realistic communication protocols remains a critical area of investigation.}

\item{{\bf{Fusion strategy in Perception}}: Information fusion among agents is central to \gls{CP}, enabling a collective understanding of the environment. However, several challenges persist in developing efficient and robust fusion methods. Firstly, information loss is a significant concern in data fusion. Techniques such as late fusion, which combine perception results using bounding boxes, often discard crucial texture information. Traditional feature fusion methods, such as average pooling, may overlook detailed features from different agents. These fine-grained details are essential for accurate scene understanding. To overcome these limitations, exploring efficient data fusion methods that retain essential information for downstream tasks without substantially increasing communication costs is necessary. Secondly, the growing volume and variety of data shared among agents introduce challenges in data management and resource allocation. Novel hybrid fusion methods that utilize features and perception results can enhance cooperation between agents, such as Hybrid-CP~\cite{RegionBasedHybridCollaborative-2024-liu}. However, the inclusion of diverse data types substantially increases the complexity of data management. Managing heterogeneous data from multiple agents poses a significant challenge, necessitating targeted solutions. Lastly, data alignment remains a bottleneck in the real-world application of \gls{CP} systems. Spatial alignment issues are not fully resolved; most current approaches are only robust against positioning errors within one meter~\cite{LatencyAwareCollaborativePerception-Lei-22,UMCUnifiedBandwidthefficient-2023-wangb,AsynchronyRobustCollaborativePerception--weic,FlowBasedFeatureFusion-2023-yue,How2commCommunicationEfficientCollaborationPragmatic--yanga,V2XSeqLargeScaleSequential-2023-yub,InterruptionAwareCooperativePerception-2024-renc}. In practice, the localization error of CAVs can vary by several meters. Achieving higher robustness is essential to scale \gls{CP} solutions across diverse conditions. For instance, resolving the positional alignment of visual features extracted from different agents' cameras remains an unsolved problem. Temporal alignment is another crucial factor. Aligning asynchronous features is particularly challenging because it often relies on predicting future features, which can constrain the accuracy of the alignment.}
\item{{\bf{Robustness}}: Ensuring the robustness of \gls{CP} systems is crucial for autonomous driving applications, which are inherently safety-critical. Various factors can degrade the performance of \gls{CP} systems. As previously discussed, issues with communication and localization significantly affect performance, making it essential to enhance the robustness of \gls{CP} systems against these challenges. In addition to these factors, challenging scenarios present further critical issues. For example, collaborative lane detection performance in complex road structures deteriorates because current models may not be sufficiently robust to infer intricate road geometries accurately in real-time \cite{CooperativeRoadGeometry-2020-sakra,CoLDFusionRealtime-2023-gamerdingera,EnhancingLaneDetection-2024-jahna}. Similarly, collaborative object detection performance declines in dense traffic conditions due to increased occlusion and the constrained bandwidth available to each CAV in the area. Although \gls{CP} has not yet been extensively evaluated under adverse weather conditions, performance is expected to degrade similar to single-entity perception systems. Therefore, enhancing the robustness of \gls{CP} systems across diverse scenarios is imperative. Beyond environmental challenges, adversarial attacks pose another significant threat to \gls{CP}. With increasing connectivity between vehicles, infrastructure, and cloud services, protecting autonomous vehicles from network attacks becomes more critical. \gls{CP} systems should be capable of identifying fraudulent messages and avoiding the fusion of malicious data to maintain system reliability. Only one study, AmongUs~\cite{UsAdversariallyRobust-2023-lia}, implement the detection of malicious activities within the \gls{CP} framework and evaluated their impact on perception performance. This remains a significantly under-explored area.}
\item{{\bf{Uncertainty}}: \gls{CP} relies heavily on \gls{AI}, which often functions as a "black box" due to its lack of explainability. This opacity makes it difficult to determine absolute confidence in the perception results. \gls{CAVs} from different manufacturers may use diverse perception models with varying performance levels. In \gls{CP} systems, receivers obtain processed data from senders -- such as detected objects -- but assessing the uncertainty associated with this data is challenging. This situation raises the issue of trustworthiness: to what extent should \gls{CAVs} trust the information received from others? Beyond uncertainties in perception results, there are inherent uncertainties within the systems. For example, depth estimation using cameras for 3D perception introduces uncertainty, as do upstream tasks whose errors can accumulate throughout the processing pipeline, ultimately degrading the outcome. To enhance the reliability and explainability of \gls{CP} systems, it is important to design uncertainty-aware models that can learn to process noisy data effectively.}
\item{{\bf{Efficiency}}: Autonomous driving systems are computationally intensive platforms that process large amounts of data in real-time. Perception is one of the most resource-consuming modules, heavily relying on complex neural networks. Compared to single-entity perception, \gls{CP} demands even more computing and communication resources. The trade-off between improved perception and additional resource consumption is a critical factor in determining the scalability of \gls{CP} systems for real-world applications. Furthermore, real-time performance necessitates that \gls{CP} systems achieve computational and communication efficiency. This ensures that accurate information is transmitted promptly to downstream modules such as planning and control. Balancing these demands is essential for effectively deploying \gls{CP} in practical autonomous driving scenarios.}
\item{{\bf{Domain shift}}: Domain shift presents a significant challenge in \gls{CP} among agents equipped with different types of LiDAR sensors. Features extracted from various LiDAR systems do not reside in the same feature space, meaning this discrepancy can significantly degrade system's performance \cite{DIV2XLearningDomainInvariant-2023-xiangb,HPLViTUnifiedPerception-2023-liuc}. To address this issue, bridging the gap between the source and target domains is essential before fusing features from multiple agents. Beyond its impact on feature fusion, the simulation-to-real-world (Sim2Real) domain shift also causes models trained on synthetic datasets to perform poorly in real-world environments~\cite{DUSADecoupledUnsupervised-2023-kongc,S2RViTMultiAgentCooperative-2024-lib}. Collecting real-world data is costly and particularly difficult for safety-critical scenarios. As a result, researchers and developers are seeking cost-efficient solutions for training neural networks using synthetic data. However, the pronounced gap between simulation and reality makes achieving this challenging. To increase the utilization of synthetic data in \gls{CP}, it is urgent to bridge the Sim2Real gap, facilitating the transfer of knowledge learned from simulations to real-world applications. This advancement would also enable training models with synthetic safety-critical data, filling current gaps in available training datasets.}
\item{{\bf{Heterogeneity}}: Research on \gls{CP} often assumes unrealistic conditions to simplify the complexity of collaborative systems, particularly regarding the heterogeneity of agents. This heterogeneity includes differences in models (model heterogeneity) \cite{BridgingDomainGap-2023-xua} and sensing modalities (modality heterogeneity) \cite{MultiModalVirtualRealFusion-2022-zhanga,HMViTHeteromodalVehicletoVehicle-2023-xiangb,ExtensibleFrameworkOpen-2024-luc}. Embracing heterogeneous collaboration is essential for making \gls{CP} technology applicable in industry and deployable in real-world scenarios. However, current approaches that address heterogeneity are limited and struggle to maintain the reliability of \gls{CP} systems under heterogeneous conditions.}
\item{{\bf{Model training}}: Model training is a crucial step in developing \gls{CP} algorithms. As models increase in size and complexity, they require larger datasets for effective training. To reduce costs, it is important to decrease the dependence on labeled data in \gls{CP}, which can significantly reduce the effort required for data annotation~\cite{CollaborativePerceptionAutonomous-2023-hanc}.}
\end{itemize}

\subsubsection{Evaluation}
The evaluation of \gls{CP} systems presents several challenges, ranging from the methods used to the metrics applied. The challenges related to evaluation, as identified in the literature, are summarized below.
\begin{itemize}

\item{{\bf{Lack of large-scale real-world dataset}}: AI-driven perception algorithms require large-scale and diverse datasets to learn the patterns and features necessary for robust model generalization. However, the current datasets available for \gls{CP} research are insufficient in size and lack diversity. They do not adequately cover a range of scenarios, such as different weather conditions or critical traffic situations. Additionally, existing datasets primarily support collaborative object detection, tracking, and prediction, but there are no datasets for tasks like collaborative semantic segmentation or lane detection. To advance \gls{CP} research forward, creating large-scale, multi-modal datasets that support multiple tasks across diverse scenarios is essential. Creating a real-world dataset presents several challenges that must be addressed in advance. Data privacy concerns and the processes required to ensure compliance can be time-intensive.  In particular, visual data captured by cameras must undergo anonymization to obscure identifiable features such as human faces and vehicle license plates, ensuring adherence to data privacy regulations. Hardware setup poses additional difficulties, particularly in achieving precise time synchronization and localization for the vehicles involved. Moreover, generating diverse annotations, covering supported tasks, object classes, or supplementary details such as occlusion, demands substantial time and financial resources. Finally, as these datasets are typically recorded at real-world intersections rather than controlled test fields, managing class distribution becomes challenging due to the lack of control over traffic conditions.}

\item{{\bf{Simulation for evaluation}}: To further advance \gls{CP} algorithms, designing fair and goal-oriented evaluation methods that quantitatively measure their performance is essential. Beyond benchmarking on public datasets, conducting online evaluations in simulations that consider more realistic network conditions can provide deeper insights. However, integrating realistic communication models into co-simulation frameworks remains challenging due to bottlenecks between multiple simulation platforms.}

\item{{\bf{Scenarios for evaluation}}: \gls{CP} is designed to address critical occlusion situations to enhance the safety of CAVs. However, collecting data on these critical scenarios from real-world environments is challenging. Such situations are rare in daily traffic and pose significant risks during data collection. Consequently, there is a gap in validating the reliability of \gls{CP} in safety-critical scenarios, which is crucial for its improvement. Developing effective methods to evaluate \gls{CP} under these conditions remains an essential yet unresolved challenge.}

\item{{\bf{Evaluation metrics}}: Metrics are essential for quantitatively analyzing the performance of \gls{CP} methods. Therefore, it is crucial to design appropriate metrics that clearly represent the advantages and limitations of \gls{CP}. Most existing metrics are adapted from single-entity perception, which may not fully capture the unique benefits of \gls{CP}, especially in addressing occlusion. Notably, only one study, UMC \cite{UMCUnifiedBandwidthefficient-2023-wangb}, evaluates performance specifically on occluded objects. To effectively evaluate \gls{CP}'s efficiency in solving occlusion problems, new metrics need to be developed, introducing novel criteria for quantitative assessment. Beyond performance measurement, there is also a need for metrics that evaluate effectiveness and safety-related aspects. Developing such metrics will enable a more comprehensive and quantitative analysis of \gls{CP} systems, facilitating their improvement and real-world application.}

\item{{\bf{Ablation study}}: Researchers employ ablation studies to evaluate the efficiency of \gls{CP} under various conditions. However, conducting these studies is more labor-intensive than in single-entity perception due to the diverse factors affecting \gls{CP}, including communication, localization, and perception. To accelerate future research, it is essential to develop tools that enable the automatic execution of ablation studies.}

\end{itemize}

\subsection{Opportunities}
\noindent We have outlined corresponding opportunities and future directions by identifying the open challenges and research gaps in \gls{CP}. These are summarized from three critical perspectives: hardware, software, and evaluation.

\subsubsection{Hardware}
\begin{itemize}

\item {{\bf{Optimal sensor configuration}}: Optimizing sensor configurations is a significant opportunity in the hardware aspect of \gls{CP}.  Researchers have ample potential to explore which hardware setups are most effective for \gls{CP} systems. Determining the optimal types and placements for infrastructure sensors is particularly important. The design and positioning of sensors at intersections directly impact roadside perception performance \cite{cai2023analyzing}. Investigating the trade-offs between sensor redundancy and safety is also valuable for enhancing system reliability.}

\item {{\bf{New modality}}: Another area for advancement is the integration of new sensor modalities. Current \gls{CP} frameworks predominantly use LiDAR and cameras to perceive the environment. However, the application of radar, infrared cameras, or event cameras remains largely unexplored. Radars can provide more accurate velocity measurements, while infrared cameras offer night vision capabilities. Incorporating these sensors can enhance the robustness of perception systems by supplementing the limitations of LiDAR and standard cameras.}

\end{itemize}

\subsubsection{Software}

\begin{itemize}

\item {{\bf{Communication}}: Enhancing communication efficiency is a critical opportunity in the software aspect of \gls{CP}. Since communication is fundamental to these systems, improving it can significantly boost overall performance. One approach is implementing data compression techniques that reduce message sizes without substantial information loss. This applies to raw data and feature data, enabling the transmission of more valuable information and enhancing the \gls{CP} process. Additionally, exploring the transmission of various data types, maps and historical perception data, can diversify \gls{CP} solutions. Designing efficient data structures for these heterogeneous data types is crucial for real-world applications. Implementing contributor selection strategies can also reduce unnecessary connections and data redundancy within the collaborative framework \cite{When2comMultiAgentPerception-2020-liua,Who2comCollaborativePerception-2020-liua,Collaborative3DObject-2023-wanga}.}

\item {{\bf{Fusion strategy in \gls{CP}}}: Advancing fusion strategies presents another promising direction. Hybrid fusion methods have shown potential in balancing resource consumption with perception performance by dynamically adapting to communication conditions, thus ensuring scalability \cite{PillarBasedCooperativePerception-2022-wanga,SoftActorCriticBased-2022-xiea,KeypointsBasedDeepFeature-2022-yuanc,RegionBasedHybridCollaborative-2024-liu,RobustCollaborativePerception-2024-lei}. Further research into hybrid fusion could unlock more benefits for \gls{CP}. Graph-based feature fusion is an underexplored area that merits attention. Graph Neural Networks (GNNs) can model agent relationships and adjust collaborations based on changing environments. Investigating the application of GNNs in communication and perception could yield significant advancements.}

\item {{\bf{Robustness and efficiency}}: Improving robustness and efficiency is vital for the practical deployment of \gls{CP} systems. While issues like communication disruptions and adversarial attacks have been studied, hardware failures, such as sensor dropouts, have been largely overlooked. Enhancing robustness against sensor failures is important for ensuring system reliability. On the efficiency front, although many state-of-the-art algorithms achieve high accuracy on public datasets, their performance concerning hardware limitations has not been thoroughly investigated. Exploring ways to improve algorithmic efficiency, such as optimizing models like V2X-ViT \cite{V2XViTVehicletoEverythingCooperative-2022-xub}, would be a valuable direction for future research. }

\item {{\bf{Compatibility}}: Lastly, ensuring compatibility between \gls{CP} and ego perception systems is essential. \gls{CP} should supplement, not replace, individual perception capabilities. Current research often treats \gls{CP} as a separate system, leading to potential resource wastage. Developing perception pipelines that can operate independently without shared information and collaborate with other agents when necessary would make systems more practical. Designing \gls{CP} as a plug-and-play module can avoid the need for complete redesign and retraining of perception models, enhancing efficiency and adaptability.}

\end{itemize}

\subsubsection{Evaluation}
Evaluation methods are essential guidelines for researchers and developers seeking to improve \gls{CP} performance. However, current evaluation approaches have notable drawbacks. This section outlines opportunities to enhance \gls{CP} evaluation from several perspectives.

\begin{itemize}

\item {{\bf{Datasets}}: A significant opportunity lies in developing large-scale \gls{CP} datasets to advance research. Besides size, diversity in datasets is equally important. Incorporating various sensor modalities and perception tasks can enrich future \gls{CP} datasets. While annotating real-world data is costly, researchers might provide unlabeled data to the community to foster collaboration. Creating an open-source framework for generating synthetic \gls{CP} datasets would also be highly beneficial, enabling broader participation and innovation.}

\item {{\bf{Evaluation methods}}: To bridge the deployment gap in \gls{CP} systems, developing a framework simplifying the entire lifecycle, from research and development to deployment and testing, is crucial. Such a framework should accelerate validation and evaluation with datasets and in real-world conditions. By streamlining these processes, \gls{CP} can more readily transition into practical applications.}

\item {{\bf{Evaluation scenarios}}: Validating the reliability of \gls{CP} under diverse conditions requires collecting a more comprehensive range of evaluation scenarios. Expanding the diversity of these scenarios ensures that \gls{CP} systems are robust across different environments. Particular attention should be given to critical traffic situations, such as those involving vulnerable road users, to assess system's performance in high-risk contexts thoroughly.}

\item {{\bf{Metrics and ablation studies}}: Quantitative metrics are important for accurately measuring \gls{CP} performance. As discussed in the challenges, new metrics that align with \gls{CP}'s objectives, such as resolving visual occlusions, are needed. Beyond developing new metrics, creating a framework that enables automatic ablation studies under varied conditions would provide valuable insights. Such a framework can help researchers understand the impact of different components and configurations, ultimately leading to more effective \gls{CP} systems.}
\end{itemize}

\subsection{Risks}
\noindent While \gls{CP} technology shows great promise in enhancing the capabilities of \gls{CAVs} and improving road safety, it also faces several significant risks.

\begin{itemize}

\item {{\bf{Deployment gap}}: A significant risk is the gap between research advancements and real-world deployment. Although various studies address individual aspects of this gap, the complexities of real-world conditions far exceed those modeled in datasets or simulations. For instance, limited communication bandwidth caused by environmental factors, unexpected synchronization failures, as well as unpredictable communication latency and interruptions can adversely affect \gls{CP} performance.  Additionally, striking the right balance between the perceptual improvements gained through collaboration and the additional costs incurred is challenging. Successful deployment of \gls{CP} also requires collaboration among different vendors to ensure that \gls{CAVs} from various manufacturers can communicate effectively and have compatible perception modules.}

\item {{\bf{Reproducibility}}: Another critical concern is the reproducibility of research findings. The research community must verify results and build upon previous work to ensure reproducibility. However, the limited availability of accessible repositories and source code in \gls{CP} research hampers this process. Providing open-source code and datasets is highly encouraged to enable other researchers to reproduce results and advance the field.}

\end{itemize}

Despite these risks, the substantial potential benefits of \gls{CP} make it a valuable area for continued exploration. Overcoming these challenges will require collaboration among researchers from various disciplines, including computer vision, communication technology, and vehicle engineering. Companies and governments should actively work together to establish standards for different V2X applications and develop compatible \gls{CP} systems. Finally, embracing open-source practices can significantly assist the research community in reproducing results and focusing on new challenges.

%% file: sections/06_conclusion.tex
\section{Conclusion} \label{sec:conclusion}

\noindent In this paper, we systematically reviewed recent research on \acrfull{CP}. We propose a structured taxonomy categorized by modality, collaboration type, and task, encompassing object detection, tracking, motion prediction, segmentation, lane detection, and multi-task or task-agnostic pipelines. The review also examined advanced techniques addressing real-world challenges, including pose errors, latency, bandwidth limitations, communication interruptions, domain shifts, heterogeneity, and adversarial attacks. Furthermore, we conducted a comparative analysis of these approaches, highlighting their strengths and limitations, and reviewed \gls{CP} evaluation methods, ranging from real-world datasets and synthetic datasets to experiments in real-world and simulated environments. Key limitations in current evaluation scenarios and metrics were identified, alongside challenges and opportunities in hardware, software, and evaluation methodologies.

A central motivation for \gls{CP} is to address visual occlusions and complement ego-perception systems. However, current research often overlooks the necessity of ensuring compatibility between \gls{CP} and ego-perception pipelines, as well as the importance of triggers to selectively activate collaboration under appropriate conditions. To assess \gls{CP}’s effectiveness in addressing visual occlusions, novel evaluation approaches aligned with its goals are essential. This review underscores the urgent need for large-scale \gls{CP} datasets that reflect realistic setups and diverse scenarios, which are pivotal for advancing the field.

Future work must prioritize the development of appropriate evaluation methodologies and large-scale datasets. An open-source co-simulation framework that represents realistic real-world scenarios and a unified collaborative driving framework encompassing the entire lifecycle, from research and development to deployment and validation, could significantly accelerate \gls{CP} advancements and its real-world implementation. Bridging the deployment gap should remain a key focus for future investigations.

Through this systematic review, we re-evaluate the concrete role of \gls{CP} in \gls{CAVs}. Revolutionizing evaluation methods and addressing deployment challenges will help transition \gls{CP} systems from lab prototypes to real-world applications. As \gls{CP} systems integrate communication, vehicular, and computer vision technologies, their progress will require interdisciplinary collaboration to enable the practical deployment of sophisticated \gls{CP} solutions.

Given the time constraints of this survey, our literature collection was finalized in March 2024, while our review focuses on research published in the past five years (2019–2023). However, our systematic review protocol and curated paper set provide a solid foundation for future researchers to extend this study. By applying forward snowballing, researchers can efficiently update the review with high-quality, cutting-edge research beyond our collection period.

\section*{Acknowledgments}
This work is developed within the framework of the ”VALISENS” project, funded by the Federal Ministry for Economic Affairs and Climate Action (BMWK). The authors are solely responsible for the content of this publication.

%% file: sections/07_appendix.tex
\begin{table*}
\centering
\caption{Overview of the methods for collaborative object detection (COD) based on BEV and 3D representations. V: Vehicle, I: Infrastructure, Raw: Raw data fusion, Trad Feat: Traditional Feature Fusion, Atten Feat: Attention Feature Fusion, Obj Fusion: Object-level Fusion, Graph: Graph-based Fusion.}
\label{tab:cod_1}
\begin{tabular}{c|c|c|c|c|c|c}
\toprule
Paper & Modality & Scheme & Year & Entity & Fusion & Code \\ \midrule
JointPerception\cite{JointPerceptionScheme-2022-ahmeda} & LiDAR & Early & 2022 & V & Raw & \xmark \\
RAO\cite{RobustRealtimeMultivehicle-2023-zhangb} & LiDAR & Early & 2023 & V & Raw &  \xmark \\
EdgeCooper\cite{EdgeCooperNetworkAwareCooperative-2024-luoa} & LiDAR & Early & 2024 & V,I & Trad Feat & \xmark \\
FastClustering\cite{FastClusteringCooperative-2024-kuanga} & LiDAR & Early & 2024 & V & Raw & \xmark  \\
\midrule
F-cooper\cite{FcooperFeatureBased-2019-chend} & LiDAR & Intermediate & 2019 & V & Trad Feat & \cmark \\
AFS-COD\cite{BandwidthAdaptiveFeatureSharing-2020-marvastia} & LiDAR & Intermediate & 2020 & V & Trad Feat & \xmark \\
FS-COD\cite{CooperativeLIDARObject-2020-marvastia} & LiDAR & Intermediate & 2020 & V & Trad Feat & \xmark \\
CoFF\cite{CoFFCooperativeSpatial-2021-guoa} & LiDAR & Intermediate & 2021 & V & Trad Feat & \xmark \\
SyncNet\cite{LatencyAwareCollaborativePerception-Lei-22} & LiDAR & Intermediate & 2022 & V & Trad Feat & \cmark \\
PillarGrid\cite{PillarGridDeepLearningBased-2022-baia} & LiDAR & Intermediate & 2022 & V,I & Trad Feat & \xmark \\
Slim-FCP\cite{SlimFCPLightweightFeatureBasedCooperative-2022-guoa} & LiDAR & Intermediate & 2022 & V & Trad Feat & \xmark \\
AdaptiveFeature\cite{AdaptiveFeatureFusion-2023-qiaob} & LiDAR & Intermediate & 2023 & V & Trad Feat & \cmark \\
CoBEVFlow\cite{AsynchronyRobustCollaborativePerception--weic} & LiDAR & Intermediate & 2023 & V & Trad Feat & \cmark \\
DI-V2X\cite{DIV2XLearningDomainInvariant-2023-xiangb} & LiDAR & Intermediate & 2023 & V,I & Trad Feat & \cmark \\
DFS\cite{DynamicFeatureSharing-2023-baia} & LiDAR & Intermediate & 2023 & V,I & Trad Feat & \xmark  \\
FFNet\cite{FlowBasedFeatureFusion-2023-yue} & LiDAR & Intermediate & 2023 & V,I & Trad Feat & \cmark \\
CoAlign\cite{RobustCollaborative3D-2023-lub} & LiDAR & Intermediate & 2023 & V & Trad Feat & \cmark \\
VINet\cite{VINetLightweightScalable-2023-baia} & LiDAR & Intermediate & 2023 & V,I & Trad Feat & \xmark  \\
FDA\cite{BreakingDataSilos-2024-lic} & LiDAR & Intermediate & 2024 & V,I & Trad Feat & \xmark \\
HP3D-V2V\cite{HP3DV2VHighPrecision3D-2024-chena} & LiDAR & Intermediate & 2024 & V & Trad Feat & \xmark  \\
MACP\cite{MACPEfficientModel-2024-maa} & LiDAR & Intermediate & 2024 & V & Trad Feat & \cmark \\
S2R-ViT\cite{S2RViTMultiAgentCooperative-2024-lib} & LiDAR & Intermediate & 2024 & V & Trad Feat & \xmark \\
Select2Col\cite{Select2ColLeveragingSpatialTemporal-2024-liuc} & LiDAR & Intermediate & 2024 & V & Trad Feat & \cmark \\
PillarAttention\cite{PillarAttentionEncoder-2024-bai} & LiDAR & Intermediate & 2024 & V,I & Trad Feat & \xmark  \\
DUSA\cite{DUSADecoupledUnsupervised-2023-kongc} & LiDAR & Intermediate & 2023 & V,I & NA & \cmark \\
UMC\cite{UMCUnifiedBandwidthefficient-2023-wangb} & LiDAR & Intermediate & 2023 & V & Graph & \cmark \\
HPL-ViT\cite{HPLViTUnifiedPerception-2023-liuc} & LiDAR & Intermediate & 2024 & V & Graph & \xmark \\
F-Transformer\cite{FTransformerPointCloud-2022-wanga} & LiDAR & Intermediate & 2019 & V & Atten Feat & \xmark \\
CRCNet\cite{ComplementarityEnhancedRedundancyMinimizedCollaboration-2022-luob} & LiDAR & Intermediate & 2022 & V & Atten Feat & \xmark \\
V2X-ViT\cite{V2XViTVehicletoEverythingCooperative-2022-xub} & LiDAR & Intermediate & 2022 & V,I & Atten Feat & \cmark \\
MPDA\cite{BridgingDomainGap-2023-xua} & LiDAR & Intermediate & 2023 & V,I & Atten Feat & \cmark \\
Co3D\cite{Collaborative3DObject-2023-wanga} & LiDAR & Intermediate & 2023 & V,I & Atten Feat & \xmark\\
FeaCo\cite{FeaCoReachingRobust-2023-guc} & LiDAR & Intermediate & 2023 & V & Atten Feat & \cmark \\
How2comm\cite{How2commCommunicationEfficientCollaborationPragmatic--yanga} & LiDAR & Intermediate & 2023 & V & Atten Feat & \cmark \\
LCRN\cite{LearningVehicletoVehicleCooperative-2023-lia} & LiDAR & Intermediate & 2023 & V & Atten Feat & \xmark  \\
SCOPE\cite{SpatioTemporalDomainAwareness-2023-yangb} & LiDAR & Intermediate & 2023 & V & Atten Feat & \cmark \\
What2comm\cite{What2commCommunicationefficientCollaborative-2023-yange} & LiDAR & Intermediate & 2023 & V,I & Atten Feat & \xmark  \\
CenterCoop\cite{CenterCoopCenterBasedFeature-2024-zhou} & LiDAR & Intermediate & 2024 & V,I & Atten Feat & \xmark  \\
V2X-INCOP\cite{InterruptionAwareCooperativePerception-2024-renc} & LiDAR & Intermediate & 2024 & V,I & Atten Feat & \xmark  \\
MKD-Cooper\cite{MKDCooperCooperative3D-2024-lia} & LiDAR & Intermediate & 2024 & V & Atten Feat & \cmark \\
Self-Adaptive\cite{SelfSupervisedAdaptiveWeighting-2024-liu} & LiDAR & Intermediate & 2024 & V & Atten Feat & \xmark  \\
SemanticComm\cite{SemanticCommunicationCooperative-2024-shenga} & LiDAR & Intermediate & 2024 & V & Atten Feat & \xmark  \\
V2VFormer\cite{V2VFormerVehicletoVehicleCooperative-2024-lina} & LiDAR & Intermediate & 2024 & V & Atten Feat & \xmark  \\
\midrule
FL-Dynamic \cite{DistributedDynamicMap-2021-zhanga} & LiDAR & Late & 2021 & V & Obj & \cmark \\
Env-T2TF\cite{EnvironmentawareOptimizationTracktoTrack-2022-volka} & LiDAR & Late & 2022 & V & Obj & \xmark  \\
Co-perception\cite{CooperativePerceptionSystem-2023-songc} & LiDAR & Late & 2023 & V & Obj & \xmark  \\
Among Us\cite{UsAdversariallyRobust-2023-lia} & LiDAR & Late & 2023 & V & Obj & \cmark \\
Collective PV-RCNN\cite{CollectivePVRCNNNovel-2023-teufela} & LiDAR & Late & 2023 & V & Trad Feat & \xmark  \\
Late-CNN\cite{CooperativePerceptionLearningBased-2023-liub} & LiDAR & Late & 2023 & V & Obj & \xmark  \\
Model-Agnostic\cite{ModelAgnosticMultiAgentPerception-2023-xub} & LiDAR & Late & 2023 & V & Obj & \cmark \\
Double-M\cite{UncertaintyQuantificationCollaborative-2023-sua} & LiDAR & Late & 2023 & V,I & Obj & \cmark \\
\midrule
Pillar-based CP\cite{PillarBasedCooperativePerception-2022-wanga} & LiDAR & Hybrid & 2022 & V & Hybrid (Raw, Trad Feat, Obj) & \xmark \\
ML-Cooper\cite{SoftActorCriticBased-2022-xiea} & LiDAR & Hybrid & 2022 & V & Hybrid (Raw, Trad Feat, Obj) & \xmark \\
FPV-RCNN\cite{KeypointsBasedDeepFeature-2022-yuanc} & LiDAR & Hybrid & 2024 & V & Trad Feat & \cmark \\
Hybrid-CP\cite{RegionBasedHybridCollaborative-2024-liu} & LiDAR & Hybrid & 2024 & V & Hybrid (Atten Feat, Obj) & \xmark \\
FreeAlign\cite{RobustCollaborativePerception-2024-lei} & LiDAR & Hybrid & 2024 & V & Trad Feat & \cmark \\
\midrule
CoCa3D\cite{CollaborationHelpsCamera-2023-hub} & Camera & Intermediate & 2023 & V,I & Trad Feat & \cmark \\
ActFormer\cite{ActFormerScalableCollaborative-2024-huangc} & Camera & Intermediate & 2024 & V & Atten Feat & \cmark \\
EMIFF\cite{EMIFFEnhancedMultiscale-2024-wangc} & Camera & Intermediate & 2024 & V,I & Trad Feat & \cmark \\
QUEST\cite{QUESTQueryStream-2023-fand} & Camera & Intermediate & 2024 & V,I & Trad Feat & \xmark \\
\bottomrule
\end{tabular}
\end{table*}

\begin{table*}[ht]
\centering
\caption{Overview of the methods for collaborative object detection (COD) based on BEV and 3D representations. (Continued)}
\label{tab:cod_2}
\begin{tabular}{c|c|c|c|c|c|c}
\toprule
Paper & Modality & Scheme & Year & Entity & Fusion & Code \\ \midrule
ViT-FuseNet\cite{ViTFuseNetMultimodalFusion-2024-zhoua} & LiDAR, Camera & Early & 2024 & V,I & Atten Feat & \xmark \\
Multi-vehicle fusion\cite{MultimodalCooperative3D-2023-chia} & LiDAR, Camera & Intermediate & 2023 & V & Trad Feat & \xmark \\
V2VFusion\cite{V2VFusionMultimodalFusion-2023-zhanga} & LiDAR, Camera & Intermediate & 2023 & V & Trad Feat & \xmark \\
HEAL\cite{ExtensibleFrameworkOpen-2024-luc} & LiDAR, Camera & Intermediate & 2024 & V & Trad Feat & \cmark \\
HGAN\cite{MultiModalVirtualRealFusion-2022-zhanga} & LiDAR, Camera & Intermediate & 2022 & V,I & Hybrid & \xmark \\
Distilled Co-Graph\cite{LearningDistilledCollaboration-2021-lib} & LiDAR, Camera & Intermediate & 2021 & V & Graph & \cmark \\
HM-ViT\cite{HMViTHeteromodalVehicletoVehicle-2023-xiangb} & LiDAR, Camera & Intermediate & 2023 & V & Graph & \cmark \\
PIXOR\cite{OPV2VOpenBenchmark-2022-xub} & LiDAR, Camera & Intermediate & 2022 & V & Atten Feat & \cmark \\
Where2comm\cite{Where2commCommunicationEfficientCollaborative-2022-huc} & LiDAR, Camera & Intermediate & 2022 & V & Atten Feat & \cmark \\
MCoT\cite{MCoTMultiModalVehicletoVehicle-2023-shia} & LiDAR, Camera & Intermediate & 2023 & V & Atten Feat & \xmark \\
PAFNet\cite{PAFNetPillarAttention-2024-wanga} & LiDAR, Camera & Intermediate & 2024 & V,I & Atten Feat & \xmark \\
V2VFormer++\cite{V2VFormerMultiModalVehicletoVehicle-2024-yina} & LiDAR, Camera & Intermediate & 2024 & V & Atten Feat & \xmark \\
\midrule
TCLF\cite{DAIRV2XLargeScaleDataset-2022-yua} & LiDAR, Camera & Late & 2022 & V,I & Obj & \cmark \\
VICOD\cite{MultistageFusionApproach-2022-yua} & LiDAR, Camera & Late & 2022 & V,I & Obj & \xmark \\
\bottomrule
\end{tabular}
\end{table*}